%% file: top.tex
\documentclass[10pt,journal,compsoc]{IEEEtran}

\usepackage[OT1]{fontenc} 
\usepackage[utf8]{inputenc}
\usepackage{amsmath,amsfonts,amssymb,amsthm}
\usepackage{graphicx}
\usepackage{color}
\usepackage{caption}
\usepackage{subcaption}
\usepackage{algorithm}
\usepackage[noend]{algpseudocode}
\usepackage{xspace}
\usepackage{ifthen}
\usepackage[nocompress]{cite}
\usepackage{array}
\usepackage{multirow}
\usepackage{multicol}
\usepackage[pagebackref=true,breaklinks=true,colorlinks,bookmarks=false]{hyperref}
\usepackage{pifont}%
\usepackage{soul}
\usepackage{tabularx}
\usepackage{booktabs}
\usepackage{dsfont}
\usepackage{hyperref}
\usepackage[table]{xcolor}
\usepackage[misc]{ifsym}

\newcommand\labelLastPage{%
  \edef\@currentlabel{\thepage}%
  \label{LastPage}%
}

\definecolor{cvprblue}{rgb}{0.21,0.49,0.74}
\definecolor{leafgreen}{rgb}{0.047,0.533,0.255}
\hypersetup{
    linkcolor = leafgreen,
    citecolor = cvprblue,
}

\graphicspath{ {gfx/} }
\input{commands}

\begin{document}
\title{HUGSIM: A Real-Time, Photo-Realistic and Closed-Loop Simulator for Autonomous Driving}
\author{Hongyu~Zhou \qquad Longzhong Lin \qquad Jiabao Wang \qquad Yichong Lu \qquad Dongfeng Bai \qquad Bingbing Liu \qquad Yue Wang \qquad Andreas~Geiger \qquad Yiyi Liao$\textsuperscript{\Letter}$

\IEEEcompsocitemizethanks{
\IEEEcompsocthanksitem H. Zhou, L. Lin, J. Wang, Y. Lu, Y. Wang and Y. Liao are affiliated with Zhejiang University, China.  
\IEEEcompsocthanksitem D. Bai and B, Liu are affiliated with Huawei, China.  
\IEEEcompsocthanksitem A. Geiger is affiliated with the Autonomous Vision Group, University of T{\"u}bingen and T{\"u}bingen AI Center, Germany.
\IEEEcompsocthanksitem Corresponding author: Yiyi Liao
\IEEEcompsocthanksitem E-mails: \href{hongyu_zhou@zju.edu.cn}{hongyu\_zhou@zju.edu.cn}, \href{yiyi.liao@zju.edu.cn}{yiyi.liao@zju.edu.cn}
\IEEEcompsocthanksitem Project Page: \href{https://xdimlab.github.io/HUGSIM}{https://xdimlab.github.io/HUGSIM}
}

}

\IEEEtitleabstractindextext{%
\input{sec_abstract}
}

\maketitle
\IEEEdisplaynontitleabstractindextext
\IEEEpeerreviewmaketitle

\input{sec_introduction}

\input{sec_related}

\input{sec_render}

\input{sec_simulation}

\input{sec_render_eval}

\input{sec_benchmark}
\input{sec_conclusion}

{
\bibliographystyle{IEEEtranS}
\bibliography{bibliography_long, references}
}

\vspace{-1.4 cm}%
\begin{IEEEbiography}[{\includegraphics[width=1in,clip,keepaspectratio]{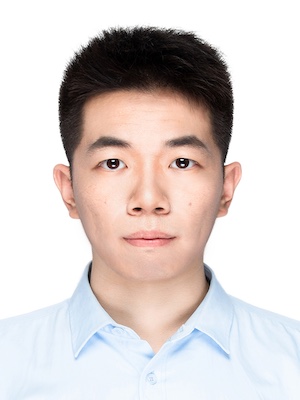}}]{Hongyu Zhou}
is a Ph.D. student at Zhejiang University. Before that, he received his B.S. degree from College of Software Engineering, Tongji University, China. His research interests include 3D computer vision and 3D reconstruction and generative models.
\end{IEEEbiography}
\vspace{-1 cm}%

\begin{IEEEbiography}[{\includegraphics[width=1in,clip,keepaspectratio]{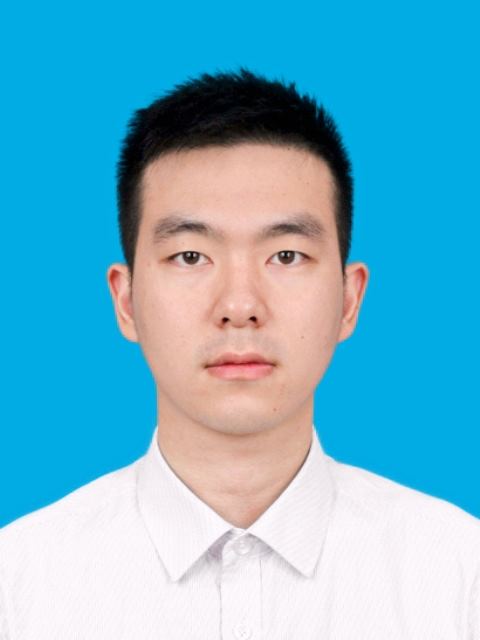}}]{Longzhong Lin}
is a Ph.D. student at Zhejiang University. Before that, he received his B.S. degree from the College of  Control Science and Engineering, Zhejiang University, China. His research interests include autonomous robots and robot learning.
\end{IEEEbiography}
\vspace{-1 cm}%

\begin{IEEEbiography}[{\includegraphics[width=1in,clip,keepaspectratio]{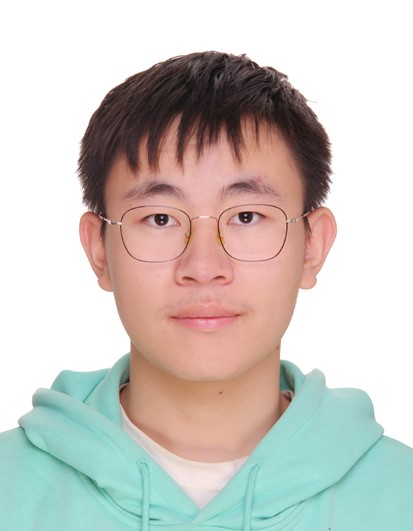}}]{Jiabao Wang} is an undergraduate student at Zhejiang University in the College of Information Science and Electronic Engineering. His research interests include 3D computer vision and reinforcement learning.
\end{IEEEbiography}
\vspace{-1 cm}%

\begin{IEEEbiography}[{\includegraphics[width=1in,clip,keepaspectratio]{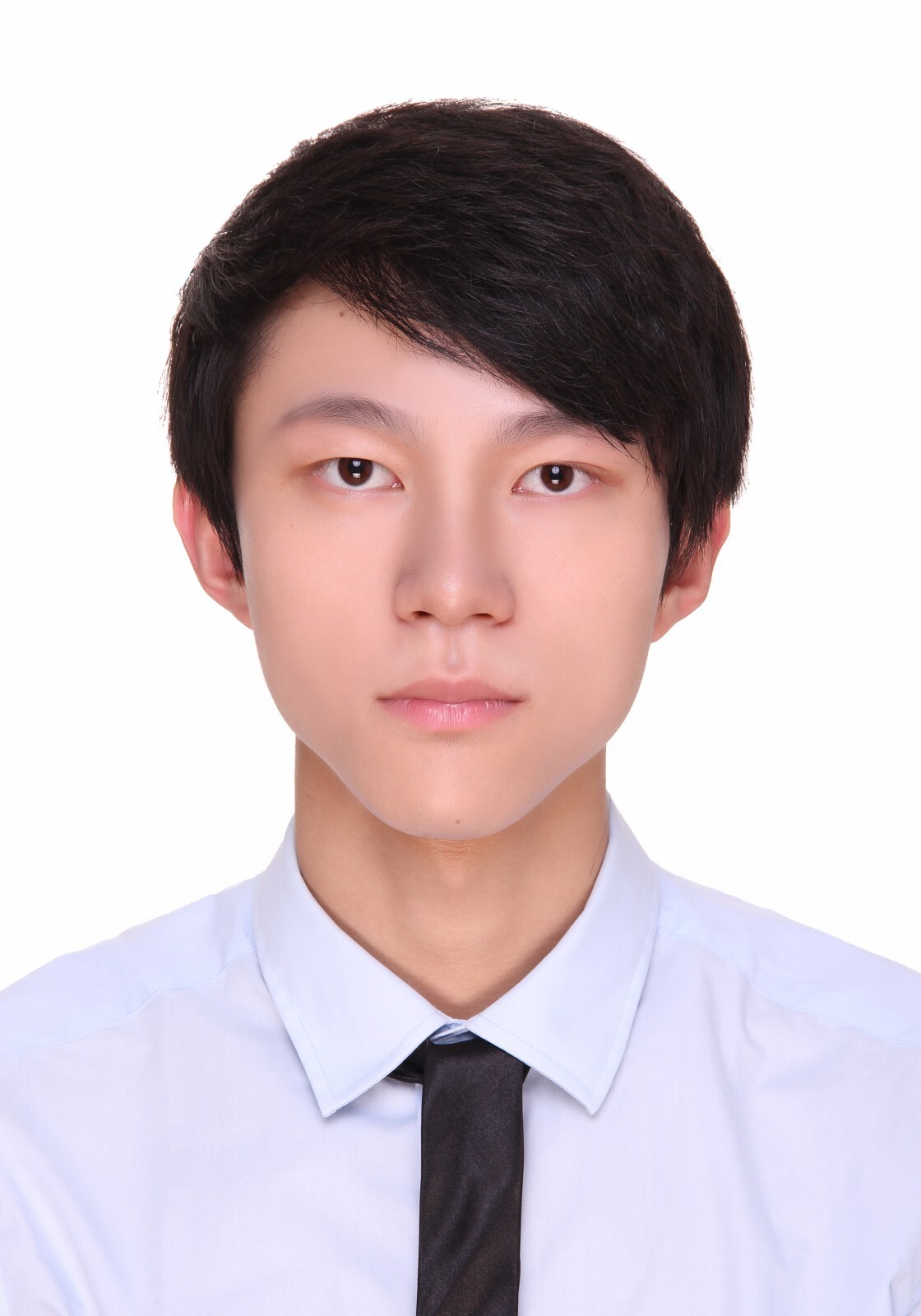}}]{Yichong Lu}
Yichong Lu is a Master's student in Zhejiang University. Before that, he obtained bachelor degree in the College of Information Science and Electronic Engineering, Zhejiang University. His research interest lies in 3D computer vision, including scene understanding, 3D reconstruction and 3D generative models.
\end{IEEEbiography}
\vspace{-1 cm}%

\begin{IEEEbiography}[{\includegraphics[width=1in,clip,keepaspectratio]{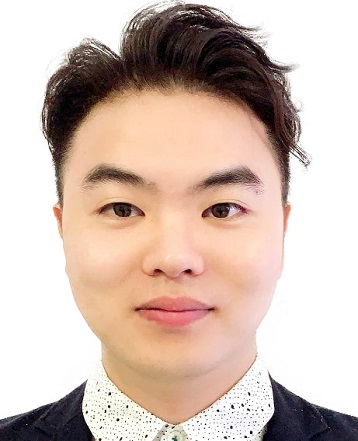}}]{Dongfeng Bai}
received the M.S. degree in the Department of Artificial Intelligence from Xi’an Jiaotong University, Xi’an, China, in 2020. He is a researcher in Noah’s Ark Lab, Huawei. His research interests include 3D reconstruction, neural rendering and 3D generation.
\end{IEEEbiography}
\vspace{-1 cm}%

\begin{IEEEbiography}[{\includegraphics[width=1in,clip,keepaspectratio]{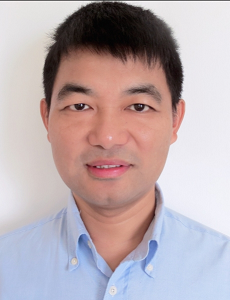}}]{Bingbing Liu}
(M09) is a Technology Specialist with Noah's Ark Lab, Huawei. Previously he worked as a Research Scientist with Institute for Infocomm Research, A*STAR, Singapore. He received his Bachelor degree from Harbin Institute of Technology, China and the Ph.D degree from Nanyang Technological University, Singapore respectively. His research interests are computer vision technologies for autonomous driving and robotics.
\end{IEEEbiography}
\vspace{-1 cm}%

\begin{IEEEbiography}[{\includegraphics[width=1in,clip,keepaspectratio]{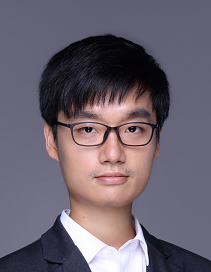}}]{Yue Wang}
received the Ph.D. degree from the Department of Control Science and Engineering, Zhejiang University, Hangzhou, China, in 2016. He is currently working as a Professor with the Department of Control Science and Engineering, Zhejiang University. His current research interests include autonomous robots and robot learning.
\end{IEEEbiography}
\vspace{-1 cm}%

\begin{IEEEbiography}[{\includegraphics[width=1in,clip,keepaspectratio]{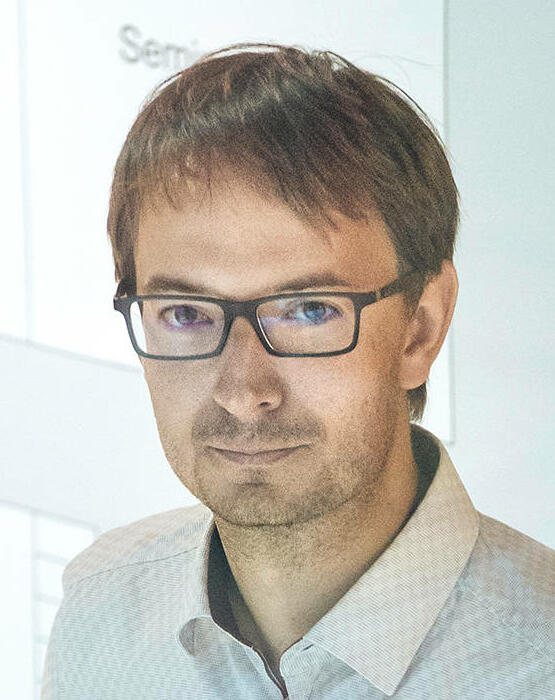}}]{Andreas Geiger}
received his Diploma in computer science and his Ph.D. degree from Karlsruhe Institute of Technology in 2008 and 2013. Currently, he is leading the Autonomous Vision Group at the University of T\"ubingen. He is also a core faculty member of the T\"ubingen AI Center. His research interests include computer vision, machine learning and scene understanding with a focus on self-driving vehicles.
\end{IEEEbiography}
\vspace{-1 cm}%

\begin{IEEEbiography}[{\includegraphics[width=1in,clip,keepaspectratio]{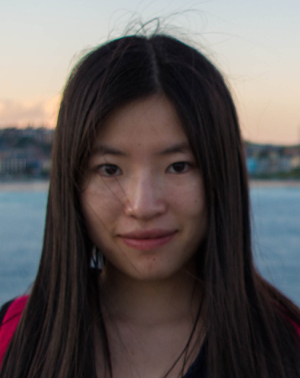}}]{Yiyi Liao} is an assistant professor at Zhejiang University, leading the X-Dimensional Representations Lab. She received her Ph.D. in Control Science and Engineering from Zhejiang University in June 2018 and her B.S. degree from Xi’an Jiaotong University in 2013. Her research interests include 3D vision and scene understanding.
\end{IEEEbiography}
\vspace{-1 cm}%

\labelLastPage
\newpage
\pagenumbering{gobble}

\input{sec_appendix}

\end{document}

%% file: commands.tex
\newcommand{\bK}{\mathbf{K}}

\newcommand{\bx}{\mathbf{x}}

\newcommand{\bI}{\mathbf{I}}

\newcommand{\bS}{\mathbf{S}}
\newcommand{\bs}{\mathbf{s}}

\newcommand{\bW}{\mathbf{W}}

\newcommand{\bq}{\mathbf{q}}
\newcommand{\bR}{\mathbf{R}}

\newcommand{\bt}{\mathbf{t}}
\newcommand{\bJ}{\mathbf{J}}

\newcommand{\bD}{\mathbf{D}}

\newcommand{\bC}{\mathbf{C}}
\newcommand{\bA}{\mathbf{A}}

\newcommand{\bff}{\mathbf{f}}
\newcommand{\bF}{\mathbf{F}}

\newcommand{\bc}{\mathbf{c}}

\newcommand{\bL}{\mathbf{L}}

\newcommand{\bd}{\mathbf{d}}

\newcommand{\bb}{\mathbf{b}}

\newcommand{\bmu}{\boldsymbol{\mu}}
\newcommand{\bSigma}{\boldsymbol{\Sigma}}

\newcommand{\nR}{\mathbb{R}}

\newcommand{\cN}{\mathcal{N}}
\newcommand{\cM}{\mathcal{M}}

\newcommand{\cL}{\mathcal{L}}

\newcommand{\cS}{\mathcal{S}}

\newcommand{\figref}[1]{\Fig~\ref{#1}}
\newcommand{\secref}[1]{Section~\ref{#1}}

\renewcommand{\eqref}[1]{Eq.~\ref{#1}}
\newcommand{\tabref}[1]{Table~\ref{#1}}

\makeatletter
\DeclareRobustCommand\onedot{\futurelet\@let@token\@onedot}
\def\@onedot{\ifx\@let@token.\else.\null\fi\xspace}

\def\wrt{wrt\onedot}

\def\Fig{Fig\onedot}   
\makeatother

\definecolor{applegreen}{rgb}{0.55, 0.71, 0.0}

\newcommand{\boldparagraph}[1]{\vspace{0.2cm}\noindent{\bf #1:} }

\newcolumntype{P}[1]{>{\centering\arraybackslash}m{#1}}

\newcommand{\del}[1]{}

\newif\ifcomment
\commenttrue
\ifcomment
	\newcommand{\ag}[1]{ \noindent {\color{red} {\bf Andreas:} {#1}} }
	\newcommand{\yl}[1]{ \noindent {\color{cyan} {\bf Yiyi:} {#1}} }

\else
	\newcommand{\ag}[1]{}
	\newcommand{\jx}[1]{}
	\newcommand{\kc}[1]{}
	\newcommand{\yl}[1]{}
\fi

%% file: sec_abstract.tex
\parbox{0.918\textwidth}{
\begin{abstract}
In the past few decades, autonomous driving algorithms have made significant progress in perception, planning, and control. However, evaluating individual components does not fully reflect the performance of entire systems, highlighting the need for more holistic assessment methods. 
This motivates the development of HUGSIM, a closed-loop, photo-realistic, and real-time simulator for evaluating autonomous driving algorithms.
We achieve this by lifting captured 2D RGB images into the 3D space via 3D Gaussian Splatting, improving the rendering quality for closed-loop scenarios, and building the closed-loop environment. In terms of rendering, We tackle challenges of novel view synthesis in closed-loop scenarios, including viewpoint extrapolation and 360-degree vehicle rendering. Beyond novel view synthesis, HUGSIM further enables the full closed simulation loop, dynamically updating the ego and actor states and observations based on control commands.
Moreover, HUGSIM offers a comprehensive benchmark across more than 70 sequences from KITTI-360, Waymo, nuScenes, and PandaSet, along with over 400 varying scenarios, providing a fair and realistic evaluation platform for existing autonomous driving algorithms. HUGSIM not only serves as an intuitive evaluation benchmark but also unlocks the potential for fine-tuning autonomous driving algorithms in a photorealistic closed-loop setting.
\end{abstract}
}

%% file: sec_introduction.tex
\IEEEraisesectionheading{\section{Introduction}\label{sec:introduction}}

\IEEEPARstart{V}{ision}-based autonomous driving (AD) offers a promising route to achieving full autonomy at low cost. 
Nonetheless, rigorously testing a vision-based AD system in real-world environments poses significant challenges, primarily due to safety concerns and the inherent inefficiencies of such evaluations.
Moreover, the risk associated with assessing safety-critical scenarios in real settings is substantial, yet these situations are precisely where AD algorithms must demonstrate their reliability and effectiveness.

Existing methods \cite{geiger2012we, liao2022kitti, caesar2020nuscenes, Sun2020CVPR, xiao2021pandaset} primarily assess vision-based AD algorithms using collected datasets, offering open-loop evaluations of specific components like perception and planning. However, this approach falls short of capturing closed-loop performance \cite{dauner2023parting, codevilla2018end, codevilla2019exploring}, as it evaluates AD algorithms solely in pre-collected, safe and normal driving environments conducted by experienced drivers. This limitation underscores the urgent need for advanced simulators capable of conducting closed-loop evaluations of vision-based AD systems, enabling the safe exploration of out-of-domain and safety-critical scenarios.

To tackle this issue, closed-loop AD simulators have been investigated. Some studies \cite{gulino2024waymax, caesar2021nuplan, wen2023limsim} implement a closed-loop system for planning and control, assuming ground truth perception results are available, neglecting the importance of perception. Other works \cite{dosovitskiy2017carla, li2022metadrive} offer an end-to-end closed-loop system based on game engines, resulting in video-game-style simulators. Although these simulators provide numerous features for testing and training AD algorithms, significant domain gaps still exist between the real world and video games. Another downside of these simulators is the considerable time and manpower investment in manually creating assets and scenarios.

\input{figures/teaser/teaser}

In this paper, we bridge the gap between real-world evaluation and simulation by building a novel photorealistic AD simulator leveraging advanced novel view synthesis techniques.
Although many previous works \cite{kundu2022panoptic, zhou2024drivinggaussian, yan2024street, Zhou2024CVPR} tackle novel view synthesis of urban scenes, there remains a significant gap between urban novel view synthesis and an AD simulator.

First, a closed-loop simulator with interactive actors poses new challenges for the rendering process: 1) The trajectories of dynamic objects are originally observed from discrete timestamps, whereas a closed-loop simulator requires continuous trajectories.
2) Unlike typical novel view synthesis tasks, which evaluate on interpolated views, simulators require photo-realistic rendering even for extrapolated views, especially for lane areas.
3) Since actors are observed under unpredictable viewpoints due to the interaction between the ego vehicle and the actors, we should ensure that actors appear realistic from 360-degree viewpoints.
Moreover, there are extra implementation steps for building a full-fledged closed-loop simulator. For instance, a communication bridge between the simulator and AD algorithms is necessary; waypoints retrieved from AD algorithms must be converted into control commands to update the ego vehicle's position and actor trajectories need to be generated efficiently.

Recently, a concurrent work called NeuroNCAP \cite{ljungbergh2024neuroncap} introduced, to the best of our knowledge, the first publicly available photorealistic closed-loop simulator that offers high-fidelity rendering. However, NeuroNCAP is not specifically designed to address the challenges of viewpoint extrapolation, 360-degree high-fidelity actor rendering, or the efficient simulation of interactive safety-critical scenarios. 

In this paper, we propose HUGSIM, a novel vision-based AD simulator to address the aforementioned challenges, see \figref{fig:teaser}.
We start by extending 3D Gaussian Splatting (3DGS)~\cite{kerbl20233d} to model dynamic 3D scenes with multiple modalities, from RGB images and noisy 2D and 3D predictions.
This approach enables holistic scene reconstruction, including appearance, semantics, flow, depth, and motion of rigidly moving objects.
Note that semantic information is essential for collision detection, while flow and depth can be helpful for AD algorithms \cite{zhou2019does}. Importantly, we apply physical constraints to both dynamic vehicles and the ground to enhance rendering quality. Specifically, we regularize the trajectory of dynamic vehicles using a unicycle model, 
allowing for improving the rendering quality of dynamic vehicles given noisy 3D bounding box predictions.
We further introduce a multi-plane ground model to address lane distortion in extrapolated views, which works across different types of ground, including sloped surfaces.
In addition, we insert \textit{non-native} actors reconstructed from 3DRealCar~\cite{du20243drealcar}, ensuring both faithful geometry and appearance from arbitrary observation views, in contrast to partially observed \textit{native} vehicles in typical driving scenes. %
Moreover, we close the simulation loop by querying waypoints from AD algorithms, applying a linear quadratic regulator (LQR) for control, extracting the multiple ground planes, and updating the pose of ego vehicles and actors. By leveraging HD maps, we deploy the  Intelligent Driver Model (IDM) planning strategy \cite{treiber2000congested} to simulate normal actor behavior. To simulate safety-critical scenarios, we design an attack planning strategy to generate aggressive actor behavior, which works effectively even in scenes without HD maps.

Additionally, we encapsulate the entire scene into an interactive closed-loop simulator and provide Gymnasium APIs \cite{brockman2016openai} to facilitate AD evaluation and the potential application of reinforcement learning in our simulator. Our simulator operates in real-time, as it introduces minimal computational overhead compared to 3DGS.
It supports scenes from KITTI-360 \cite{liao2022kitti}, Waymo \cite{Sun2020CVPR}, nuScenes \cite{caesar2020nuscenes}, and PandaSet \cite{xiao2021pandaset}. To fairly evaluate different AD algorithms, we propose an evaluation metric named HD-Score, inspired by NAVSIM \cite{dauner2024navsim} and DriveArena \cite{yang2024drivearena}. The HD-Score reflects AD performance based on $NC$ (No Collision), $DAC$ (Drivable Area Compliance), $TTC$ (Time to Collision),  $COM$ (Comfort), and $R_c$ (Route Completion).

We summarize the contributions of this paper as follows:
\begin{itemize}
\item We present a novel simulator for autonomous driving, characterized by being closed-loop, photo-realistic, and real-time, bridging the gap between urban scene novel view synthesis and autonomous driving simulators.
\item To address the specific rendering challenges in the simulator, we leverage physical constraints and non-native actors to improve fidelity, surpassing previous novel view synthesis approaches.
\item We propose an efficient actor trajectory generation strategy, enabling the simulation of aggressive driving behaviors even without HD maps.
\item We introduce a new benchmark for fair evaluation of AD algorithms, offering more photorealistic simulation environments compared to existing closed-loop vision-based AD simulators. Additionally, the benchmark provides a variety of scenes built upon multiple datasets, with varying scenarios with different levels of difficulty.
\end{itemize}

This journal paper is an extension of a conference paper
published at CVPR 2024, HUGS \cite{Zhou2024CVPR}. In comparison to HUGS, we 1) extend the novel view synthesis task by constructing a closed-loop simulator; 2) design an efficient strategy for simulating safety-critical scenarios without relying on HD maps; 3) propose a multi-plane ground Gaussian model and real captured vehicle models to enhance appearance fidelity; 4) introduce a novel benchmark for fair evaluation of vision-based AD algorithms.

%% file: figures/teaser/teaser.tex
\begin{figure*}
    \centering
    \includegraphics[width=\textwidth]{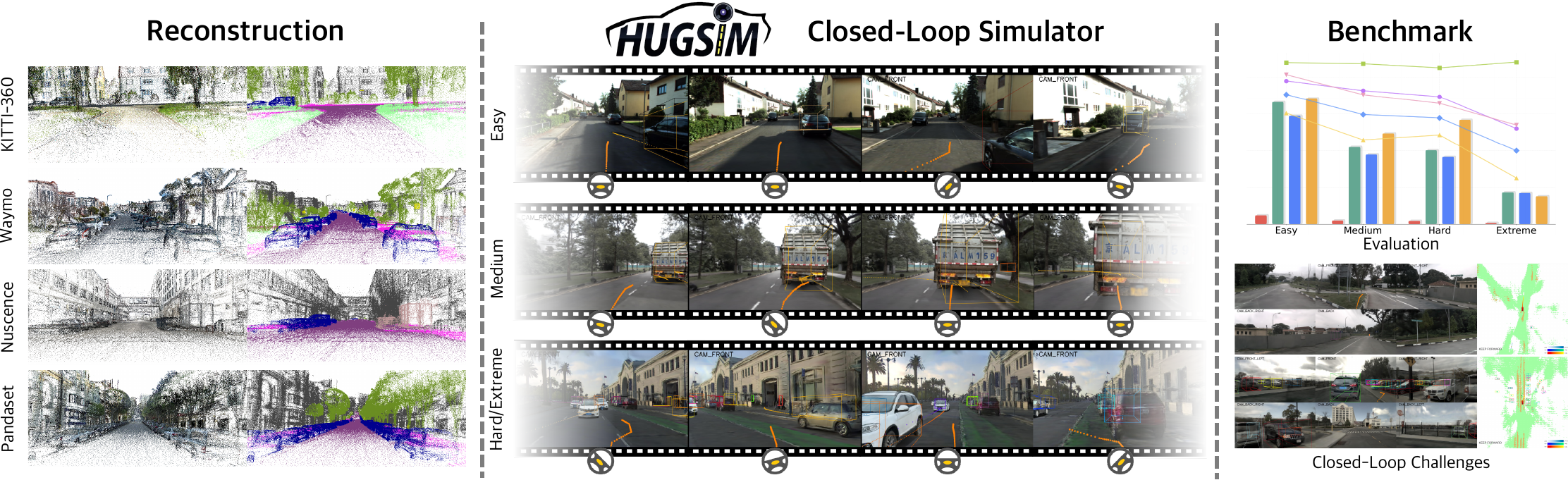}
    \caption{\textbf{Overview of HUGSIM}. We propose HUGSIM, a photorealistic closed-loop simulator for AD, supporting a variety of scenes from different datasets and scenarios with varying levels of difficulty. HUGSIM also provides a closed-loop benchmark and presents challenges for existing AD algorithms. }
\label{fig:teaser}
\vspace{-0.5cm}
\end{figure*}

%% file: sec_related.tex
\input{tables/relatedwork}

\section{Related Work}

In this section, we first review methods in urban scene reconstruction, a core task for enabling AD simulators. Then we discuss existing datasets, benchmarks and simulators used for training or evaluating AD algorithms. We compare various features of representative relative work in \tabref{tab:relatedwork}.

\subsection{Urban Scene Reconstruction}

\boldparagraph{Static Scenes}
Numerous studies have been conducted to reconstruct urban scenes using various methods. These methods can be categorized into four classes: point-based \cite{agarwal2011building, schonberger2016structure}, mesh-based \cite{gallup2010piecewise, lafarge2012hybrid}, NeRF-based \cite{martin2021nerf, rematas2022urban, wimbauer2023behind, tancik2022block, zhang2023nerflets, guo2023streetsurf, lu2023urban, miao2024efficient} and 3DGS-based \cite{han2024ggs, li2024ggrt, shi2024dhgs}. While point-based and mesh-based methods demonstrate faithful reconstructions, they struggle to recover all aspects of the scene, especially when it comes to high-quality appearance modeling. In contrast, NeRF-based models not only reconstruct scene appearance but also enable high-quality rendering from novel viewpoints, while 3DGS-based models support real-time rendering without sacrificing quality. However, these approaches are primarily designed for static scenes, lacking the ability to handle dynamic urban environments. In this study, our focus lies in addressing the challenges of dynamic urban scenes.

\boldparagraph{Dynamic Scenes}
Several methods have also been developed to address the reconstruction of dynamic urban scenes. Many of these approaches rely on the availability of accurate 3D bounding boxes for moving objects in order to separate the dynamic elements from the static components, as seen in \cite{ost2021neural, wu2023mars, yang2023unisim, zhou2024drivinggaussian, fischer2024dynamic, tonderski2024neurad}. PNF\cite{kundu2022panoptic}, NeuRAD \cite{tonderski2024neurad} and StreetGaussians \cite{yan2024street} takes a different approach by leveraging monocular-based or LiDAR-based 3D bounding box predictions and proposes a joint optimization of object poses during the reconstruction process. However, our experimental observations indicate that the straightforward optimization of object poses yields unsatisfactory results due to the absence of physical constraints. Another method, SUDS \cite{turki2023suds}, avoids the use of 3D bounding boxes by grouping the scene based on learned feature fields. However, the accuracy of this approach lags behind. In parallel, EmerNeRF \cite{yang2023emernerf} and PVG \cite{chen2023periodic} follows a similar idea to SUDS by decomposing the scene purely into static and dynamic components. 
In our research, we have the capability to further decompose individual dynamic objects within the scene and estimate their motion. This reconstructed motion can then be used to simulate driving behavior within the simulator.

\boldparagraph{Extrapolated View Rendering}
Although the urban reconstruction methods mentioned above can render high-fidelity images in interpolated views, most struggle with rendering in extrapolated views, where artifacts particularly on the lanes are common. Some methods, such as \cite{yu2024sgd, wang2024freevs}, leverage diffusion models to add additional views for supervision, addressing the problem of sparse input views in urban scene reconstruction. However, diffusion models introduce challenges such as inconsistency between views, increased training complexity, and a slowdown in training speed due to the heavy computational burden of these models.
GGS \cite{han2024ggs} presents a generalizable model based on MVSplat \cite{chen2024mvsplat}, incorporating a virtual lane generation strategy during training to address the extrapolated view problem. While this approach significantly improves fidelity in extrapolated views, MVSplat allows only a few frames as input, which may restrict scalability and lead to multi-view misalignment.
In comparison, the concurrent works AutoSplat \cite{khan2024autosplat} and RoGS \cite{feng2024rogs} apply physical priors to constrain ground Gaussians, similar to our approach. However, AutoSplat relies on LiDAR data for initialization, and RoGS uniformly distributes Gaussians across the ground plane, with both approaches fixing Gaussian positions on the plane. These methods use a large number of Gaussians to model non-textured areas, which we find unnecessary and inefficient. By optimizing additional parameters such as position and scale, HUGSIM achieves better performance without requiring as many Gaussians.

The related work mentioned above primarily focus on novel view synthesis but still fall short of being fully developed closed-loop simulator, as shown in \tabref{tab:relatedwork}.

\input{figures/method/pipeline/pipeline}

\subsection{Benchmarks}

\boldparagraph{Open-Loop}
Most existing datasets and benchmarks \cite{geiger2012we, liao2022kitti, caesar2020nuscenes, xiao2021pandaset, Sun2020CVPR}  for AD algorithms follow an open-loop approach, evaluating individual components of the algorithms separately. For instance, in the perception component, tasks such as semantic segmentation, bounding box detection, and lane detection are assessed, while the planning component evaluates tasks like route planning, behavior prediction, and trajectory prediction. 
Although these open-loop benchmarks provide detailed and convincing performance evaluations of each part, they evaluate the performance of AD algorithms using sensory data collected by experts, lacking scenarios deviated from the expert-collected sensory data. Without closed-loop feedback, the long-term consequences of these deviations remain unexplored, which could be critical for understanding the robustness and safety of AD systems in real-world conditions. 
NAVSIM \cite{dauner2024navsim} provides a benchmark positioned between open-loop and closed-loop evaluations. Although it is a non-reactive simulator and lacks the capability for novel view synthesis, it computes closed-loop metrics by forecasting the planning trajectory for a few seconds into the future. However, NAVSIM is limited to short time horizons and does not address how accumulated deviations over time impact driving safety.

\boldparagraph{Closed-Loop}
A number of closed-loop simulators have attempted to address the limitations of open-loop benchmarks in autonomous driving. Some simulators \cite{gulino2024waymax, caesar2021nuplan, wen2023limsim} provide ground-truth positions and rotations of other vehicles, lacking the perception aspect of the AD system, which is crucial for a comprehensive evaluation. Other works \cite{dosovitskiy2017carla, li2022metadrive} use game engines to create simulators, offering high edit-ability but often lacking photorealism and requiring extensive manual design of scenes, making them costly and less scalable. Rapid advances in video diffusion models \cite{wang2023drivedreamer, hu2023gaia, yang2024generalized, gao2024vista} have shown promise in generating photorealistic driving videos. DriveArena \cite{yang2024drivearena} uses a video diffusion model to build a closed-loop simulator, controlling scene generation through scene layouts. However, these models still suffer from issues like temporal inconsistency and significant computational demands. UniSim \cite{yang2023unisim} and NeuroNcap \cite{ljungbergh2024neuroncap} takes a different approach by creating a NeRF-based simulator that enables photorealistic, closed-loop simulations. However, UniSim is not an open-source work, while NeuroNCAP has several shortcomings, including non-real-time rendering, sub-optimal quality in extrapolated views, and fully manual actor behavior design. In contrast, HUGSIM addresses these challenges by offering real-time performance, improved fidelity in extrapolated views, and efficient, automated generation of actor behaviors.

%% file: tables/relatedwork.tex
\begin{table*}
\centering
\small

\definecolor{leafgreen}{rgb}{0.05, 0.54, 0.25}
\definecolor{orange}{rgb}{1.0, 0.6, 0.0}
\def \tick{\color{leafgreen}{\ding{52}}}
\def \cross{\color{red}{\ding{56}}}
\def \semitick{\textcolor{orange}{\ding{52}}{\small\textcolor{orange}{\kern-0.7em\ding{56}}}}

\setlength{\tabcolsep}{2.1pt}
\begin{tabular}{@{\extracolsep{2pt}}lcccccccc@{}} 
\toprule
& Closed-Loop & RGB & MV-Consistency & P-Realistic & FV-Actors & IA-Actors & Ext-Views & Real-Time \\
\cline{2-9}
\cellcolor{blue!25}MARS \cite{wu2023mars} & \cross & \tick & \tick & \tick & \cross & \cross & \cross & \cross \\
\cellcolor{blue!25}StreetGaussian \cite{yan2024street} & \cross & \tick & \tick & \tick & \cross & \cross & \cross & \tick \\
\cellcolor{yellow!25}KITTI \cite{geiger2012we} & \cross & \tick & \tick & \tick & \cross & \cross & \cross & \tick \\
\cellcolor{yellow!25}NAVSIM \cite{dauner2024navsim} & \semitick & \tick & \tick & \tick & \cross & \cross & \cross & \tick \\
\cellcolor{green!25}nuplan \cite{caesar2021nuplan} & \tick & \cross & \cross & \cross & \cross & \tick & \cross & \tick \\
\cellcolor{green!25}Carla \cite{dosovitskiy2017carla} & \tick & \tick & \tick & \cross & \tick & \tick & \tick & \tick \\
\cellcolor{green!25}NeuroNCAP \cite{ljungbergh2024neuroncap} & \tick & \tick & \tick & \tick & \cross & \cross & \cross & \cross \\
\cellcolor{green!25}DriveArena \cite{yang2024drivearena} & \tick & \tick & \cross & \tick & \tick & \tick & \tick & \cross \\
\cellcolor{green!25}HUGSIM (Ours) & \tick & \tick & \tick & \tick & \tick & \tick & \tick & \tick \\

\bottomrule
\end{tabular}
\vspace{-0.2cm}
\caption{\textbf{Comparison of Related Work,} including autonomous driving-related {\colorbox{blue!25}{reconstruction methods}}, {\colorbox{yellow!25}{open-loop benchmarks}}, and {\colorbox{green!25}{closed-loop benchmarks}}.
We compare various features like closed-loop evaluation (Closed-Loop), RGB images (RGB), multi-view image consistency (MV-Consistency), photo-realistic rendering (P-Realistic), full-view actors (FV-Actors), interactive actors (IA-Actors), fidelity of extrapolated views (Ext-Views), and real-time simulation (Real-Time) capabilities. Note that only one representative work is shown for overlapping features.
}
\vspace{-0.5cm}
\label{tab:relatedwork}
\end{table*}

%% file: figures/method/pipeline/pipeline.tex
\begin{figure*}
    \centering
    \includegraphics[width=\textwidth]{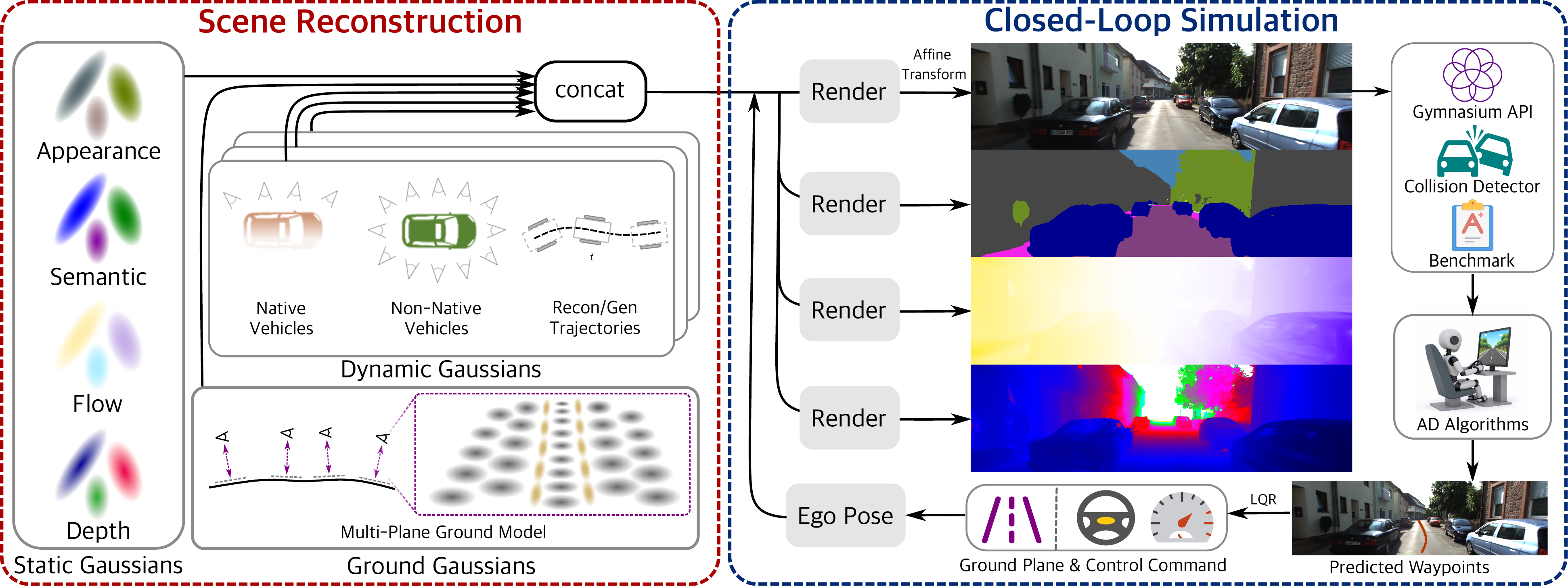}
    \vspace{-0.4cm}
    \caption{\textbf{The Pipeline of HUGSIM.} We reconstruct urban scenes by disentangling them as ground, non-ground static background,  and dynamic objects, while also enabling multi-modal rendering. Subsequently, we implement a closed-loop simulator based on the reconstructed results, providing a benchmark for evaluating autonomous driving algorithms.}
\vspace{-0.5cm}
\label{fig:pipeline}
\end{figure*}

%% file: sec_render.tex
\section{Urban Scene Reconstruction}\label{sec:method_render}

In this section, we begin by introducing the preliminaries. We then present our decomposed urban scene representation, designed to address rendering challenges in vision-based AD simulation. Next, we describe our holistic urban Gaussian splatting approach, covering appearance, semantics, optical flow, and depth. Finally, we detail the loss functions employed in our scene reconstruction.

\subsection{Preliminaries}

\boldparagraph{3D Gaussian Splatting} 3DGS \cite{kerbl20233d} represents scenes using a set of anisotropic 3D Gaussians with differentiable properties, allowing for efficient image rendering through tile-based rasterization. Each 3D Gaussian is defined by its position $\mu$, rotation $R$ (also represented as quaternion $\bq$ during training), scale $S$ (comprising $s_x, s_y, s_z$), opacity $\alpha$, and spherical harmonic coefficient $SH$. The 3D covariance matrix $\bSigma\in \nR^{3 \times 3}$ of each Gaussian is defined as:
\begin{equation}
\bSigma = RSS^TR^T
\end{equation}
A 3D Gaussian is defined as follow:
\begin{equation}
G(\bx ) =  \alpha \exp \left(-\frac{1}{2} (\bx-\mu)^T \bSigma^{-1} (\bx-\mu) \right)
\end{equation}

The color $\bc$ of each Gaussian can be computed based on the view direction and its corresponding spherical harmonics ($SH$). Given a set of sorted 3D Gaussians $\cN$ along the ray, we obtain the accumulated color via volume rendering:
\begin{equation}
\pi: \quad \bC = \sum_{i \in \mathcal{N}} \bc_i \alpha'_i \prod_{j=1}^{i-1}(1-\alpha'_j)
\label{eq:alpha_blending}
\end{equation}
Here $\alpha'$ is determined by the projected 2D Gaussian and the 3D opacity $\alpha$ following \cite{zwicker2002ewa}, see the appendix for details.

\boldparagraph{Coordinate System Definition} 
Given a sequence of urban images, we define the world coordinates based on the first frame throughout this paper. Specifically, the origin is set at the front camera of the first frame.
The coordinate axes are aligned with OpenCV conventions \cite{bradski2000opencv}, with the $x$ axis pointing to the right, the $y$ axis pointing downward, and the $z$ axis pointing forward.

\subsection{Decomposed Scene Representation}
\input{figures/method/lane_distortion/lane_distortion}
\input{figures/method/ground_model/ground_model}
We assume the scene consists of static regions and dynamic vehicles exhibiting rigid motion. Static regions are decomposed into ground and non-ground regions, allowing the application of planar constraints to the ground to preserve lane structure in extrapolated views. We consider two categories of dynamic vehicles: native dynamic vehicles present in the original driving dataset and non-native dynamic vehicles reconstructed from 360-degree captured images.

\boldparagraph{Non-Ground Static Gaussians}
Following 3DGS \cite{kerbl20233d}, we model all regions of urban scenes using 3D Gaussians. 
In addition to the original definition of 3D Gaussians, We propose to additionally model semantic logits $\bs \in \nR^{S}$ of each 3D Gaussian, allowing for rendering 2D semantic labels. Furthermore, we can naturally obtain a rendered optical flow $\bff_{t_1 \rightarrow t_2 }\in \nR^2$ for each 3D Gaussian by projecting the 3D position $\mu$ to the image space at two different timestamps, $t_1$ and $t_2$, and calculating the motion. We provide details of the multi-modality rendering in \secref{sec:holistic}.

\boldparagraph{Ground Gaussians}\label{sec:ground_model}
Lanes play a crucial role in the perception of AD algorithms. However, most existing reconstruction methods struggle to accurately render lane geometry in extrapolated views, as shown in \figref{fig:lanedistort}. The cause of these distortions is that ground Gaussians tend to overfit to the training views, failing to reconstruct the correct ground geometry.
Our preliminary experiments show that directly supervising the render depth fails to solve the problem, as Gaussians with incorrect geometry can still render accurate-looking 2D depth maps. Insteand, we regularize ground Gaussians to form planar structures, yielding correct geometry as shown in \figref{fig:groudmodel}.

A naive assumption would be to consider the ground of scenes as a single planar surface, allowing ground Gaussians to be distributed at the same height. However, this assumption overlooks more complex scenarios, such as sloped roads. To tackle this problem, 
we propose a multi-plane ground model, where we assume the ground to be planar only within a limited distance, denoted as $\Delta Z$. In this model, each local plane is assumed to have a fixed height relative to the nearest camera. Since the camera poses reflect the surface slopes, this multi-plane approach effectively models such complex scenarios.
Specifically, we optimize ground Gaussians and constrain the distribution of Gaussians in 3D space by limiting the height variance of sampled Gaussian patches within a small $\Delta Z$ in corresponding camera coordinates. 
Note that the local planes overlap with each other, hence avoiding boundary artifacts.
More formally, the constraints of our ground model can be expressed as the optimization target:
\begin{align}
    & \underset{\{\mu_{x, y, z}, s_{x, z}, \bc, \alpha\}}{\text{minimize}}
    & & (1-\lambda_{SSIM}) \| \hat{\bI} - \Tilde{\bI} \|_1 + \lambda_{SSIM}\text{SSIM}(\hat{\bI}, \Tilde{\bI}) \nonumber \\
    & \text{subject to}
    & & \lim_{\Delta Z \rightarrow 0} \sqrt{\frac{1}{N-1} \sum_{i=1}^{N} (\mu_{y_i}^{cam} - \bar{\mu}_y^{cam})^2} = 0 \label{eq:ground_model}
\end{align}
where $\hat{\bI}$, $\Tilde{\bI}$ represent the rendered image and the ground truth image, respectively. $N$ denotes the number of Gaussians in one local plane. $\mu_{y_i}^{cam}$ and $\bar{\mu}_y^{cam}$ represent heights of 3D Gaussians in camera coordinates and average heights in a patch.
Note that the $s_y$ and $\bq$ of the Gaussians are kept constant to ensure they remain flat and oriented upward.

Unlike previous approaches that use densely tiled or LiDAR-initilaized Gaussians, as seen in RoGS \cite{feng2024rogs} and AutoSplat \cite{khan2024autosplat}, we find that the ground can be efficiently represented using sparser distributed Gaussians since the ground textures are not uniformly distributed. We therefore retain color, position, opacity, two dimensions of scale as optimizable parameters, while also incorporating the density control strategy \cite{kerbl20233d}. Our approach enables high-quality ground rendering without the need for an excessive number of 3D Gaussians, as demonstrated in our experiments.

\input{figures/method/single/single}

\boldparagraph{Native Dynamic Vehicle Gaussians and Unicycle Model}\label{sec:unicycle}
For dynamic vehicles, we assume 3D bounding boxes are predicted from the input RGB images, enabling 3D Gaussian modeling in object coordinate space. To address noise in predictions, we jointly optimize them by regularizing with a unicycle model.
Specifically, we parameterize the transformations $(\bR_t, \bt_t)$ of each dynamic vehicle following the unicycle model\footnote{While it is more accurate to model vehicles using a bicycle model, we observe that using the simpler unicycle model is sufficient here.}. The state of a unicycle model is parameterized by three elements: $(x_t, z_t, \theta_t)$, where $x_t$ and $z_t$ represent horizontal coordinates of $\bt$ with $\bt_t = [x_t,y_t,z_t]$, and $\theta_t$ is the yaw angle of $\bR_t$.
To adapt the continuous unicycle model to discrete frames, we derive the calculus of the unicycle model for the vehicle transition from timestamp $t$ to $t+1$ as follows:
\begin{align}
    x_{t+1} & = x_t + \frac{v_t}{\omega_t} (\sin \theta_{t+1} - \sin \theta_t)  \nonumber \\ 
    z_{t+1} & = z_t - \frac{v_t}{\omega_t} (\cos \theta_{t+1} - \cos \theta_t)  \nonumber  \\ 
    \theta_{t+1} & = \theta_t + \omega_t \label{eq:unicycle}
\end{align}
Here, $v_t$ represents the forward velocity, and $\omega_t$ is the angular velocity. This model integrates physical constraints when compared to directly optimizing the transformations of dynamic vehicles at every frame independently, thus enabling smoother motion modeling of moving objects and making them less prone to local minima.

While it is possible to define an initial state $(x_1, z_1, \theta_1)$ and derive the following states recursively based on velocities, $v_t$ and $\omega_t$, such a recursive parameterization is challenging to optimize.
In practice, we define a set of trainable states $\{(x_t, z_t, \theta_t)\}_{t=1}^{T}$ along with trainable velocities $\{v_t, \omega_t\}_{t=1}^{T-1}$,
and add a regularization term to ensure that the vehicle's states adhere to the characteristics of a unicycle model in \eqref{eq:unicycle}. The regularization terms will be described in \secref{sec:loss}. Additionally, we retrieve the vertical locations of the vehicle $\{y_t\}_{t=1}^{T}$ from our multi-plane ground model. During simulation, we interpolate the states from \eqref{eq:unicycle} by given timestamps.

\boldparagraph{Non-Native Full-Observed Vehicle Gaussians}
The AD simulator requires rendering high-fidelity actors from all 360 degrees, particularly when integrating interactive actors into closed-loop simulations. However, vehicles in the original reconstructed scenes are only captured from a limited set of viewpoints, resulting in noticeable artifacts when viewed from angles far outside the training perspectives. To address this, we reconstruct vehicles using a densely captured real-world dataset, 3DRealCar~\cite{du20243drealcar}, which provides 360-degree observations of real-world vehicles. Our experiments show that real-world captured vehicles outperform vehicles from original scenes when inserted into the simulation scenes with random viewing angles.

The 3DRealCar dataset provides masks of the vehicles. We leverage the mask information to ensure that the 3D Gaussians only model the car foreground. This is achieved by considering an alpha mask loss in addition to the vanilla rendering loss.
Importantly, directly inserting the foreground vehicles without shadows often appear as if they are floating in the air. However, inverse rendering requires an accurate environment map, which is difficult to obtain from perspective cameras. Although some work \cite{moenne20243d, shi2023gir, liang2024gs, 3dgrt2024} has addressed the challenge of inverse rendering in Gaussian Splatting, it remains a computationally expensive operation. To simplify the problem, we assume the light source (the sun) is directly overhead, meaning shadows should appear beneath the vehicles. To render vehicle shadows, we place flat Gaussians at the bottom of the vehicle in canonical space, as shown in \figref{fig:single}. The $\alpha$ attribute of these Gaussians decreases smoothly based on their distance from the bottom center. Although this is a simplified assumption, we observe that the inserted non-native vehicles appear plausible in many scenarios, striking a good balance between efficiency and photorealism.

\input{figures/method/smt_softmax/softmax}

\subsection{Holistic Urban Gaussian Splatting}
\label{sec:holistic}
Given the representation specified above, we are able to render images, semantic maps and optical flow to supervise the model or make predictions at inference time. We now elaborate on the rendering of each modality.

\boldparagraph{Novel View Synthesis}
The combination of static and dynamic Gaussians can be sorted and projected onto the image plane via $\alpha$-blending $\pi$ in \eqref{eq:alpha_blending}.

In contrast to single-object scenes, urban scenes typically involve more complex lighting conditions and the images are usually captured with auto white balance and auto exposure. NeRF-based methods \cite{martin2021nerf} typically feed a per-frame appearance embedding along with the 3D positions into a neural network to compute the color, thereby compensating exposure. However, when working with 3D Gaussians, there is no neural network capable of processing appearance embeddings.  Inspired by Urban Radiance Field \cite{rematas2022urban}, we generate an exposure affine matrix for each camera by mapping the camera's extrinsic parameters to an affine matrix $\bA \in \nR^{3 \times3}$ and vector $\bb \in \nR^3$ via a small MLP:
\begin{equation}
\Tilde{\bC} = \bA \times \bC + \bb
\end{equation}
We demonstrate that modeling the exposure improves rendering quality in the experimental section.

\boldparagraph{Semantic Reconstruction}
Similarly to \eqref{eq:alpha_blending}, we can obtain 2D semantic labels via $\alpha$-blending based on the 3D semantic logit $\bs$:
\begin{equation}
\pi_\bS: \quad \bS = \sum_{i \in \mathcal{N}} \text{softmax}(\bs_i) \alpha'_i \prod_{j=1}^{i-1}(1-\alpha'_j)
\label{eq:semantic_render}
\end{equation}
Note that we perform the softmax operation on 3D semantic logits $\bs_i$ prior to $\alpha$ blending, in contrast to most existing methods that apply softmax to 2D semantic logits $\bar{\bS}$ obtained by accumulating unnormalized 3D semantic logits $\bs_i$~\cite{zhi2021place, fu2022panoptic, kundu2022panoptic}. As shown in \figref{fig:softmax_cmp}, applying softmax in 2D space leads to noisy 3D semantic labels. This is due to the fact that 2D space softmax can produce accurate 2D semantics by adjusting the scale of the 3D semantic logits,  allowing a single sampled point with a substantial logit value to significantly influence the volume rendering outcome.
For example, an undesired floating point labeled with ``car'' may not be penalized despite the target rendered label being ``tree'', as long as a 3D Gaussian is providing a large logit value of ``tree'' along this ray. Our solution instead removes such floaters by normalizing logits in 3D space. See the appendix for more quantitative and qualitative details.
The removal of these floaters is also essential for building our simulator, as the background collision detection is implemented based on Gaussian semantics and geometry.

\boldparagraph{Optical Flow} The 3D Gaussian representation also enables the rendering of optical flow. 
Given two timestamps $t_1$ and $t_2$, we first calculate the optical flow of each 3D Gaussian's center $\mu$ as $\bff_{t_1\rightarrow t_2}$. Specifically, we project $\mu$ to the 2D image space based on the camera's intrinsic and extrinsic parameters: 
\begin{equation}
    \mu'_1=\bK{[\bR_{t_1}^{\text{cam}};\bt_{t_1}^{\text{cam}}]}\mu, \quad 
    \mu'_2=\bK{[\bR_{t_2}^{\text{cam}};\bt_{t_2}^{\text{cam}}]}\mu,
\end{equation}
and then calculate the motion vector as $\bff_{t_1\rightarrow t_2}=\mu'_2-\mu'_1$.
Next, we render the optical flow via accumulate the optical flows via volume rendering:
\begin{equation}
\pi_{\bF}: \quad \bF = \sum_{i \in \mathcal{N}} \bff_i \alpha'_i \prod_{j=1}^{i-1}(1-\alpha'_j)
\label{eq:flow_render}
\end{equation}
Note that this rendering process assumes that any pixel of a rendered 2D Gaussian splat shares the same optical flow direction as the corresponding Gaussian center but with scaled magnitude. While this is indeed a simplified approximation, we observe this to work well in practice.

\boldparagraph{Depth}
Depth maps are also provided during rendering. Each Gaussian's depth distance relative to the observed camera is defined as $\bd_i$. The depth map is rendered by accumulating $\bd_i$ using volume rendering:
\begin{equation}
\pi_{\bD}: \quad \bD = \sum_{i \in \mathcal{N}} \bd_i \alpha'_i \prod_{j=1}^{i-1}(1-\alpha'_j)
\end{equation}

\subsection{Loss Functions} \label{sec:loss}
Our loss functions are divided into two parts. 
The first is image-based losses, which aim to fit the observed ground truth or pseudo-ground truth 2D images. The second is geometry-based losses, which constrain the reconstructed result to align with physical priors from the real world. We leverage the pre-trained perception model \cite{borse2021inverseform} to provide noisy 2D semantic. To compensate for the absence of manually labeled 3D bounding boxes, we use pre-trained recognition models \cite{hu2022monocular} to provide noisy 3D tracking results. These easy-to-obtain predictions are crucial for enabling RGB-only holistic scene understanding in both 2D and 3D space, without relying on LiDAR input or 3D semantic supervision.

\boldparagraph{Image-Based Losses}
Our model is supervised with the ground truth images using a combination of L1 and SSIM losses. Our rendering loss is defined as follows:
\begin{equation}
    \cL_{\bI} = (1-\lambda_{SSIM}) \| \hat{\bI} - \Tilde{\bI} \|_1 + \lambda_{SSIM}\text{SSIM}(\hat{\bI}, \Tilde{\bI})
\end{equation}
where $\hat{\bI}$ and $\Tilde{\bI}$ represent the rendered and the ground truth image, respectively.
We additionally apply the cross-entropy loss to the rendered semantic label \wrt pseudo-2D semantic segmentation ground truth $\hat{\bS}$:
\begin{equation}
    \cL_{\bS} = - \sum_{k=0}^{S-1} \hat{\bS}_k \log (\bS_k) 
\end{equation}
During the reconstruction of the non-native vehicles, we apply an alpha loss to prevent Gaussians from representing the masked background and to ensure the reconstructed vehicle appears as a non-translucent entity:
\begin{equation}
    \cL_{\mathcal{A}} = \| \mathcal{A} - \Tilde{\bI}_{\cM} \|_2
\end{equation}
where $\Tilde{\bI}_{\cM}$ denotes the ground truth mask and $\mathcal{A}$ the rendered alpha map.

\boldparagraph{Physical-Based Regularizations}
A 3D regularization loss is applied to the multi-plane ground model to limit the variance in the height of the Gaussians, forcing the model learns an almost flat ground patch. We express the optimization problem \eqref{eq:ground_model} as the multi-plane ground model regularization loss during training:
\begin{equation}
    \cL_{ground} = \frac{1}{N-1} \sum_{z_i - z_0 < \Delta z} (\mu_{y_i}^{cam} - {\bar{\mu}_y^{cam}})^2
\end{equation}

\input{figures/method/attack/attack}

We use a unicycle model to guide the noisy 3D bounding box predictions:
\begin{equation}
\label{eq:L_3dbox}
\cL_{\bt} = \sum_t \Vert x_t - \hat{x}_t \Vert_2 + \sum_t \Vert z_t - \hat{z}_t\Vert_2
\end{equation}
where $\hat{x}_t$ and $\hat{y}_t$ are the $x$ and $y$ locations of a noisy 3D bounding box at timestamp $t$.

As mentioned earlier, we parameterize the vehicle's states $(x_t, y_t, \theta_t)$ and the velocities $v_t, \omega_t$ as learnable parameters. Hence, we add the following regularization to make the states adhere to the unicycle model as follows:
\begin{align}
\cL_{uni} = & \sum_t \| x_{t+1} - x_t - \frac{v_t}{\omega_t} (\sin \theta_{t+1} - \sin \theta_t) \| + \nonumber \\
           &   \sum_t\|  z_{t+1} - z_t + \frac{v_t}{\omega_t} (\cos \theta_{t+1} - \cos \theta_t) \| +    \nonumber\\
           &  \sum_t \|  \theta_{t+1} - \theta_t  - \omega_t \| 
            \label{eq:uni_restrain}
\end{align}
In addition, we regularize the acceleration of the forward velocity $v_t$ and angular velocity $\omega_t$ to be smooth:
\begin{align}
\cL_{reg} = & \sum_t \Vert v_{t+1} + v_{t-1} - 2v_t \Vert_2 + \nonumber \\
            & \sum_t \Vert \theta_{t+1} + \theta_{t-1} - 2\theta_t \Vert_2
            \label{eq:uni_reg}
\end{align}

%% file: figures/method/lane_distortion/lane_distortion.tex
\begin{figure}
     \centering
     \small 
     \setlength{\tabcolsep}{0pt}
     \def\mywidth{4.3cm}
     \begin{tabular}{P{\mywidth}P{\mywidth}}

     \includegraphics[width=\mywidth]{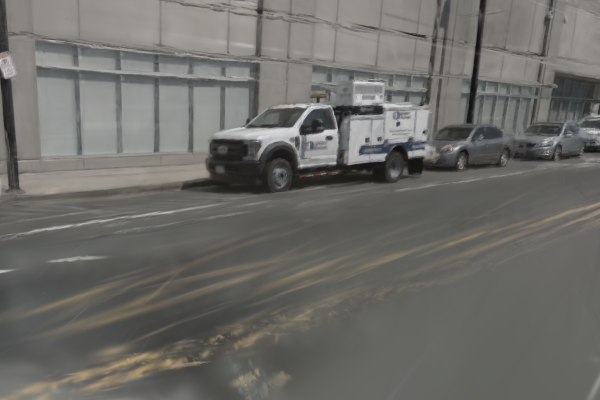}  &
     \includegraphics[width=\mywidth]{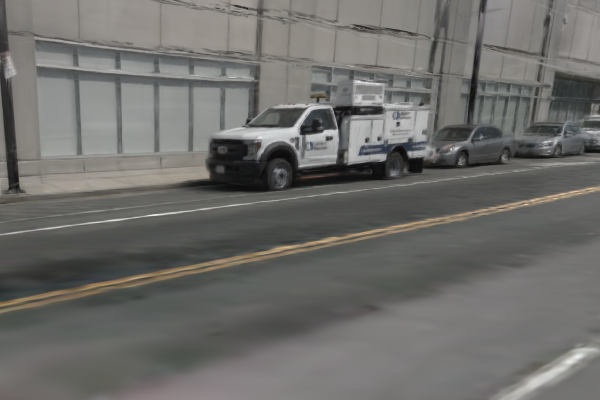} \\

     w/o. Ground Model & w. Ground Model \\
     
     \end{tabular}
     \vspace{-0.2cm}
     \caption{\textbf{Lane Distortion} In the left image of the extrapolated view, the lane appears highly distorted due to incorrect geometry of ground gaussians. In contrast, the right image, which applies the multi-plane ground model, shows significantly fewer artifacts in the same view.}

\vspace{-0.3cm}
\label{fig:lanedistort}
\end{figure}

%% file: figures/method/ground_model/ground_model.tex
\begin{figure}
    \def\mywidth{8.8cm}
    \centering
    \includegraphics[width=\mywidth]{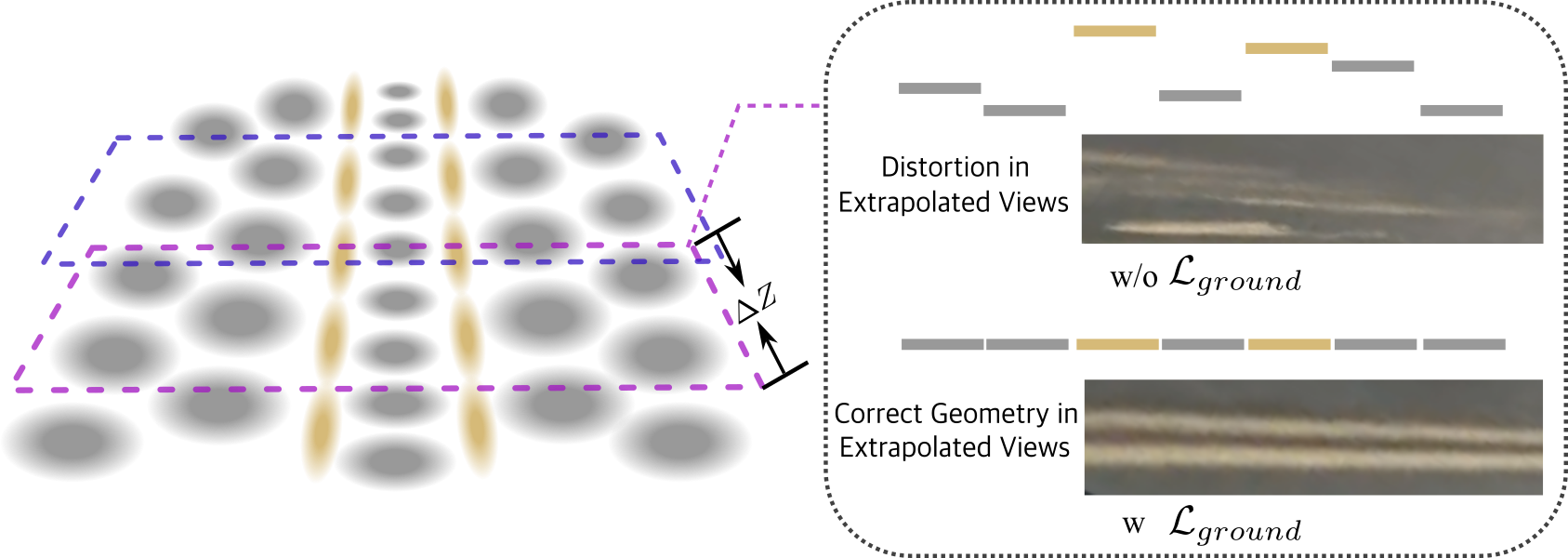}
    \vspace{-0.2cm}
    \caption{\textbf{The Multi-Plane Ground Model.} Our ground model is initialized by multiple planes, each containing a set of flat 3D Gaussians. During training, we sample small patches of these Gaussians and constrain the height variance.
    }
\label{fig:groudmodel}
\vspace{-0.5cm}
\end{figure} 

%% file: figures/method/single/single.tex
\begin{figure}
     \centering
     \small 
     \setlength{\tabcolsep}{0pt}
     \def\mywidth{4.5cm}
     \begin{tabular}{P{\mywidth}P{\mywidth}}

     \includegraphics[width=\mywidth]{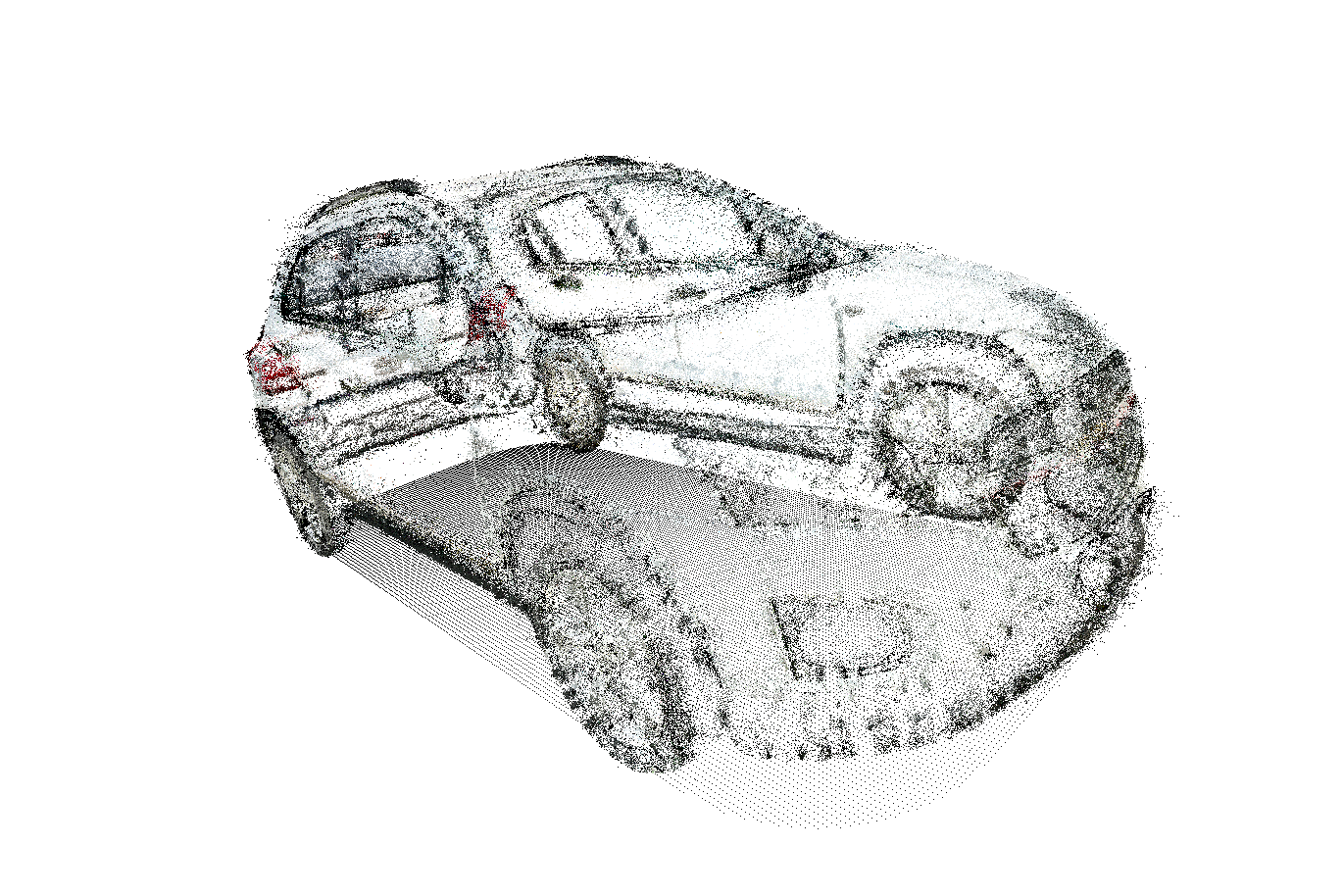}  &
     \includegraphics[width=\mywidth]{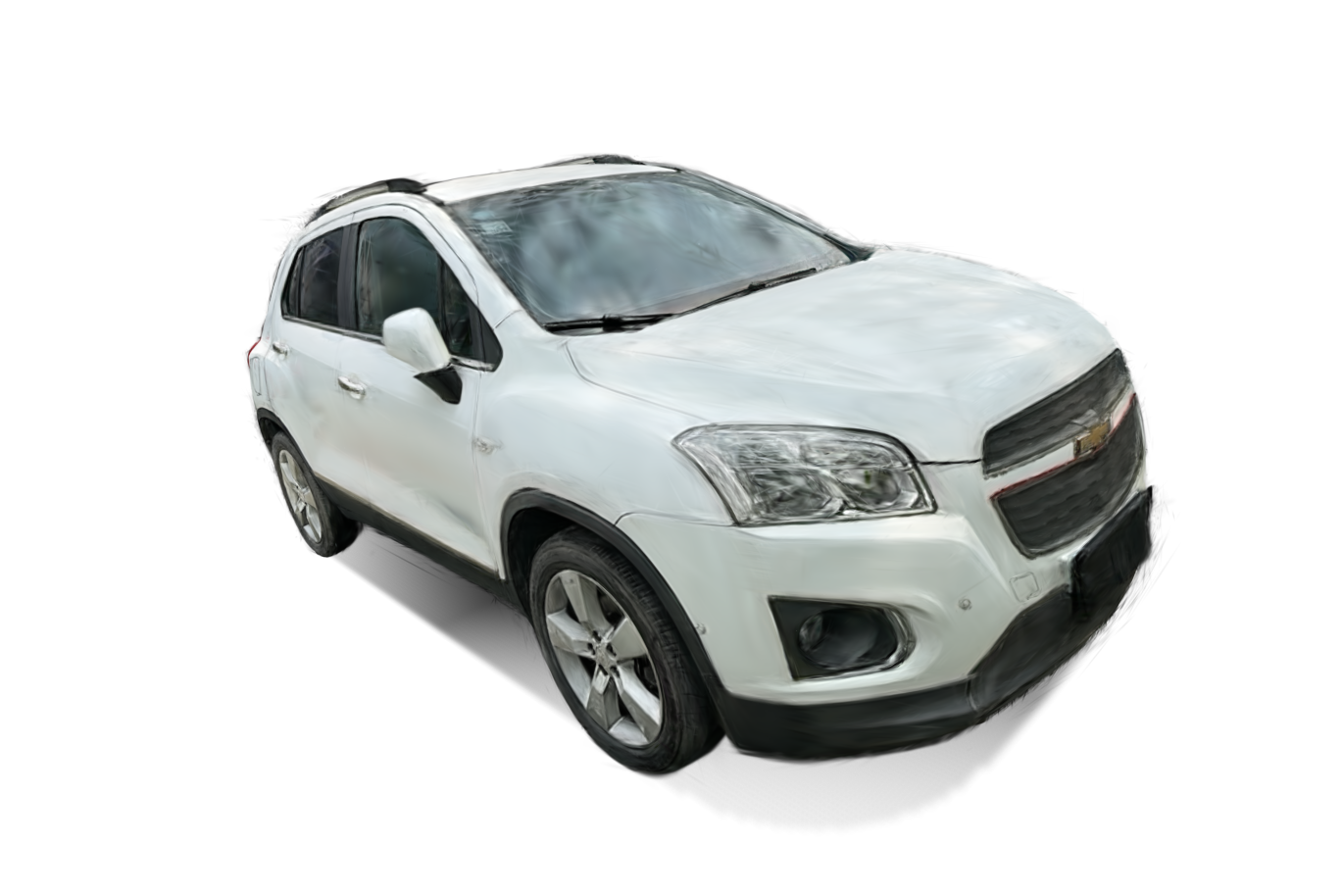} \\

     \includegraphics[width=\mywidth]{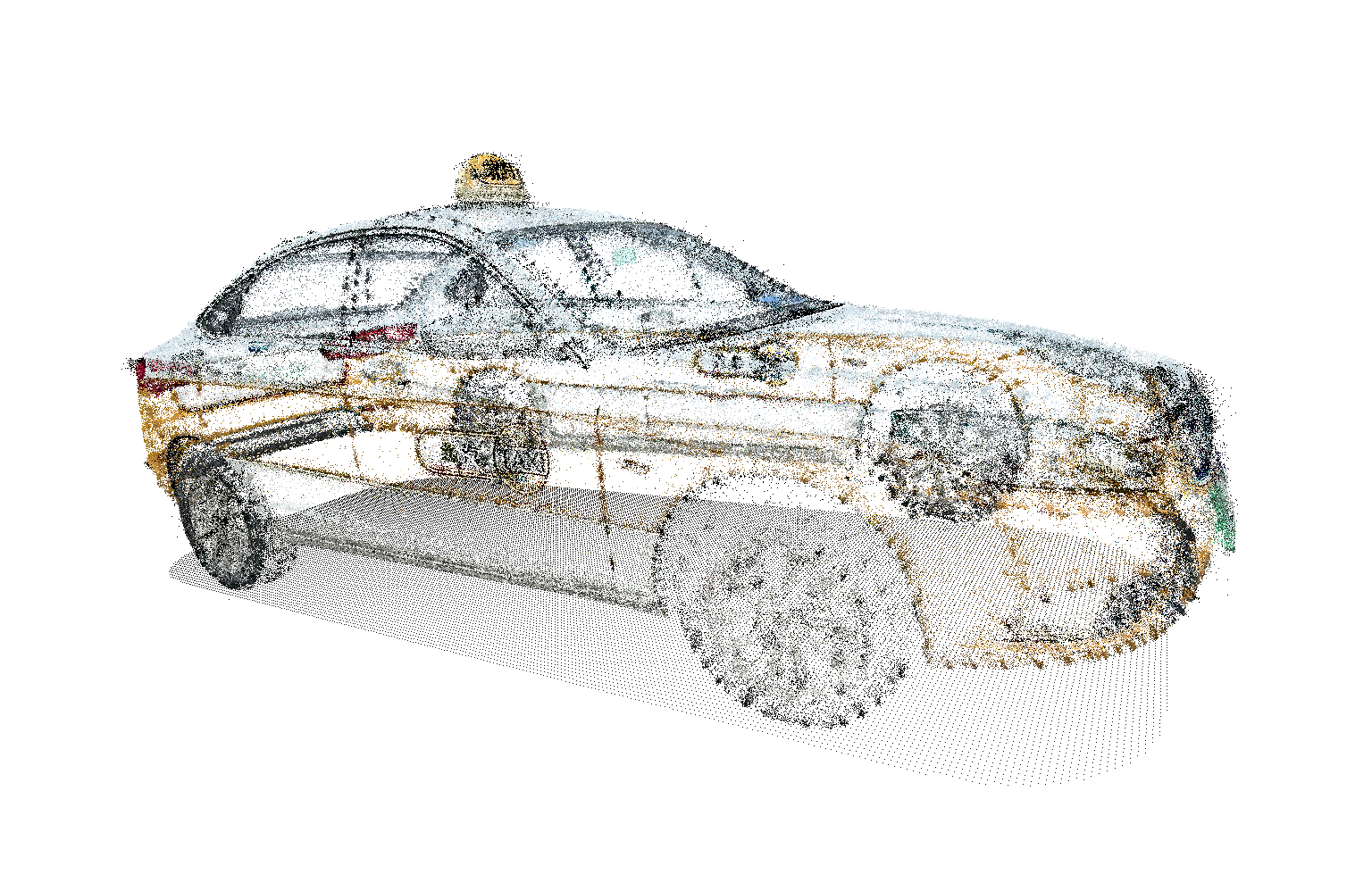}  &
     \includegraphics[width=\mywidth]{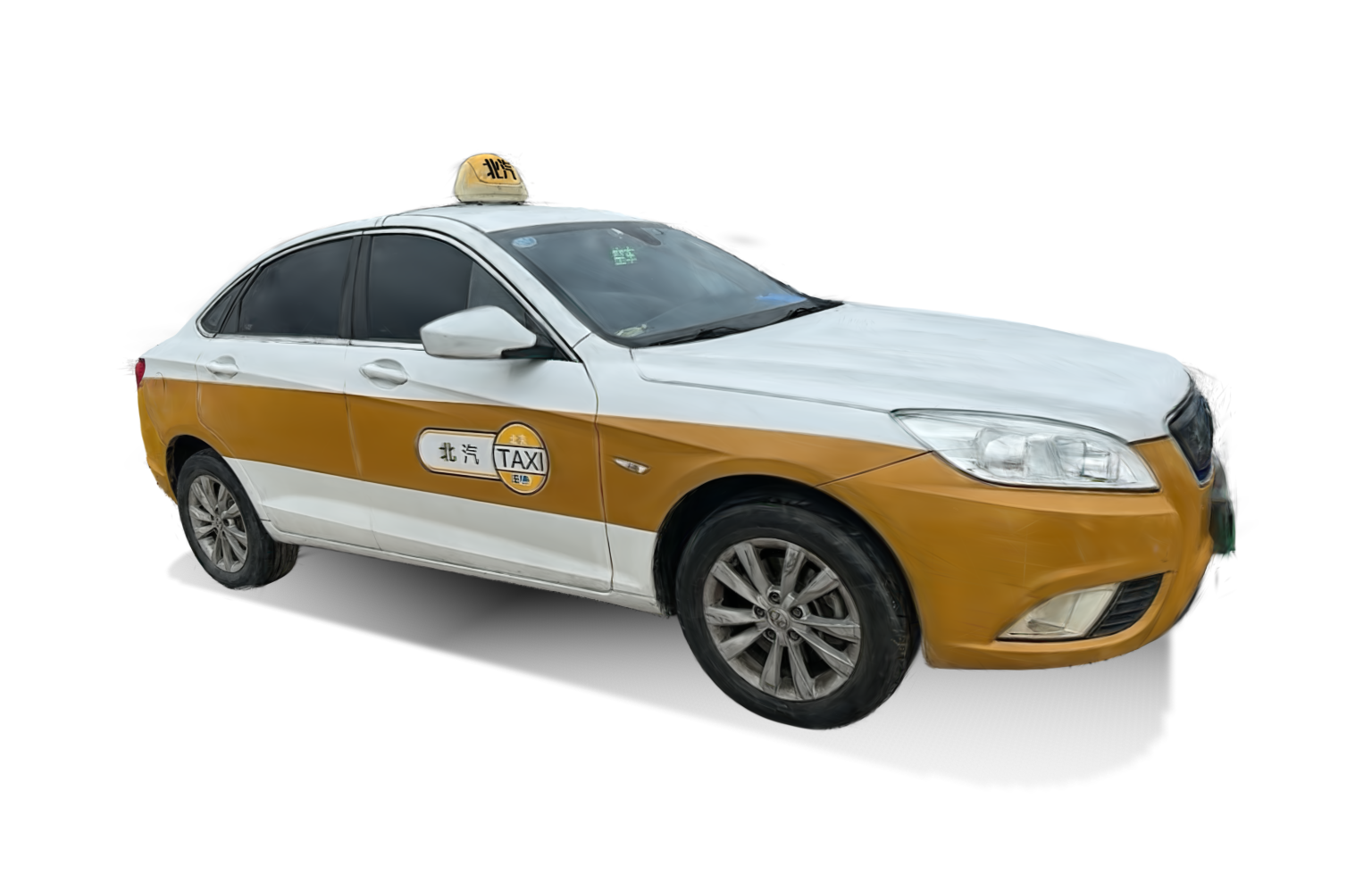} \\
     
     \end{tabular}
     \vspace{-0.2cm}
     \caption{\textbf{Non-Native Vehicle Reconstruction} We reconstruct more than 100 vehicles from 3DRealCar \cite{du20243drealcar} to enable 360-degree high-fidelity actor observation. Additionally, we place shadow Gaussians with gradually changing opacity inside a rounded rectangle to enhance the visual realism of the inserted actor. }
     \vspace{-0.4cm}
\label{fig:single}
\end{figure}

%% file: figures/method/smt_softmax/softmax.tex
\begin{figure}
     \centering
     \small 
     \setlength{\tabcolsep}{0pt}
     \def\mywidth{4.4cm}
     \begin{tabular}{P{\mywidth}P{\mywidth}}

     \includegraphics[width=\mywidth]{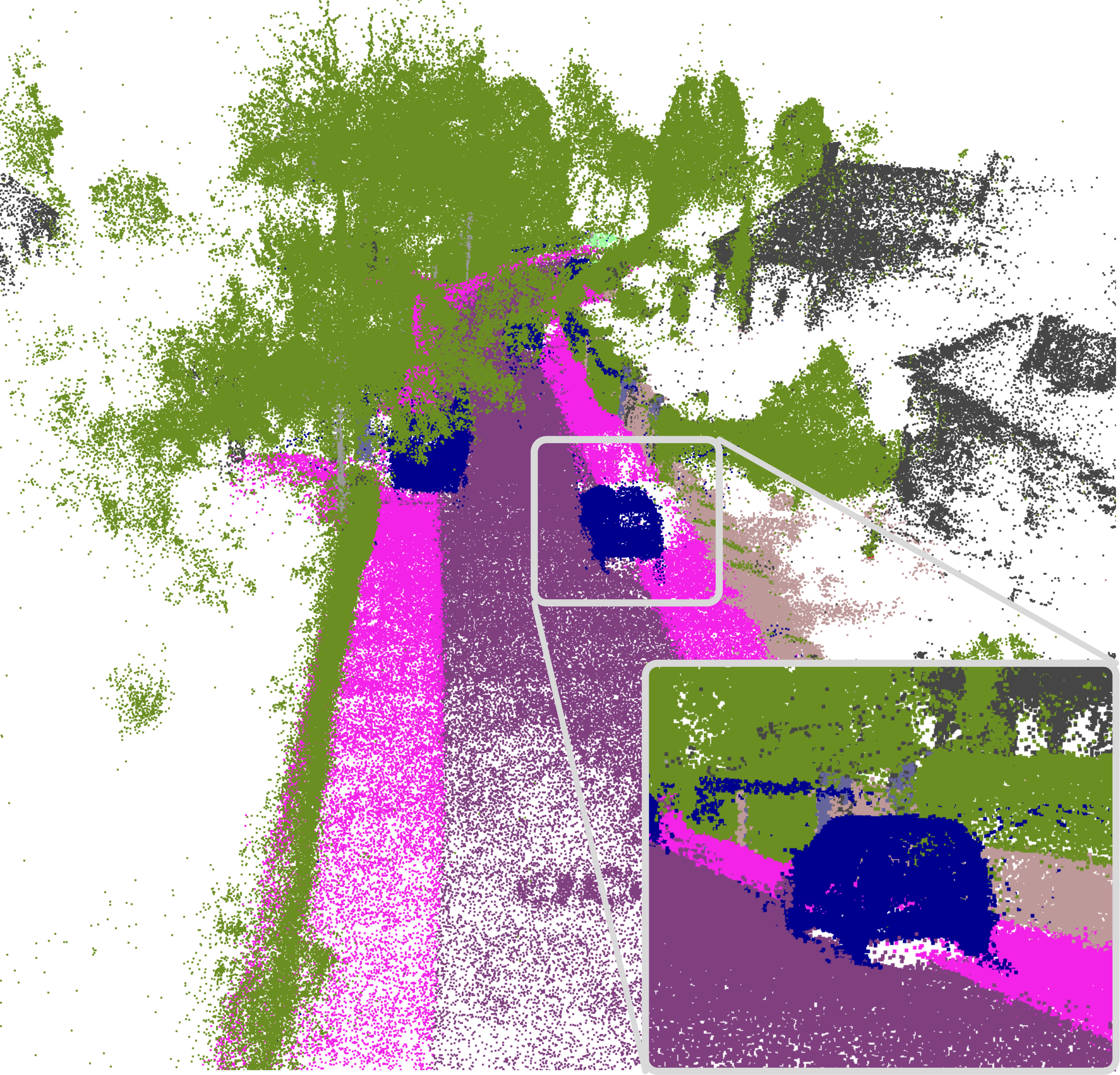}  &
     \includegraphics[width=\mywidth]{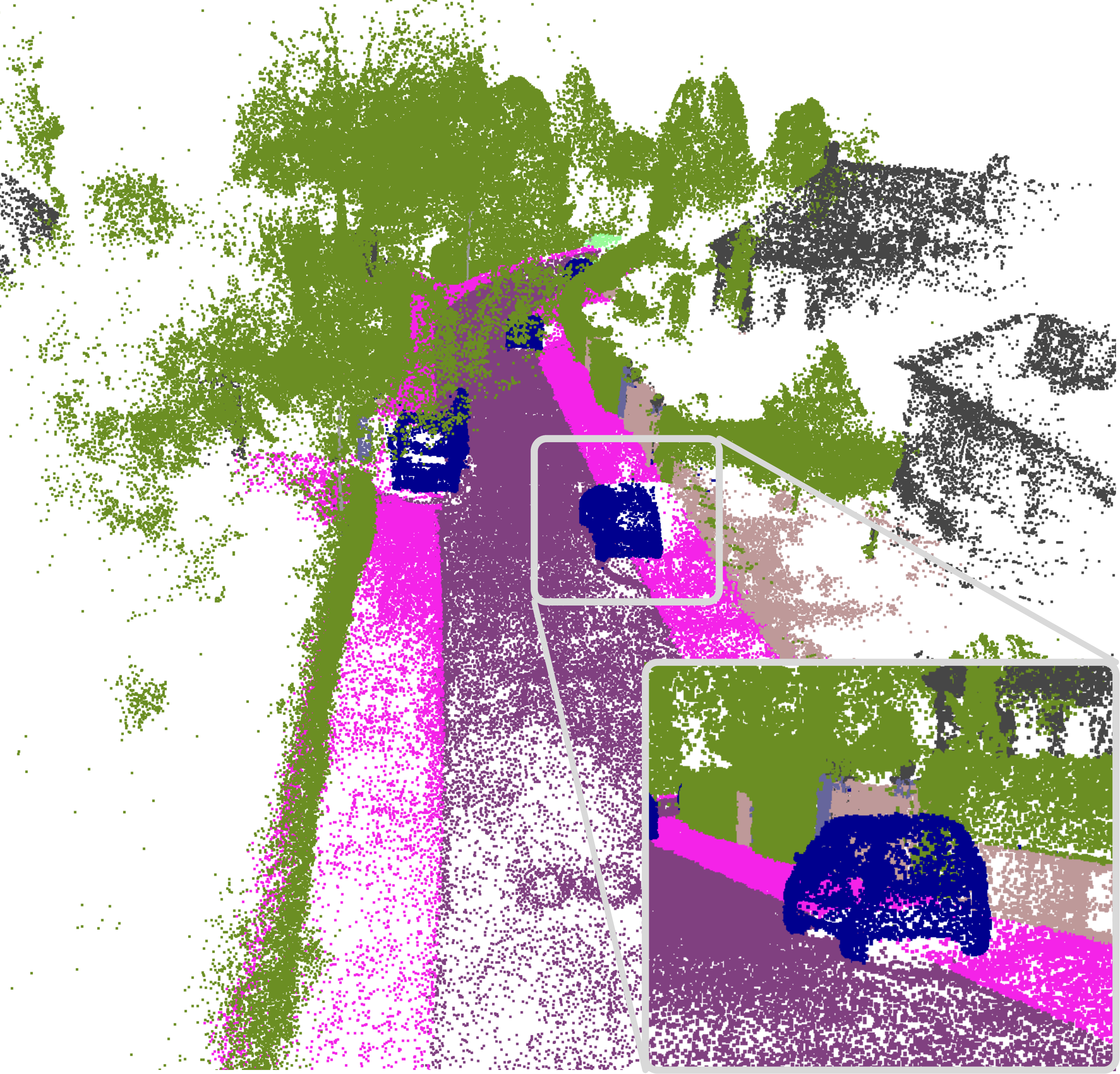} \\

     \end{tabular}
     \caption{\textbf{3D Semantic Reconstruction.} Comparison between applying softmax to accumulated 2D semantic logits (left) and to 3D semantic logits (right). Normalizing semantic logits in 3D space clearly reduces floaters and yields better 3D semantic reconstruction than the 2D normalization counterpart. This improvement is also crucial for our simulator collision detection.}
     \vspace{-0.3cm}
\label{fig:softmax_cmp}
\end{figure}

%% file: figures/method/attack/attack.tex
\begin{figure*}
    \centering
    \includegraphics[width=\textwidth]{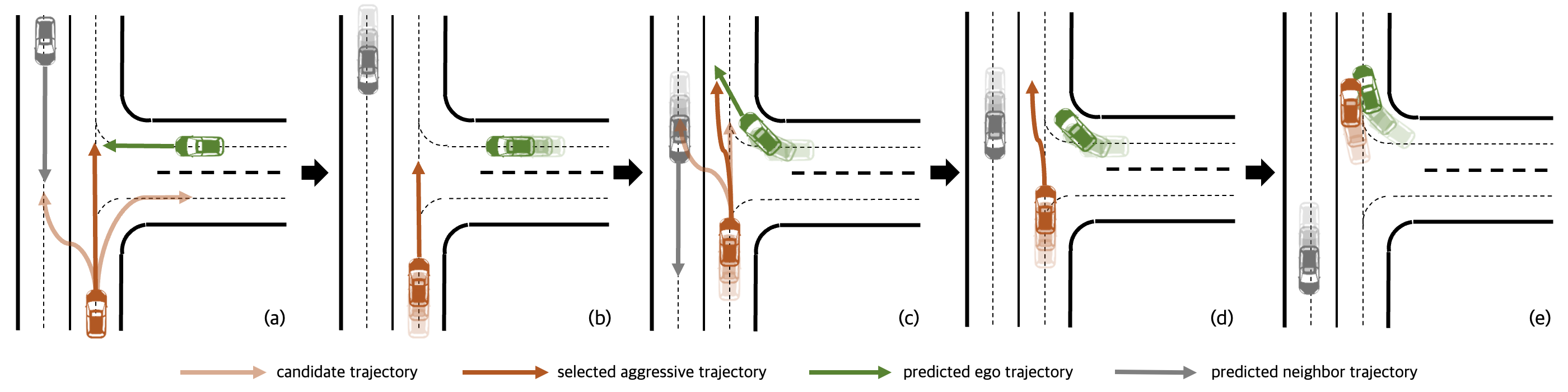}
    \vspace{-0.3cm}
    \caption{\textbf{The Aggressive Driving Behavior Model.} From (a) to (e), we illustrate the behavior of an aggressive actor (orange) attempting to collide the ego-vehicle (green) over a time period.
    In (a), the attacker selects the trajectory with the highest probability of collision by estimating the future trajectory of the ego vehicle based on its current state. In (c), the attacker re-estimates and updates the trajectory, successfully colliding with the ego vehicle in (e).}
\vspace{-0.2cm}
\label{fig:attack_model}
\end{figure*}

%% file: sec_simulation.tex
\section{Simulation}\label{sec:method_simulation}
In this section, we first elaborate on the construction of a closed-loop simulator based on the reconstruction of the previous section. Next, we present our method for efficiently generating actor behaviors, including both normal and aggressive driving patterns. Finally, we introduce the evaluation metrics used to assess AD algorithms within our simulator.

\subsection{Graphicial Configuration Interface}
We develop a graphicial user interface (GUI) to facilitate the configuration of testing scenarios in our simulator. The GUI configuration involves several steps. The first step is configuring the camera settings, including the number of cameras, camera intrinsics, and and the extrinsic parameters wrt. the vehicle. 
The second step is configuring the ego vehicle parameters, including specifying the kinematic model, the control frequency, and the start state of the ego vehicle. The final step involves configuring the actors, including native and non-native vehicles with different behaviors specified in \secref{sec:actor_behavior}. The appearance of all these actors can be selected from a pool of more than 100 candidate 3D vehicles reconstructed from 3DRealCar \cite{du20243drealcar}.

\subsection{Closed-Loop Simulation}
\boldparagraph{Simulator-User communication}
Given the 3D Gaussian reconstruction results, we encapsulate the scenes as Gymnasium environments \cite{brockman2016openai}, 
providing functions such as environment resetting, retrieving camera observations, accessing information about the ego vehicle and environment, and advancing to the next state with specified control actions for the ego vehicle.
Ideally, the simulator should support evaluating various AD algorithms without requiring code adaptation. To achieve this, we maintain separate repositories for the simulator and AD algorithms, implementing them independently while enabling parallel execution and communication. 
When both the simulator and AD algorithms are run on the same machine, we use named pipes to establish a communication bridge, minimizing communication costs. In other cases, web sockets are available for communication.

\boldparagraph{Controller}
By providing observations and ego vehicle information to the AD algorithms, our simulator expects feedback in the form of either planned waypoints or a sequence of control commands for the next several seconds. When the feedback is planning waypoints, these waypoints need to be converted into a sequence of control commands, including steering angle ($\delta$) and acceleration ($\dot{a}$). To achieve this conversion, we leverage a Linear Quadratic Regulator (LQR) control following \cite{karnchanachari2024towards, ljungbergh2024neuroncap, yang2024drivearena}. 

\boldparagraph{Ego-Vehicle Kinematic Model}
Given the control commands of steering angle ($\delta$) and acceleration ($\dot{a}$), the next states $\cS$ of the ego vehicle can be computed by applying a discrete version of the kinematic bicycle model, following \cite{karnchanachari2024towards, samak2021control}.
In this model, $\bL$ denotes the length of the ego vehicle:
\begin{equation}
    \cS = 
    \left(
    \begin{matrix}
        x \\ y \\ \theta \\ v
    \end{matrix}
    \right), 
    \frac{d\cS}{dt} = 
    \left(
    \begin{matrix}
        v\cos{\theta} \\
        v\sin{\theta} \\
        \frac{v\tan{\delta}}{\bL} \\
        \dot{a}
    \end{matrix}
    \right)
\end{equation}

\boldparagraph{Collision Detection}
We define two types of collisions for the ego vehicle: collisions with the foreground (actors) and collisions with the background. Collisions with the foreground are detected by checking if the Bird's Eye View (BEV) bounding boxes of the ego vehicle and actors overlap. Collisions with the background are identified by counting the number of background Gaussians inside the ego vehicle's 3D bounding box, excluding Gaussians with ground semantic labels and low opacities. If the number exceeds a preset threshold, a collision is considered to have occurred.

\subsection{Actor Driving Behaviors}
\label{sec:actor_behavior}
We design three patterns of driving behavior: replayed, normal and aggressive. 

\boldparagraph{Replayed Driving Behavior}
For native dynamic vehicles in scenes, we reconstruct distinct actor pose observations into a continuous trajectory using our unicycle models. The reconstructed trajectory can be integrated into the HUGSIM closed-loop simulation by providing corresponding timestamps. It is important to note that this replayed driving behavior does not interact with the ego vehicle and is only applied in some easy scenarios.

\boldparagraph{Normal Driving Behavior}
Normal driving behavior follows IDM \cite{treiber2000congested}, a simple car-following model that relies on HD maps for lane tracking and vehicle following. If no vehicle is present in front, the model drives at a constant speed.
Among the datasets we consider in this work, NuScenes is the only one that provides paired RGB images and HD maps. Therefore, we apply the IDM-based normal driving behavior on nuScenes alone. For other datasets, we use a constant speed following a predefined driving direction as a simple alternative.

\boldparagraph{Aggressive Driving Behavior}
The design of aggressive driving behavior is illustrated in \figref{fig:attack_model}, this adversarial model performs effectively both with and without the use of HD maps.
Inspired by KING \cite{Hanselmann2022ECCV} and CAT \cite{zhang2023cat}, we design an optimization-based generative trajectory planner. Unlike KING, we generate adversarial behavior during simulation rather than in a post-processing manner. In contrast to CAT, our approach can generate candidate trajectories both with and without HD maps.
First, based on road information or current actor states, we generate a set of feasible candidate trajectories $\{s_{1:T}^{a(i)}\}_{i=1}^{N}$ based on a spline-planner for the attacking actor. $T$ is the planning horizon and $N$ is the number of candidate trajectories $s_{1:T}^{a(i)}$.
The spline-planner generates feasible trajectories through grid-based terminal state sampling, spline-based trajectory interpolation, and feasibility filtering using dynamic constraints.
This process enables the generation of diverse and robust trajectory sets while ensuring adherence to physical and environmental constraints.
Next, we select from the feasible trajectories with the goal of attacking the ego vehicle while avoiding collision with other actors, achieved by predicting the future trajectory of the ego vehicle and other actors. More formally, we select the aggressive actor trajectory by minimizing the composite cost:
\begin{align}
    & \min_{i} C_{total}(s_{1:T}^{a(i)}) = C_{attack}(s_{1:T}^{a(i)}) + \lambda C_{collision}(s_{1:T}^{a(i)}), \nonumber \\
    & C_{attack}(s_{1:T}^{a(i)}) = \min_{t=1:T} \left \| s_{t}^{e} - s_{t}^{a(i)} \right \|, \nonumber \\
    & C_{collision}(s_{1:T}^{a(i)}) = \sum_{j=1}^{M}\mathds{1}(\min_{t=1:T} \left \| s_{t}^{n(j)} - s_{t}^{a(i)} \right \| < tolerance).
\end{align}
where $s_{1:T}^{e}$ denotes the ego future trajectory and $s_{1:T}^{n(j)}$ denote the trajectories of other inserted actors, both predicted based on their current states. $C_{attack}$ denotes the distance score for successful attacks, and $C_{collision}$ penalizes collision with other actors.
Furthermore, we can adjust the attack intensity by randomly choosing from the top-k trajectories sorted by $C_{total}(s_{1:T}^{a(i)})$, instead of choosing the most aggressive one. Additionally, we can adjust the attack planning frequency to further control the intensity of the adversarial attack.
\input{figures/method/HD_Score/hdscore}

\subsection{Evaluation}
Inspired by NAVSIM \cite{dauner2024navsim} and DriveArena \cite{yang2024drivearena}, we propose the HD-Score (HUGSIM Driving Score) to measure the performance of AD algorithms in our simulator. At each timestamp, the HD-Score is defined as:
\begin{align}
    \text{HD-Score}_t = 
    & \underbrace{\left(
        \prod_{m \in \{NC, DAC\}} score_m
    \right)}_{\text{driving policy items}} \times \nonumber \\
    & \underbrace{\left(
        \frac{\sum_{w \in \{TTC, COM\}} weight_w \times score_w}{\sum_{w \in \{TTC, COM\}} weight_w}
    \right)}_{\text{contributory items}}
\end{align}
The driving policy items include driving with no collisions ($NC$) and drivable area compliance ($DAC$), these sub-scores are crucial for driving safety. The contributory items include time-to-collision ($TTC$) and comfort ($COM$), which may not directly cause failure cases when they are low. Still, they can lead to feelings of insecurity and discomfort for passengers. We illustrate each sub-score in \figref{fig:hdscore}. The detailed definition of these sub-scores can be found in \cite{dauner2024navsim}.
In the calculation of $NC$ and $TTC$, we consider collisions with static background entities such as buildings, fences, vegetation, etc., leveraging our semantic information. This background entity collision is not included in NAVSIM nor DriveArena. 

Additionally, unlike NAVSIM and DriveArena, which evaluate ego progress ($EP$) using pseudo ground truth generated by planning algorithms such as the Predictive Driver Model (PDM)\cite{dauner2023parting}, we argue that in a closed-loop simulator, other metrics alone can sufficiently reflect the performance of AD algorithms. 
Furthermore, the PDM is not a flawless algorithm, even provided with ground-truth perception results. Additionally, there is often multiple acceptable driving style in most scenarios.
For these reasons, evaluating AD algorithms using $EP$ in a closed-loop manner becomes unsuitable. Instead of measuring ego progress at each frame, we introduce a global route completion score $R_c \in [0,1]$, which represents the percentage of the driving distance completed by AD algorithms.

The final HD-Score is averaged across all simulation timestamps and multiplied by the global route completion score $R_c$:
\begin{equation}
    \text{HD-Score} = R_c \times \frac{\sum_{t=0}^{T} \text{HD-Score}_t}{T}
\end{equation}

%% file: figures/method/HD_Score/hdscore.tex
\begin{figure}
    \centering
    \includegraphics[width=0.38\textwidth]{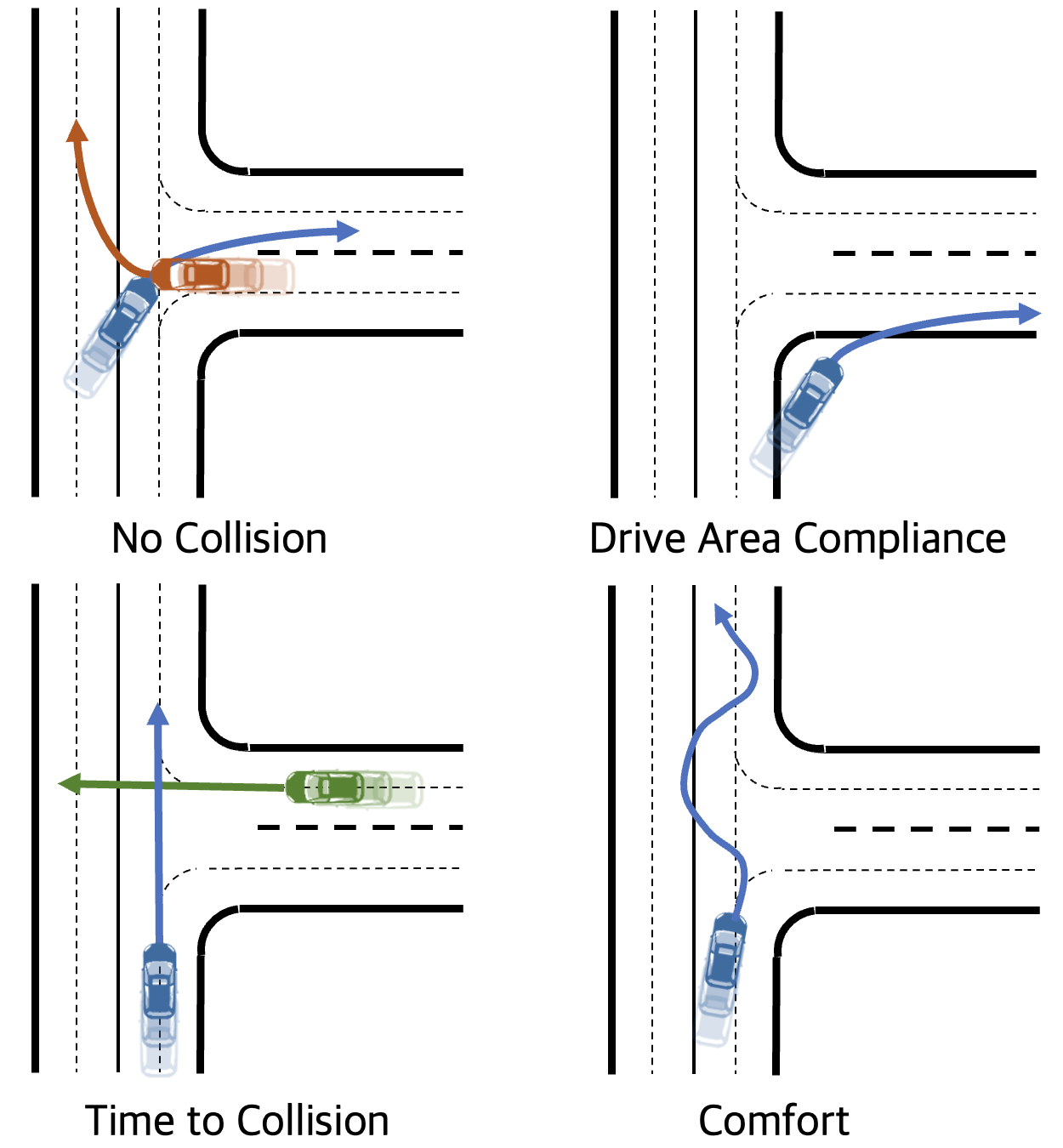}
    \caption{\textbf{Illustration of sub-scores combined in HUGSIM.} We present cases that result in the failure or reduction of the corresponding sub-scores.}
\label{fig:hdscore}
\vspace{-0.2cm}
\end{figure}

%% file: sec_render_eval.tex
\section{Rendering Evaluation}\label{sec:render_eval}

\input{figures/eval/kitti360_leaderboard/mars_comparison}
\input{figures/eval/kitti360_leaderboard/pnf_comparison}
\input{tables/kitti360_leaderboard}
\input{tables/semantic3d}

In this section, we evaluate HUGSIM from several perspectives. In \secref{sec:exp_urban_recon}, we focus on evaluating interpolated views in both dynamic and static scenes. In \secref{sec:exp_smt_recon}, we assess the semantics and geometry of reconstructed scenes. In \secref{sec:exp_extrapolated}, we evaluate extrapolated views, which are significant in a closed-loop driving simulator. Finally, in \secref{sec:exp_ablation}, we conduct ablation studies on each component of HUGSIM.

\input{tables/kitti_dynamic_noisy}
\input{figures/eval/kitti_comparison/kitti_comparison}

\subsection{Novel View Synthesis on Interpolated Views} \label{sec:exp_urban_recon}

We first evaluate HUGSIM for interpolated novel view synthesis on KITTI \cite{geiger2012we}, Virtual KITTI2 \cite{cabon2020virtual} and KITTI-360 \cite{liao2022kitti} including dynamic and static scenes. For dynamic scenes, we leverage noisy 3D bounding box predictions as input, instead of using the ground truth. 
We evaluate the dynamic scene rendering results on KITTI and vKITTI in this section because previous dynamic urban reconstruction baselines have been tested on these two datasets. We apply 50\% dropout rate following existing evaluation protocols \cite{liao2022kitti, wu2023mars} on all of these datasets.
We evaluate the dynamic scene novel view synthesis task by comparing our method with NSG \cite{ost2021neural} and MARS \cite{wu2023mars}, which are two open-source methods for dynamic urban scenes. Additionally, we compare the static novel view appearance task with mip-NeRF \cite{barron2021mip}, PNF \cite{kundu2022panoptic}, and MARS \cite{wu2023mars}. We adopt the default setting for interpolated views quantitative assessments, including the evaluation of PSNR, SSIM and LPIPS \cite{zhang2018unreasonable}. Although not our main foucus, we include a comparison using ground truth 3D bounding boxes in the appendix.

\boldparagraph{Dynamic Scene with Noisy 3D Bounding Boxes}
Following \cite{ost2021neural, wu2023mars}, we evaluate our performance on dynamic scenes of the KITTI and vKITTI datasets. In contrast to these methods that leverage ground truth poses, we investigate a more practical scenario where the bounding boxes are generated by a monocular-based 3D tracking algorithm, QD-3DT~\cite{hu2022monocular}, in \tabref{tab:dynamic_noisy}. Here, the predicted 3D bounding boxes are only provided for training views, as testing views should not be used as inputs for the tracking model. In experiments where the unicycle model is not utilized, the bounding boxes of testing views are obtained through linear interpolation from neighbour training views. Where the unicycle model is used, the bounding boxes of testing views are computed using \eqref{eq:unicycle}. For vKITTI, there is no pre-trained monocular tracking algorithm. We hence jitter the ground truth poses to simulate noisy monocular predictions, with an average noise of 0.5 meters in translation and 5 degrees in rotation. Our model's robustness wrt. various levels of noise will be analyzed in the ablation study.

\tabref{tab:dynamic_noisy} demonstrate that our method consistently outperforms against the baselines. Note, that QD-3DT yields reasonable predictions on the KITTI dataset\footnote{In fact, following the evaluation protocol of MARS, the sequences we evaluate on are used as training sequences for QD-3DT.}. Hence, NSG and MARS reconstruct the dynamic objects reasonably well, but with more blurriness and artifacts (see \figref{fig:kitti_comparison}), as they do not model the optimization of the object poses. In contrast, our method allows for reconstructing dynamic objects with sharp details, not only in cases of minor pose error on the KITTI dataset but also on the vKITTI dataset with more severe noise. We present the optimization progress of noisy bounding boxes and a comparison before and after optimization in the appendix.

\boldparagraph{Static Scene Leaderboard} We further evaluate our performance on the KITTI-360 leaderboard, which contains 5 static sequences. Our method achieves state-of-the-art performance on the leaderboard as in \tabref{tab:leaderboard_kitti360} (left), demonstrating the effectiveness of the 3D Gaussian representation in modeling complex urban scenes. As we will discuss in the ablation study, incorporating the affine transform to model camera exposure is important for reaching high fidelity. \figref{fig:compare_mars_detail} shows the qualitative comparison of our proposed method to another top-ranking method, MARS, on the leaderboard.

\subsection{Semantic and Geometric Scene Understanding} \label{sec:exp_smt_recon}

Next, we evaluate our model on various semantic and geometric scene understanding tasks on the KITTI-360 dataset \cite{liao2022kitti} as it provides dense 2D and 3D labels. We compare the semantic synthesis task with mip-NeRF \cite{barron2021mip}, PNF \cite{kundu2022panoptic}, and MARS \cite{wu2023mars}. Furthermore, we assess the quality of 3D semantic scene reconstruction by comparing it with Semantic Nerfacto \cite{tancik2023nerfstudio}. 
We follow KITTI-360, which reports the mean Intersection over Union on class (mIoU$_\text{cls}$) and category (mIoU$_\text{cat}$), respectively.  Further, we evaluate our performance on \textit{3D Semantic Segmentation} against a ground truth semantic LiDAR point cloud, measuring both geometric reconstruction quality and semantic accuracy. The geometric quality is evaluated as the chamfer distance between two point clouds, including completeness and accuracy, whereas the semantic accuracy is also measured using mIoU$_\text{cls}$. 
Correct 3D semantics, the reduction of floaters, and the reconstruction of accurate geometry are essential for building our closed-loop simulator and benchmark.

\boldparagraph{Novel View Semantic Synthesis}
Our holistic representation also enables novel view semantic synthesis. Hence, we submit our novel view semantic synthesis performance to the KITTI-360 leaderboard for comparison as well, see \tabref{tab:leaderboard_kitti360} (right). Despite not leveraging category-level prior as done in previous work~\cite{kundu2022panoptic}, our approach achieves comparable performance to the SOTA~\cite{kundu2022panoptic} as shown in \figref{fig:compare_pnf}.

\boldparagraph{3D Semantic Scene Reconstruction}
While existing 2D-to-3D semantic lifting methods solely evaluate their performance in the 2D image space, we further evaluate our performance in the 3D space to examine the underlying 3D geometry. To this goal, we leverage the ground truth LiDAR points provided by the KITTI-360 dataset for evaluation.
With each Gaussian possessing semantic information, we can obtain a semantic point cloud by extracting the Gaussian's center $\mu$ and its semantic label. 
We evaluate the geometric quality and semantic accuracy of this semantic point cloud in \tabref{tab:semantic_3d}. We compare our method with Semantic Nerfacto \cite{tancik2023nerfstudio}, a Semantic NeRF implemented using a more advanced backbone, as the state-of-the-art novel view semantic synthesis method, PNF, in \tabref{tab:leaderboard_kitti360} is not open-source. For this baseline, we extract a semantic point cloud by specifying a threshold to the density field.
While Semantic Nerfacto enables rendering faithful 2D semantic labels as shown in the appendix, the underlying 3D semantic point cloud is significantly worse in comparison. The Gaussian based representation instead allows for extracting a much more accurate semantic point cloud in comparison. Accurate geometry is crucial for implementing the closed-loop simulator.

\input{tables/extrapolated}
\input{figures/eval/ext_comparison/comp_waymo}

\subsection{Novel View Synthesis on Extrapolated Views} \label{sec:exp_extrapolated}
The ego vehicle can change lanes or view scenes from a wide range of perspectives beyond the training views, which may lead to artifacts in novel view synthesis, particularly causing distortion in lanes and other ground-level signals.
However, since most extrapolated camera poses are not directly observed, quantitative metrics like PSNR that require ground-truth images cannot be computed directly. Instead, we use the Kernel Inception Distance (KID) \cite{binkowski2018demystifying} to measure the similarity in distribution between extrapolated views and real captured images.

We evaluate extrapolated views on the Waymo Open Dataset \cite{Sun2020CVPR} and nuScenes \cite{caesar2020nuscenes} by comparing our method to NeuRAD \cite{tonderski2024neurad} and StreetGaussian \cite{yan2024street}, two recent approaches for novel view synthesis in urban scenes. StreetGaussian supports the Waymo dataset, while NeuRAD supports both Waymo and nuScenes datasets. Notably, NeuRAD is also the reconstruction and rendering method used in NeuroNCAP \cite{ljungbergh2024neuroncap}. 
We further compare our results to RoGS \cite{feng2024rogs}, a method designed specifically for ground reconstruction in nuScenes.

\boldparagraph{Full Scene Rendering}
We selected two sequences from the Waymo dataset and two from nuScenes for comparison. During training, HUGSIM, StreetGaussian \cite{yan2024street}, and NeuRAD \cite{tonderski2024neurad} drop every 5th frame.  To generate extrapolated views, we shift the original views horizontally to simulate lane changes and slightly adjust the yaw angle to simulate varied observation angles. In this experiment, we evaluate only the front camera rendering results, as the front view plays the most crucial role for driving. For a fair comparison, all methods use the same extrapolated views within the same sequences.

\input{figures/eval/ext_comparison/comp_nusc}
\input{tables/comp_rogs}

\input{figures/eval/ext_comparison/comp_rogs}
\input{tables/ablation/ablation_dynamic}
\input{tables/ablation/ablation_static}

\tabref{tab:extrapolated} shows that our method achieves comparable or better performance in interpolated views and state-of-the-art performance on extrapolated views. Additionally, while both StreetGaussian and NeuRAD require LiDAR point clouds as input, HUGSIM relies solely on RGB images. HUGSIM also outperforms the compared methods in rendering speed and number of Gaussians. \figref{fig:compare_waymo} and \figref{fig:compare_nusc} provides a qualitative comparison of HUGSIM and other methods. Without ground constraints, significant distortion is evident in NeuRAD and StreetGaussian, while our method produces realistic results in extrapolated views.

\boldparagraph{Ground Rendering}
We further compare our ground model with RoGS \cite{feng2024rogs}, an approach specifically designed for ground reconstruction using 2D Gaussian Splatting \cite{huang20242d}. 
The primary difference between RoGS and our method is that RoGS uniformly distributes Gaussians on the ground plane and fixes their attributes, allowing only spherical harmonics and height to be learned. In contrast, our method fixes only the y-axis scale and rotations of Gaussians and applies $\cL_{ground}$ to constrain the ground flatness, as demonstrated in Section \ref{sec:ground_model}.

The experiments are conducted on four sequences from nuScenes. We utilize a semantic mask to color non-ground pixels black. Since the top half of the images does not contain any ground pixels, we focus our evaluation on the lower half of the images for this experiment. 
\tabref{tab:comp_rogs} and \figref{fig:comp_rogs} show our method surpasses RoGS both quantitatively and qualitatively, while also using fewer Gaussians to reconstruct the scene and exhibiting greater flexibility without the need to assign the hyperparameter of road width. We attribute the advantages of our method to its learnable position, scaling, and opacity, allowing it to allocate more Gaussians to texture-rich areas and fewer to regions with minimal texture. This leads to to a reduced number of Gaussians while maintaining high texture detail. In contrast, the uniform distribution of large numbers of Gaussians, as implemented in RoGS, still exists holes and results in less satisfactory rendering outcomes.

\input{tables/ablation/ablation_ground}

\subsection{Ablation} \label{sec:exp_ablation}
We conduct ablation studies on dynamic, static scenes, ground model and inserted vehicles.

\input{figures/eval/ablation/unicycle/unicycle}

\boldparagraph{Dynamic Scene} As KITTI provides accurate 3D bounding box ground truth, we ablate the effectiveness of our unicycle model on KITTI by manually adding noise to the 3D bounding boxes and evaluate both the novel view synthesis results and the tracking performance, see \tabref{tab:ablation_dynamic}. 
In this experiment, we compare our full model to two variants, i.e., using the noises without optimization (w/o opt., w/o uni.), and performing na\"ive per-frame optimization without using the unicycle model (w/ opt., w/o uni.). We evaluate \textit{3D tracking} performance by measuring the rotation and translation error $e_\bR$ and $e_\bt$ of our optimized 3D bounding boxes wrt. the ground truth following \cite{chen2020category}. 
The results validate the effectiveness of the unicycle model, which obviously improves the rendering quality and 3D tracking accuracy. Qualitative results in \figref{fig:comp_noisy} further verify the effectiveness of our unicycle model in enabling accuracy object reconstruction given noisy 3D bounding boxes.

\boldparagraph{Static Scene} We study the effect of different components on three static scenes of KITTI-360 in \tabref{tab:ablation_static}. This allows us to ablate design choices without mixing up the impact of dynamic objects. The results indicate the significance of exposure modeling, which is particularly important for scenes with strong exposure variance. The semantic loss has little contribution to improving novel view synthesis. It is rational as imposing a constraint on the semantics does not necessarily contribute to appearance. However, note that incorporating the semantic supervision improves the underlying geometry, see the appendix for qualitative comparison. Moreover, we utilize the semantic information for collision detection of our closed-loop simulator.

\boldparagraph{Ground Model} We ablate the effectiveness of $\cL_{ground}$ and importance of learnable position, scaling and opacity, as shown \tabref{tab:ablation_ground} and \figref{fig:ablation_ground}. PSNR, SSIM and LPIPS are evaluated in interpolated views, while KID is evaluated in extrapolated views. The results demonstrate the importance of $\cL_{ground}$ in significantly reducing floating Gaussians. 
Additionally, the learnable position, scaling, and opacity improve the reconstruction fidelity of our road model.

\input{figures/eval/ablation/ground/ground}

\boldparagraph{Inserted Vehicles} To compare simulation quality with native versus non-native vehicles, we inserted vehicles with random poses into a sequence, as shown in \figref{fig:ablation_insert}. The results indicate that native vehicles yield less satisfactory rendering results due to limited observations.

%% file: figures/eval/kitti360_leaderboard/mars_comparison.tex
\begin{figure}[t!]
     \centering
     \small 
     \setlength{\tabcolsep}{0pt}
     \def\mywidth{2.0cm}
     \begin{tabular}{P{\mywidth}P{\mywidth}P{\mywidth}P{\mywidth}}
     \includegraphics[width=\mywidth]{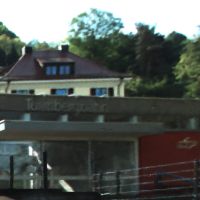}  
     & \includegraphics[width=\mywidth]{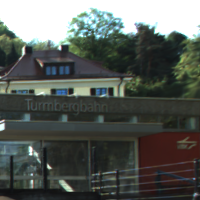}
     & \includegraphics[width=\mywidth]{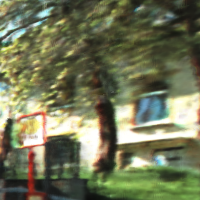}
     & \includegraphics[width=\mywidth]{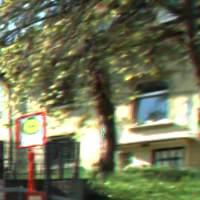} \\
     \includegraphics[width=\mywidth]{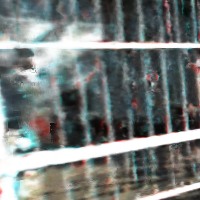}  
     & \includegraphics[width=\mywidth]{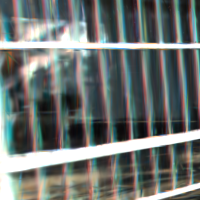}
     & \includegraphics[width=\mywidth]{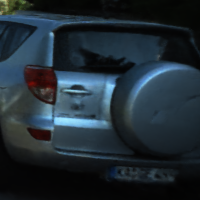}
     & \includegraphics[width=\mywidth]{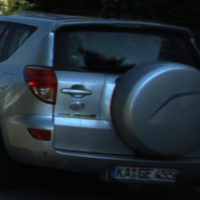} \\
     \rotatebox{0}{MARS} & \rotatebox{0}{Ours} & \rotatebox{0}{MARS} & \rotatebox{0}{Ours}\\
     \end{tabular}
     \vspace{-0.3cm}
     \caption{\textbf{Details Qualitative Comparison} with MARS on KITTI-360 Leaderboard.}
     \vspace{-0.2cm}
\label{fig:compare_mars_detail}
\end{figure}

%% file: figures/eval/kitti360_leaderboard/pnf_comparison.tex
\begin{figure}[t!]
     \centering
     \small 
     \setlength{\tabcolsep}{0pt}
     \def\mywidth{4.2cm}
     \begin{tabular}{P{\mywidth}P{\mywidth}}
     \includegraphics[width=\mywidth]{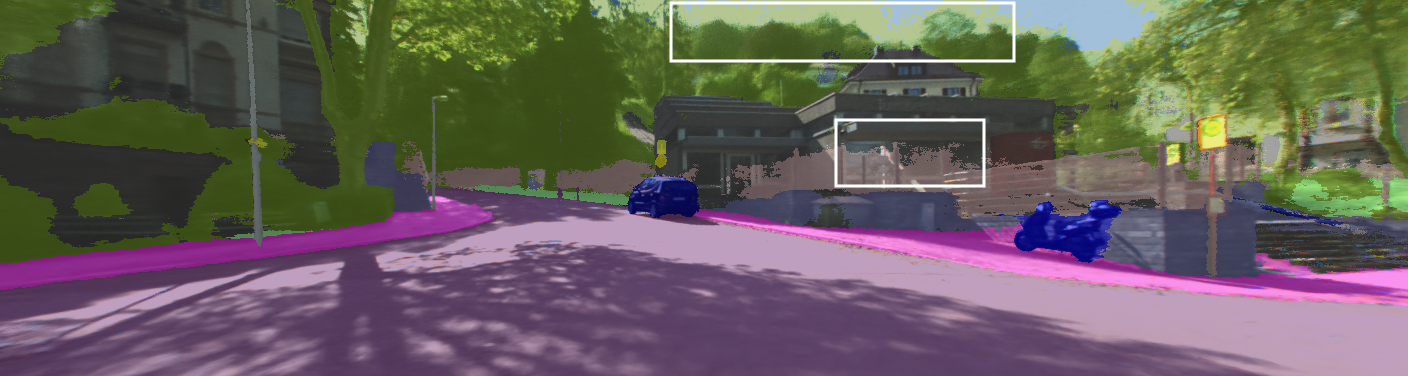}  & \includegraphics[width=\mywidth]{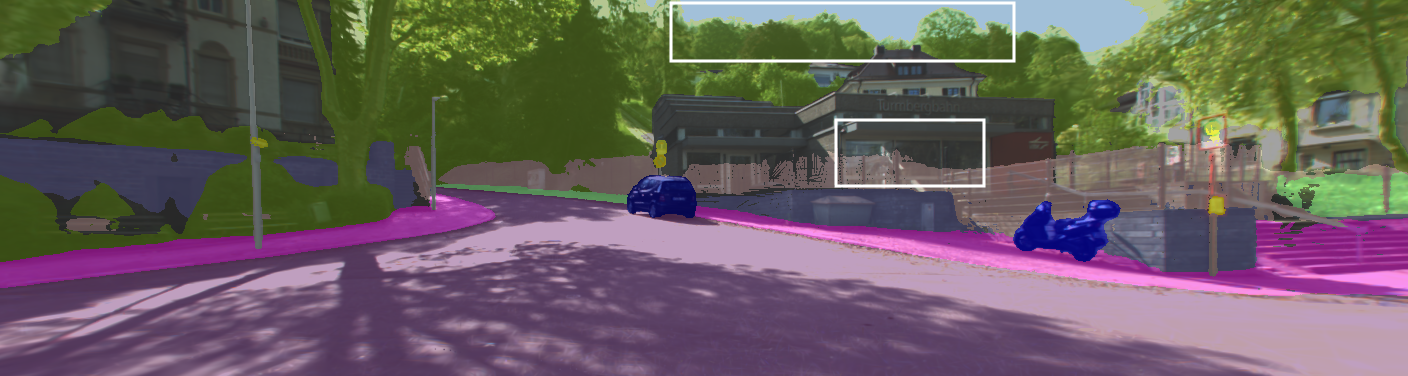} \\
     \includegraphics[width=\mywidth]{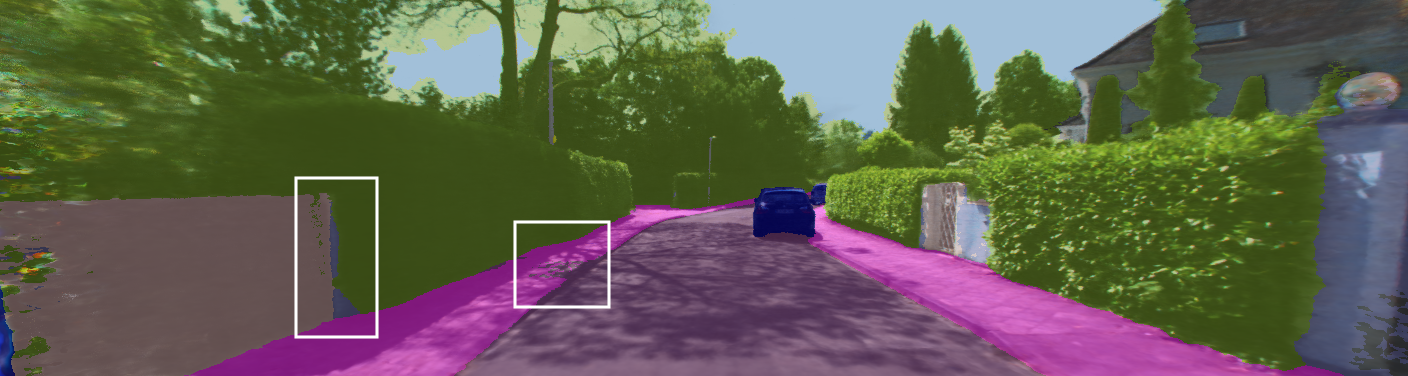}  & \includegraphics[width=\mywidth]{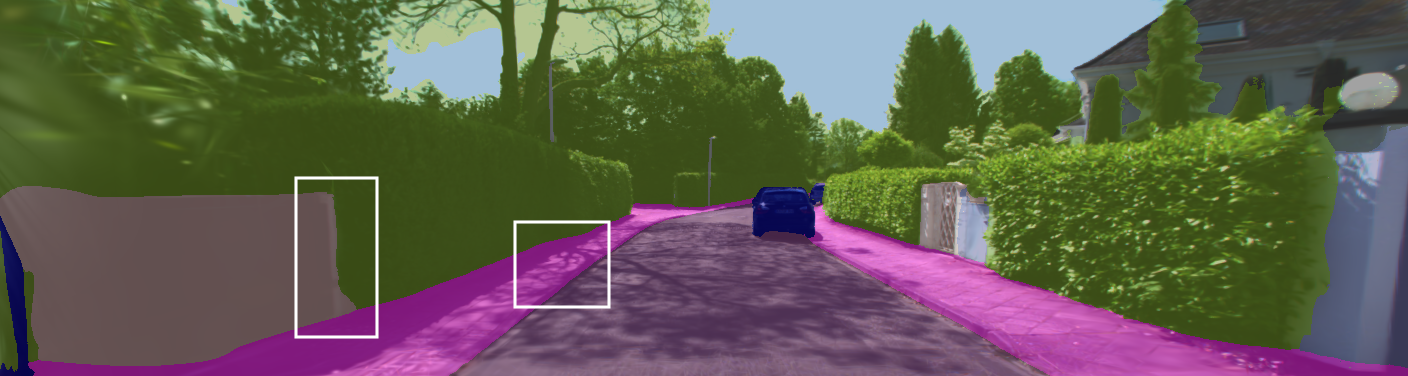} \\
     \includegraphics[width=\mywidth]{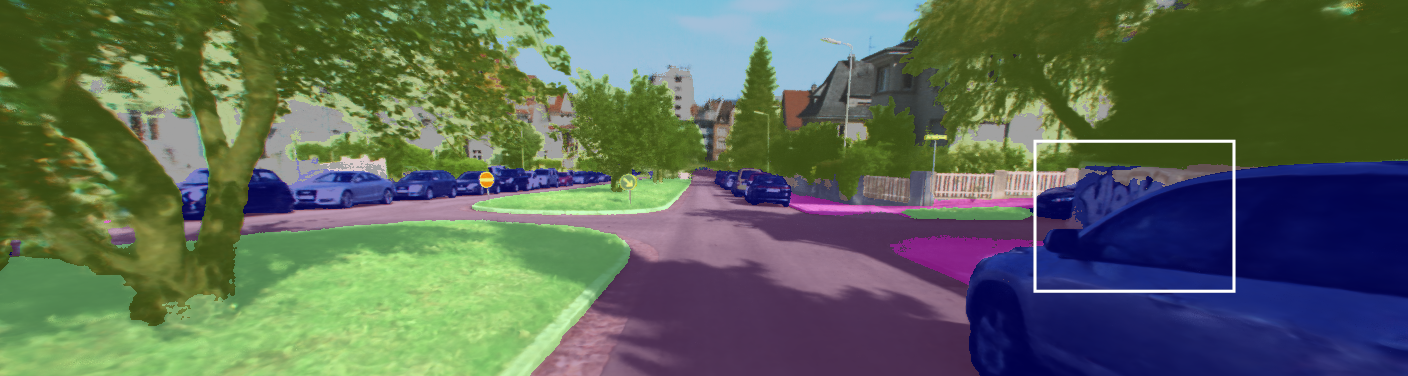}  & \includegraphics[width=\mywidth]{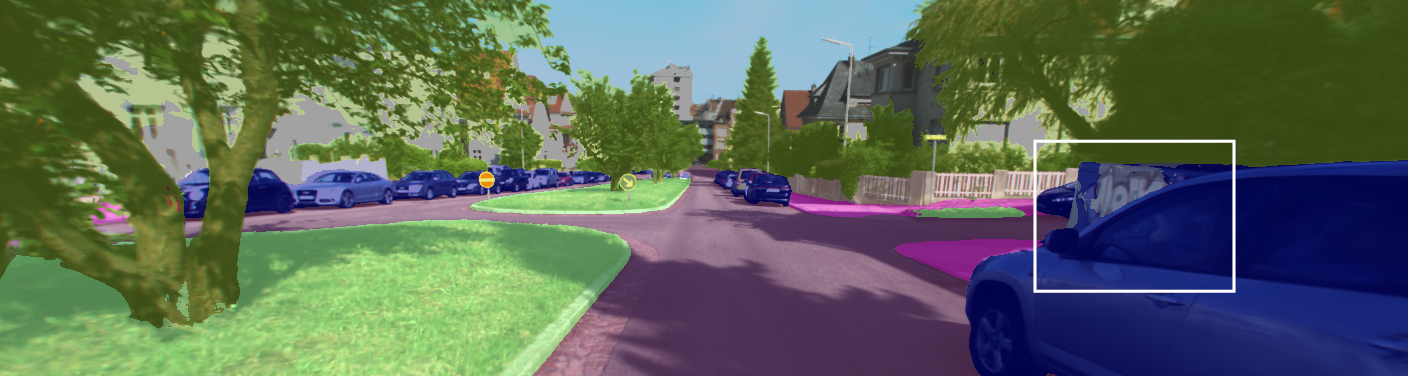} \\
    \rotatebox{0}{PNF} & \rotatebox{0}{Ours}
     \end{tabular}
    \vspace{-0.3cm}
     \caption{\textbf{Qualitative Comparison} with PNF on KITTI-360 Leaderboard.}
     \vspace{-0.5cm}
\label{fig:compare_pnf}
\end{figure}

%% file: tables/kitti360_leaderboard.tex
\begin{table}[t]
\centering
\small
\setlength{\tabcolsep}{0.6pt}
\def\mywidth{1.3cm}
\begin{tabular}{lP{1.2cm}P{1.2cm}P{1.2cm}|P{\mywidth}P{\mywidth}}
\toprule
& PSNR$\uparrow$     & SSIM$\uparrow$  & LPIPS$\downarrow$   & mIoU$_\text{cls}\uparrow$  & mIoU$_\text{cat}\uparrow$       \\
\cline{2-6}
mip-NeRF \cite{barron2021mip}
& 21.54                 & 0.778          & 0.365          & 48.25          & 67.47             \\
PNF \cite{kundu2022panoptic}   
& 22.07                 & 0.820          & 0.221          & \textbf{73.06} & 84.97       \\
MARS \cite{wu2023mars}
& 23.09                 & 0.857          & 0.174          & -              & - \\
Ours     
& \textbf{23.38}        & \textbf{0.870} & \textbf{0.121} & 72.65          & \textbf{85.64}        \\ 
\bottomrule
\end{tabular}
\vspace{-0.2cm}
\caption{\textbf{Novel View Semantic and Appearance Synthesis} on KITTI-360.}
\label{tab:leaderboard_kitti360}
\vspace{-0.3cm}
\end{table}

%% file: tables/semantic3d.tex
\begin{table}[t]
\centering
\small 
\begin{tabular}{lcccc}
\toprule 
& acc.$\downarrow$ & comp.$\downarrow$  & mIoU$_\text{cls}\uparrow$ \\
\cline{2-4}
Semantic Nerfacto
& 1.508 & 24.28 & 0.055 \\
Ours
& \textbf{0.233} & \textbf{0.214} & \textbf{0.505} \\ 
\bottomrule
\end{tabular}
\vspace{-0.2cm}
\caption{\textbf{3D Semantic Reconstruction} on KITTI-360. Note that all metrics are calculated in 3D space. }
\label{tab:semantic_3d}
\vspace{-0.4cm}
\end{table}

%% file: tables/kitti_dynamic_noisy.tex
\begin{table*}
\centering
\small
\setlength{\tabcolsep}{4pt}
\begin{tabular}{@{\extracolsep{2pt}}lcccccccccccc@{}} 
\toprule
\multicolumn{1}{c}{} & \multicolumn{3}{c}{KITTI Scene02} & \multicolumn{3}{c}{KITTI Scene06} & \multicolumn{3}{c}{vKITTI Scene02} & \multicolumn{3}{c}{vKITTI Scene06} \\
& PSNR$\uparrow$  & SSIM$\uparrow$  & LPIPS$\downarrow$  & PSNR$\uparrow$  & SSIM$\uparrow$  & LPIPS$\downarrow$ & PSNR$\uparrow$  & SSIM$\uparrow$  & LPIPS$\downarrow$  & PSNR$\uparrow$  & SSIM$\uparrow$  & LPIPS$\downarrow$  \\
\cline{2-4} \cline{5-7} \cline{8-10}  \cline{11-13} 
NSG \cite{ost2021neural} 
& 23.00 & 0.664 & 0.373 & 23.78 & 0.717 & 0.234 & 21.40 & 0.689 & 0.376 & 20.60 & 0.719 & 0.255 \\
MARS \cite{wu2023mars} 
& 23.30 & 0.731 & 0.139 & 25.09 & 0.856 & 0.083 & 22.67 & 0.882 & 0.128 & 21.67 & 0.856 & 0.134 \\
Ours 
& \textbf{25.42} & \textbf{0.821} & \textbf{0.092} & \textbf{28.20} & \textbf{0.919} & \textbf{0.027} & \textbf{26.21} & \textbf{0.911} & \textbf{0.040} & \textbf{26.65} & \textbf{0.921} & \textbf{0.030} \\
\bottomrule
\end{tabular}
\vspace{-0.2cm}
\caption{\textbf{Novel View Synthesis on Dynamic Scenes} with predicted or noisy 3D trackings.}
\label{tab:dynamic_noisy}
\vspace{-0.2cm}
\end{table*}

%% file: figures/eval/kitti_comparison/kitti_comparison.tex
\begin{figure*}[t!]
     \centering
     \small 
     \setlength{\tabcolsep}{0pt}
     \def\mywidth{4.2cm}
     \begin{tabular}{P{0.5cm}P{\mywidth}P{\mywidth}P{\mywidth}P{\mywidth}}
     \multirow{2}{*}{\rotatebox{90}{KITTI~~~~}} &
     \includegraphics[width=\mywidth]{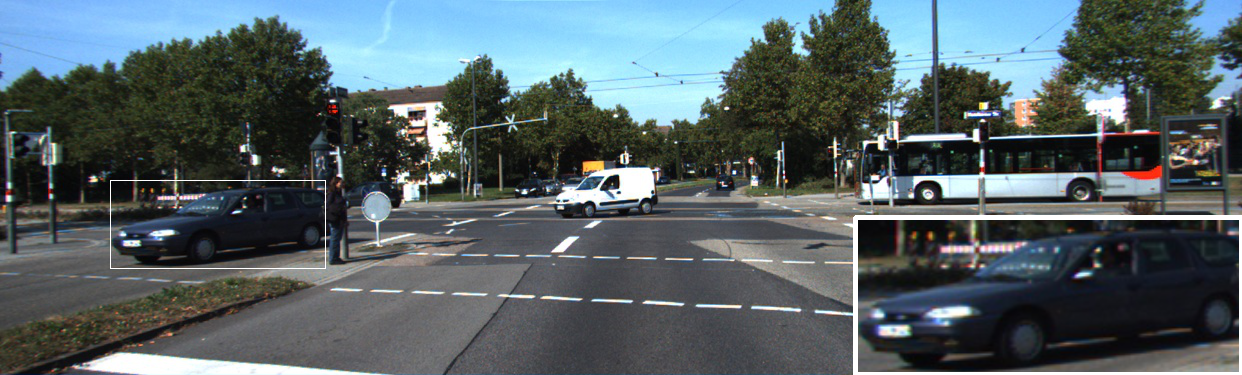}  & 
     \includegraphics[width=\mywidth]{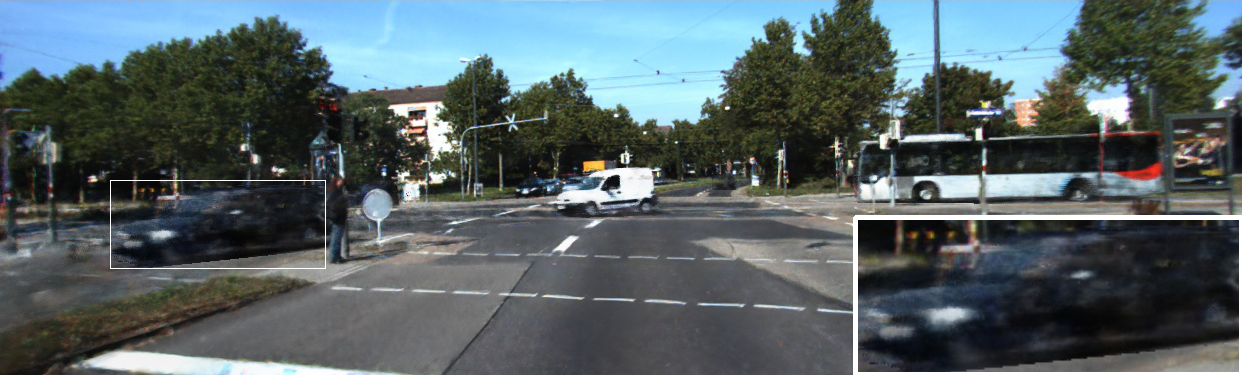} &
     \includegraphics[width=\mywidth]{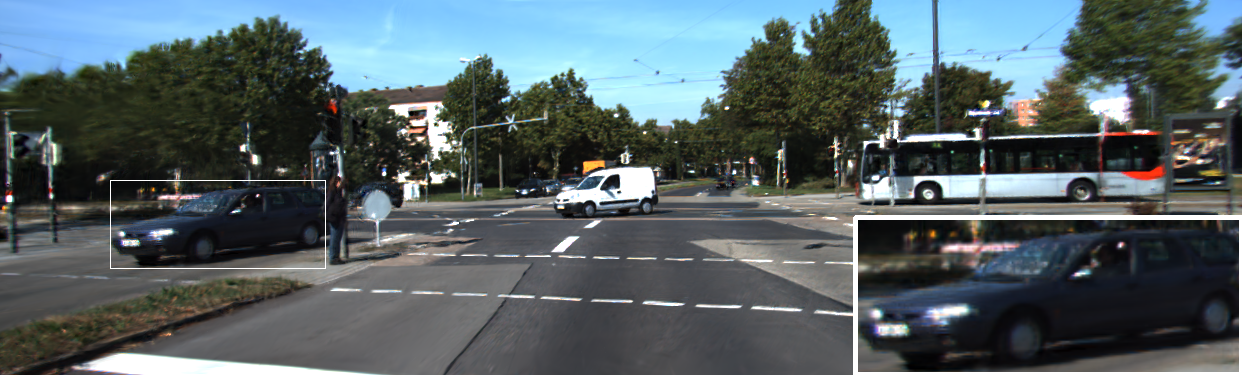} &
     \includegraphics[width=\mywidth]{figures/eval/kitti_comparison/kitti02_pseudo/gt_paste.png} \\
     
     &
     \includegraphics[width=\mywidth]{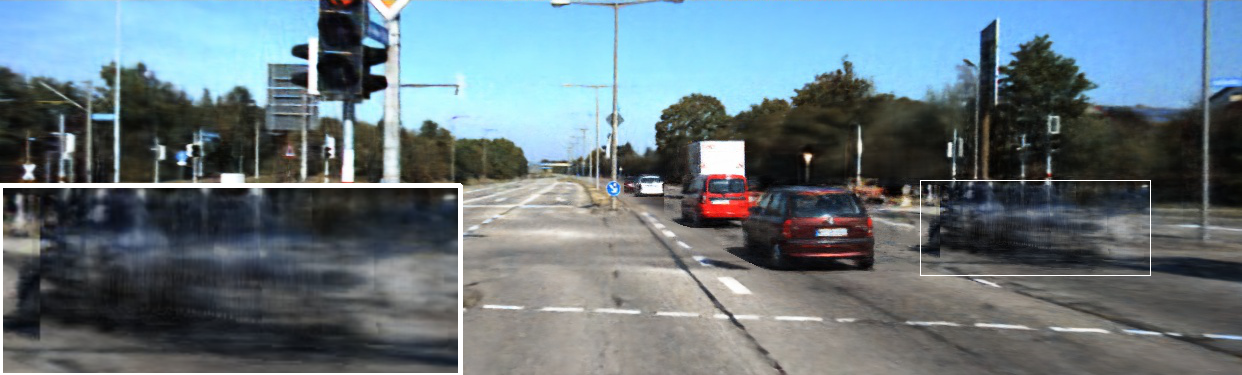}  & 
     \includegraphics[width=\mywidth]{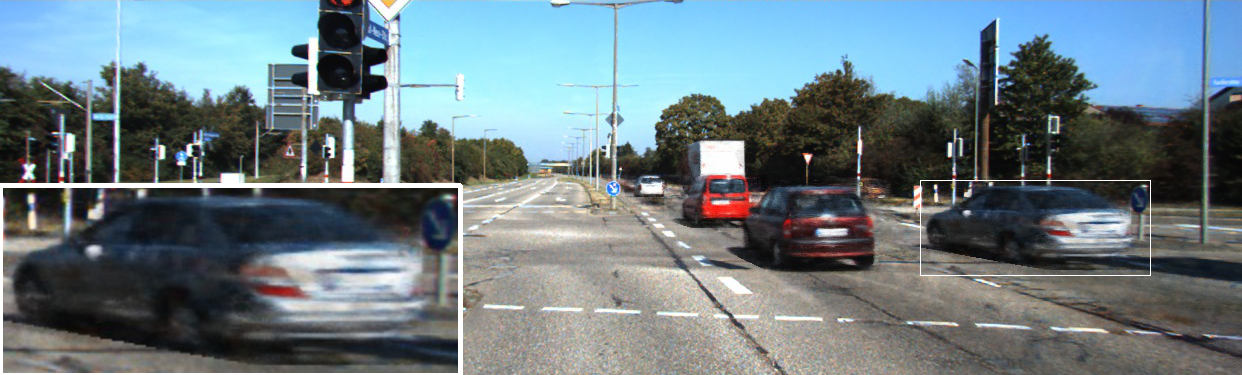} &
     \includegraphics[width=\mywidth]{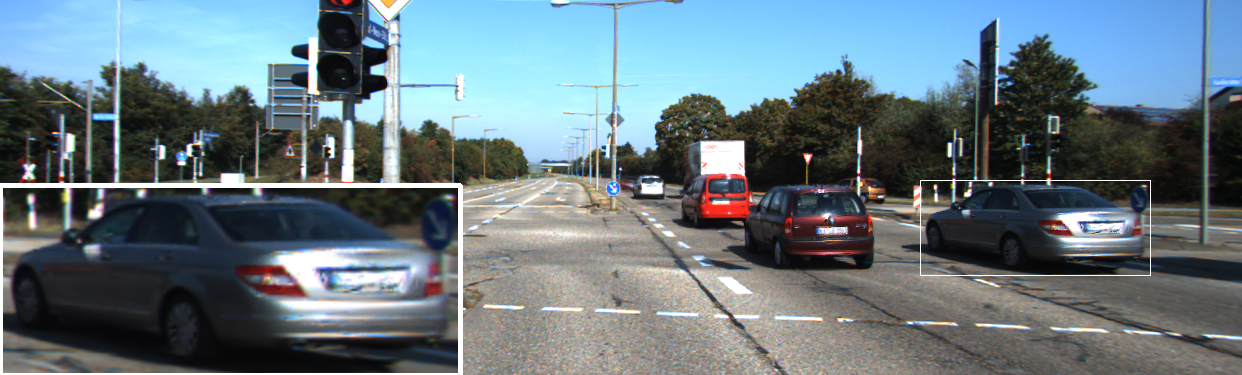} & 
     \includegraphics[width=\mywidth]{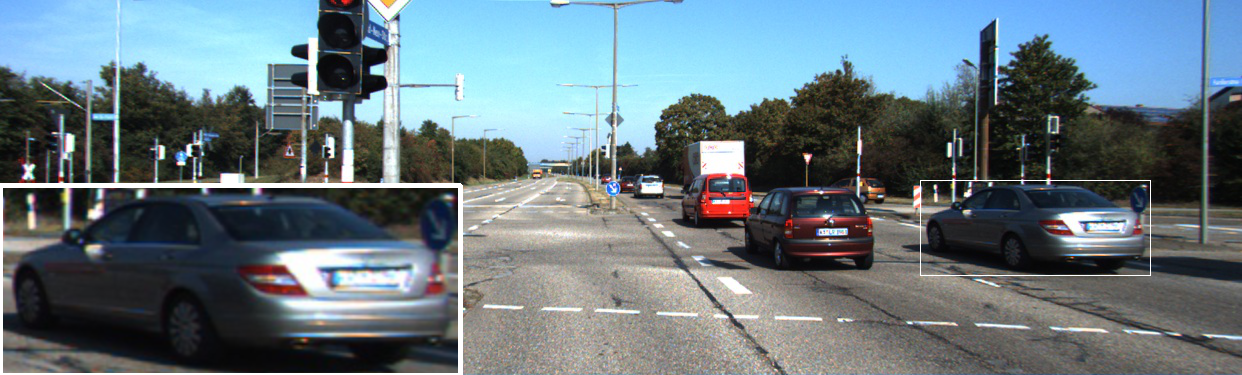} \\

     \multirow{2}{*}{\rotatebox{90}{vKITTI~~~~}} &
     \includegraphics[width=\mywidth]{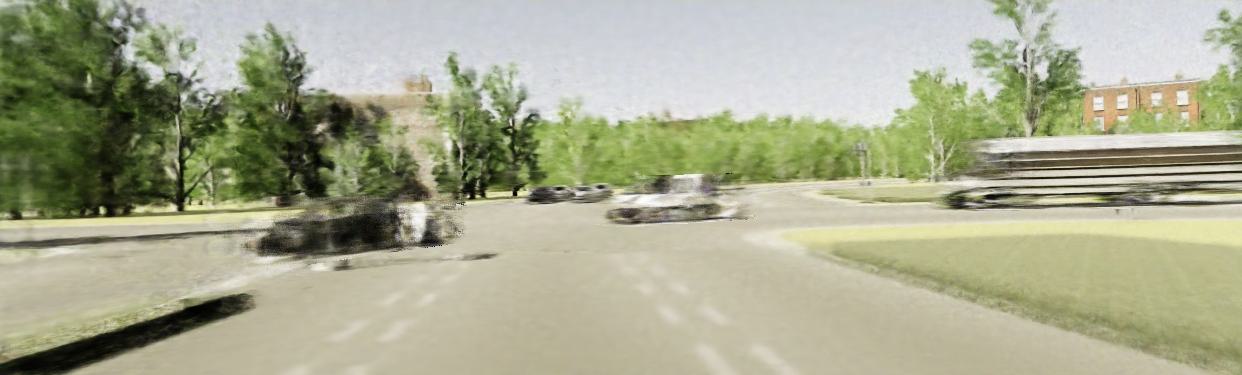}  & 
     \includegraphics[width=\mywidth]{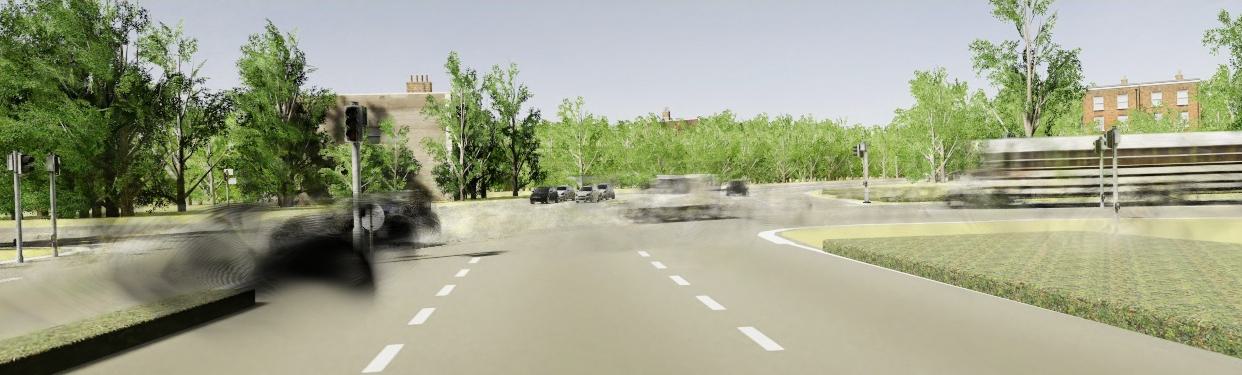} &
     \includegraphics[width=\mywidth]{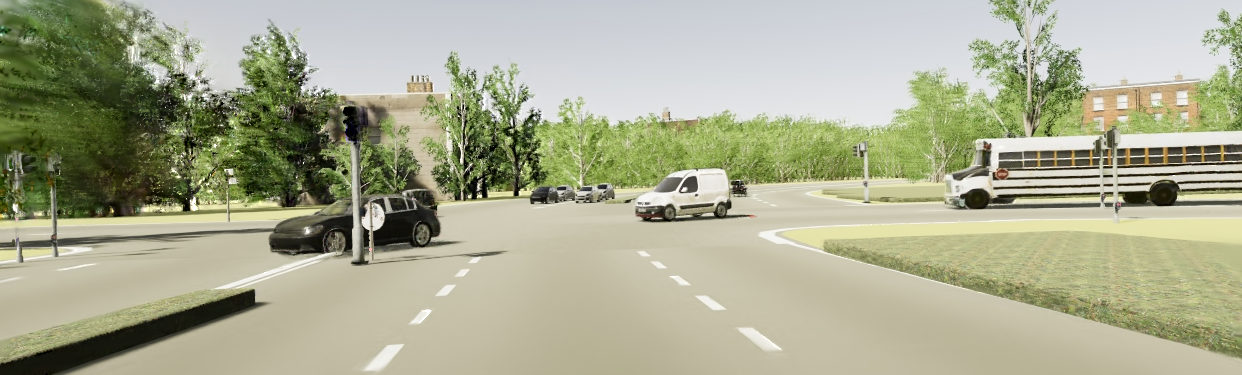} & 
     \includegraphics[width=\mywidth]{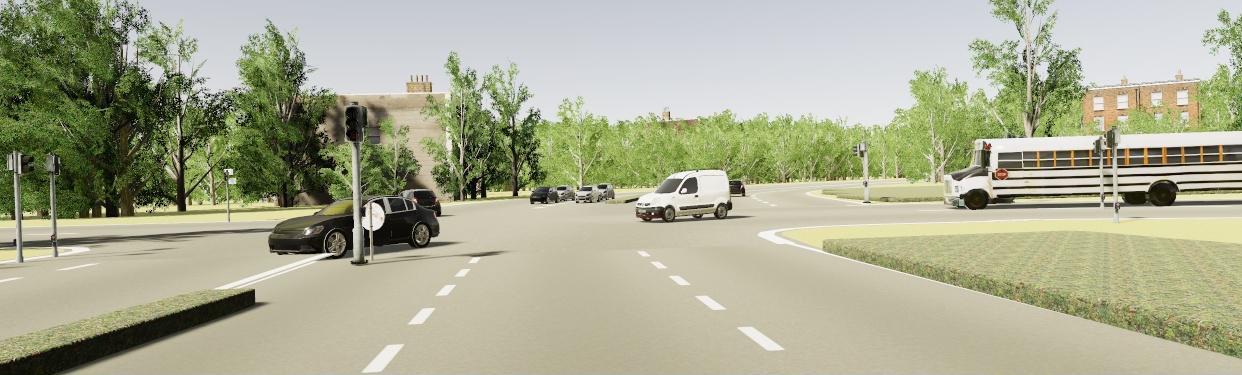} \\
     
     &
     \includegraphics[width=\mywidth]{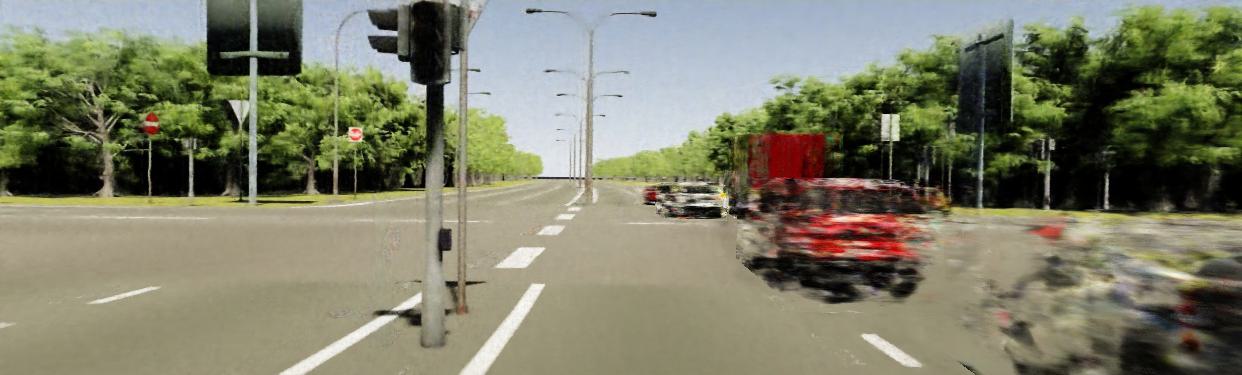}  & 
     \includegraphics[width=\mywidth]{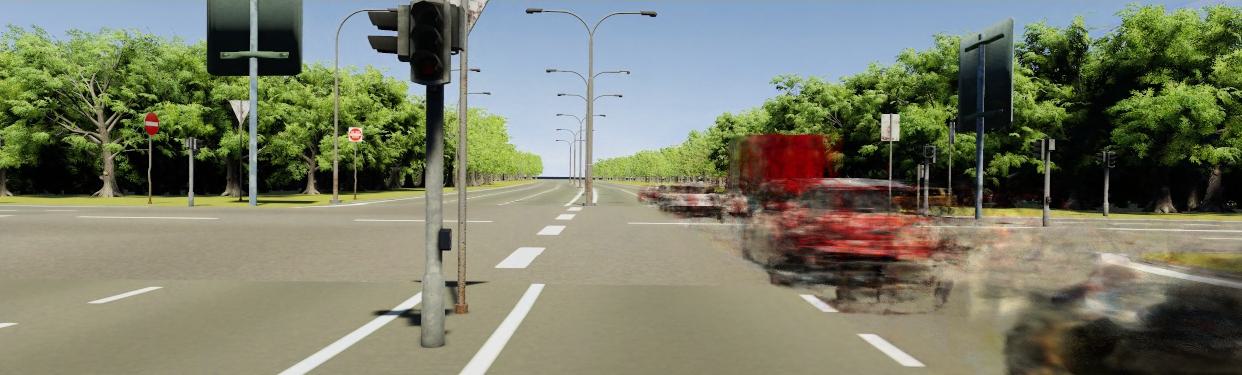} &
     \includegraphics[width=\mywidth]{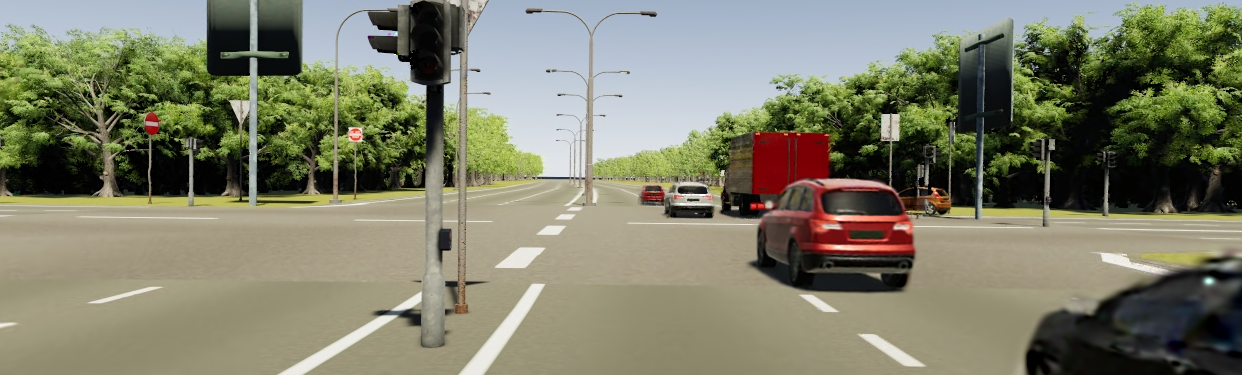} & 
     \includegraphics[width=\mywidth]{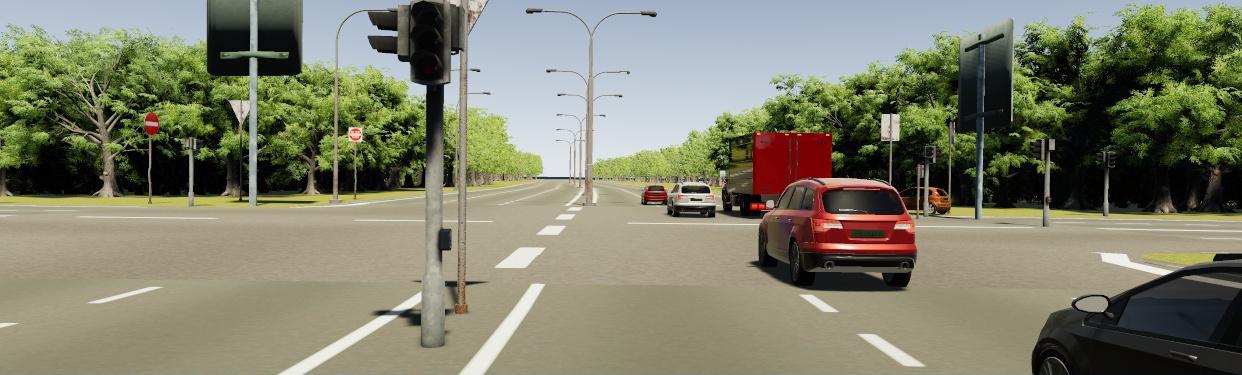} \\
     \rotatebox{0}{} & \rotatebox{0}{NSG} & \rotatebox{0}{MARS} & \rotatebox{0}{Ours}  & \rotatebox{0}{GT}  \\

     \end{tabular}\vspace{-0.2cm}
     \caption{\textbf{Qualitative Comparison} on KITTI and vKITTI. We use monocular-based 3D bounding box predictions for KITTI, and manually jittered 3D bounding boxes for vKITTI. We zoom in on a patch of a dynamic object for each KITTI scene.}
    \vspace{-0.4cm}
\label{fig:kitti_comparison}
\end{figure*}

%% file: tables/extrapolated.tex
\begin{table*}
\centering
\small
\setlength{\tabcolsep}{4pt}
\begin{tabular}{@{\extracolsep{0pt}}lcccc|cccc|ccc@{}} 
\toprule
\multicolumn{1}{c}{} & \multicolumn{4}{c|}{Waymo} & \multicolumn{4}{c|}{NuScenes} & \multicolumn{3}{c}{Attributes} \\
& PSNR$\uparrow$  & SSIM$\uparrow$  & LPIPS$\downarrow$  & $\text{KID}_{extrap}\downarrow$ & PSNR$\uparrow$  & SSIM$\uparrow$  & LPIPS$\downarrow$  & $\text{KID}_{extrap}\downarrow$ & FPS$\uparrow$ & \#GS$\downarrow$ & Inputs \\
\cline{2-5} \cline{6-9} \cline{10-12}
NeuRAD \cite{tonderski2024neurad} & \textbf{29.18} & 0.839 & 0.120 & 0.094 & 24.49 & 0.692 & 0.287 & 0.082 & 2.61 & - & LiDAR+RGB  \\
StreetGaussian \cite{yan2024street} & 27.20 & 0.868 & 0.113 & 0.151 & - & - & - & - & 66.50 & 6.85M & LiDAR+RGB \\
Ours & 28.97 & \textbf{0.872} & \textbf{0.110} & \textbf{0.077} & \textbf{25.36} & \textbf{0.776} & \textbf{0.171} & \textbf{0.062} & \textbf{89.15} & \textbf{4.45M} & \textbf{RGB} \\
\bottomrule
\end{tabular}
\vspace{-0.2cm}
\caption{\textbf{Quantitative Comparison} with NeuRAD and StreetGaussian. All the experiments are conducted on a NVIDIA RTX 3090.}
\vspace{-0.6cm}
\label{tab:extrapolated}
\end{table*}

%% file: figures/eval/ext_comparison/comp_waymo.tex
\begin{figure}[t!]
     \centering
     \small 
     \setlength{\tabcolsep}{0pt}
     \def\mywidth{4.2cm}
     \begin{tabular}{P{0.5cm}P{\mywidth}P{\mywidth}}
     \rotatebox{90}{StreetGaussian} & \includegraphics[width=\mywidth]{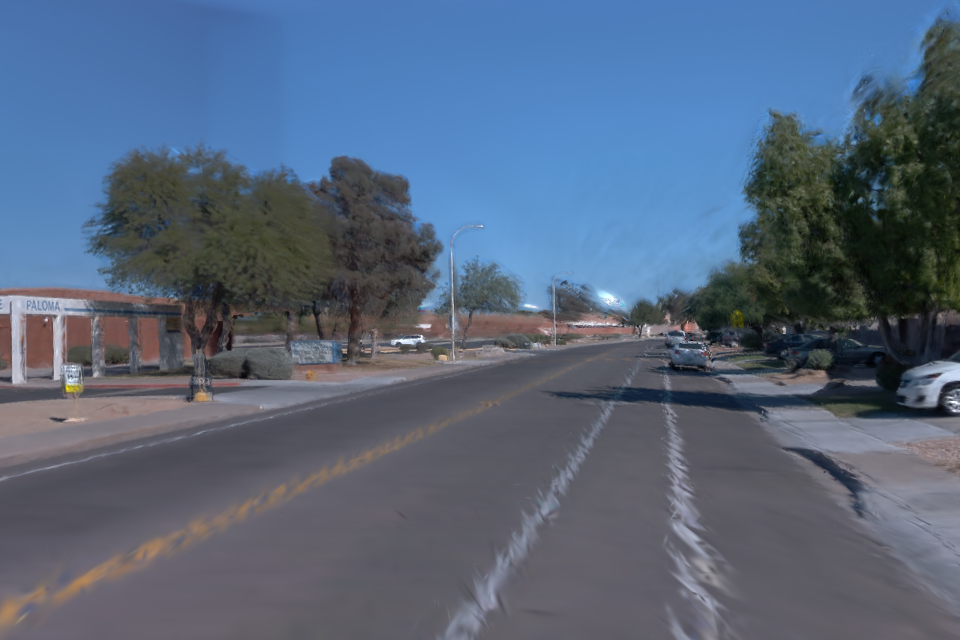}
     & \includegraphics[width=\mywidth]{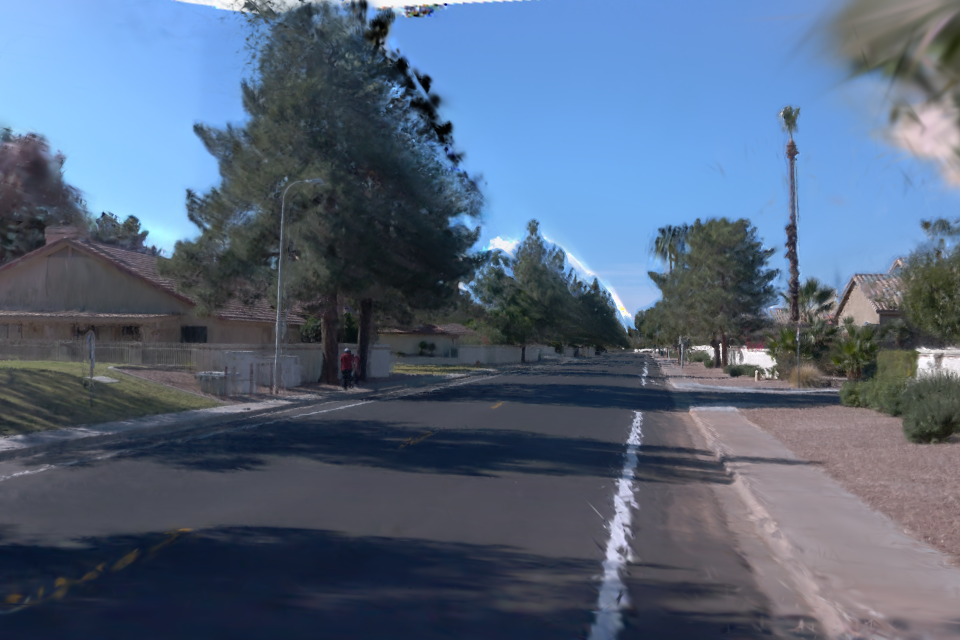} \\
     
     \rotatebox{90}{NeuRAD} & \includegraphics[width=\mywidth]{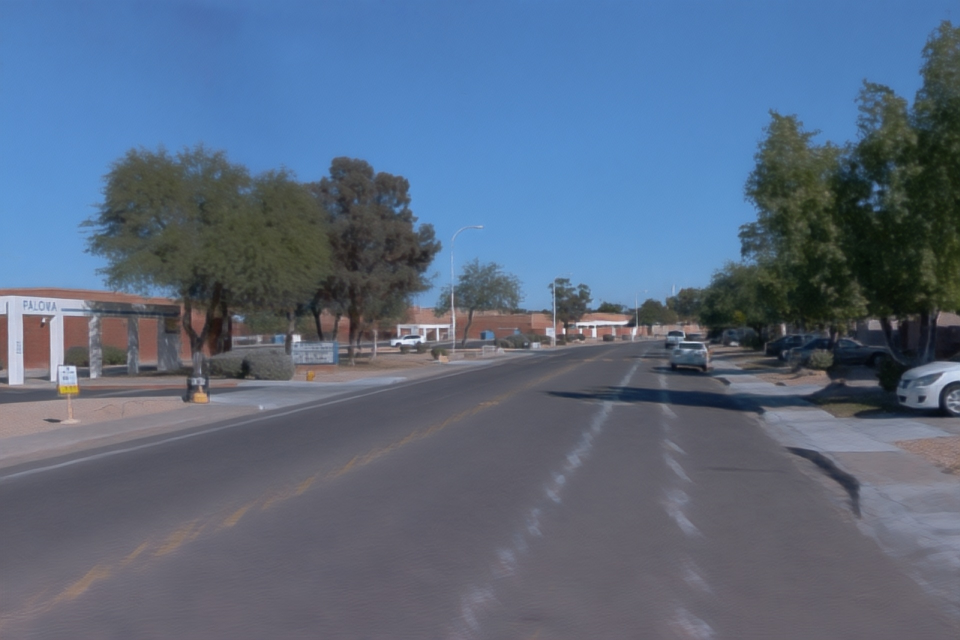}
     & \includegraphics[width=\mywidth]{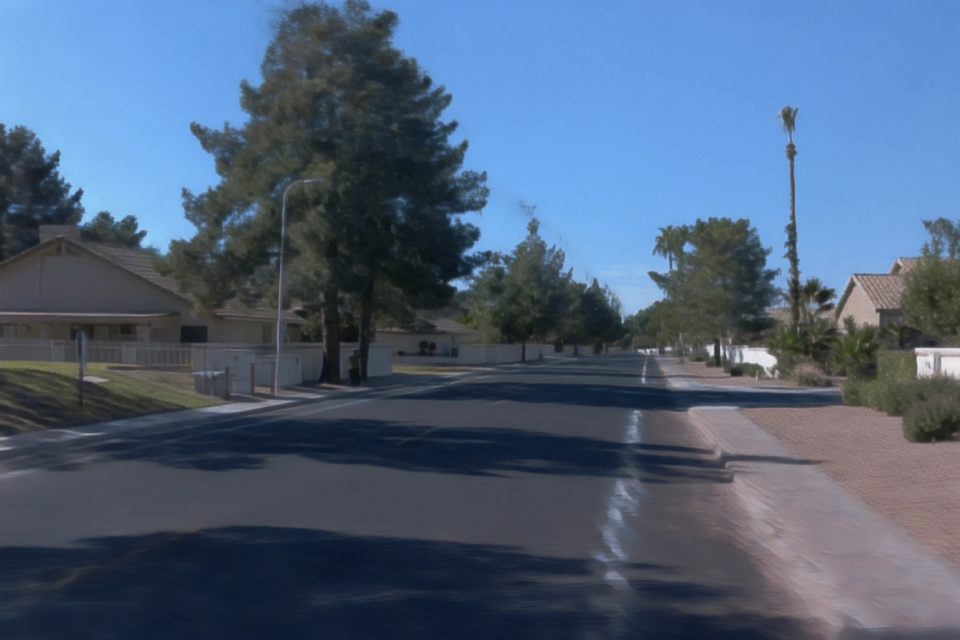} \\
     
     \rotatebox{90}{Ours} & \includegraphics[width=\mywidth]{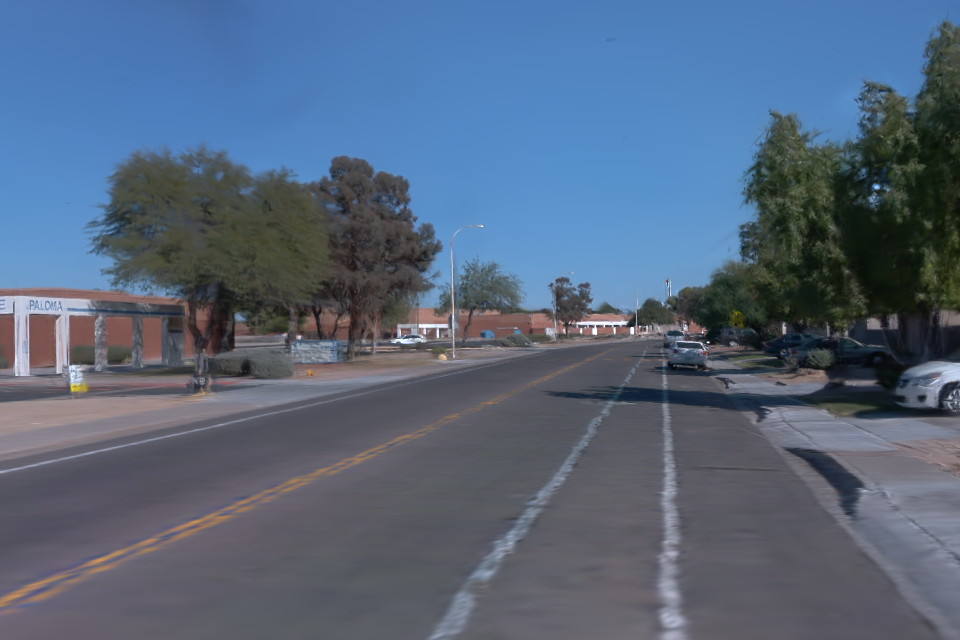}
     & \includegraphics[width=\mywidth]{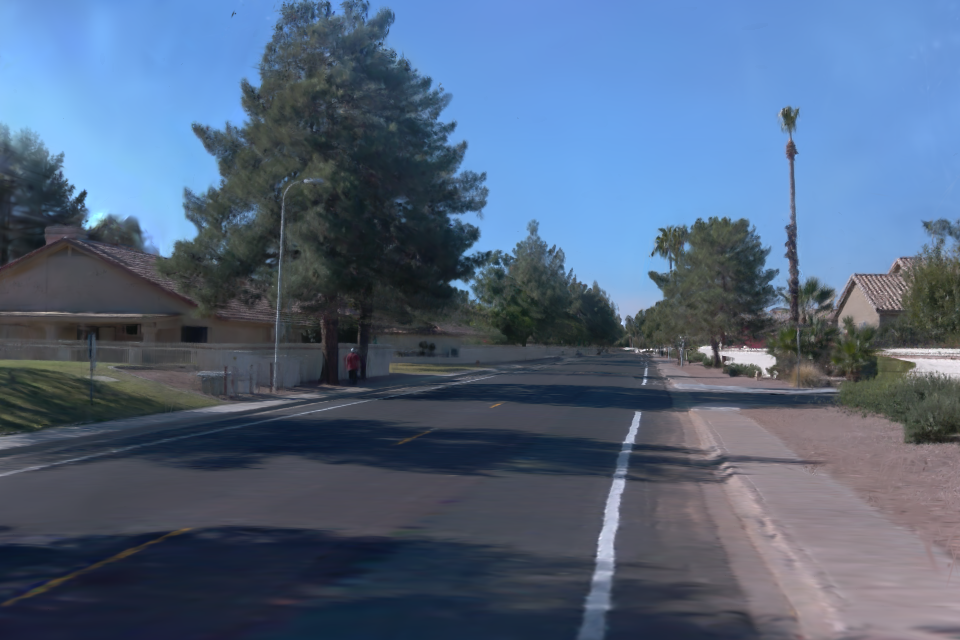} \\
     \end{tabular}
     \vspace{-0.3cm}
     \caption{\textbf{Extrapolated Views Qualitative Comparison} with StreetGaussian and NeuRAD \cite{yan2024street} on Waymo.} 
     \vspace{-0.2cm}
\label{fig:compare_waymo}
\end{figure}

%% file: figures/eval/ext_comparison/comp_nusc.tex
\begin{figure}[t!]
     \centering
     \small 
     \setlength{\tabcolsep}{0pt}
     \def\mywidth{4.2cm}
     \begin{tabular}{P{0.5cm}P{\mywidth}P{\mywidth}}
     
     \rotatebox{90}{NeuRAD} & \includegraphics[width=\mywidth]{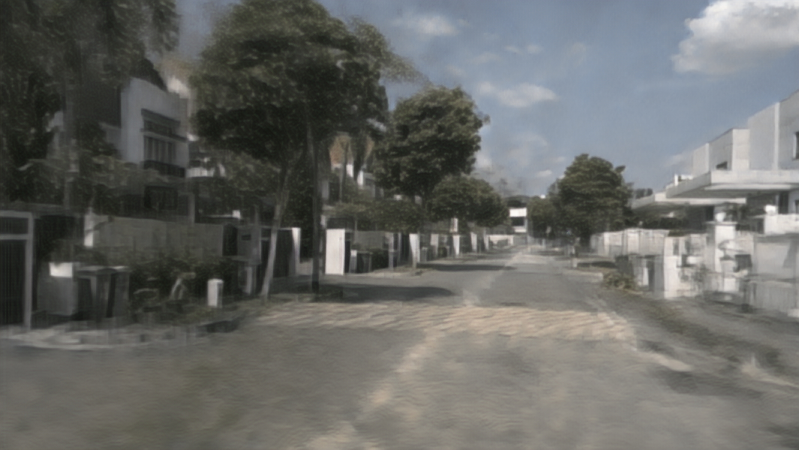}
     & \includegraphics[width=\mywidth]{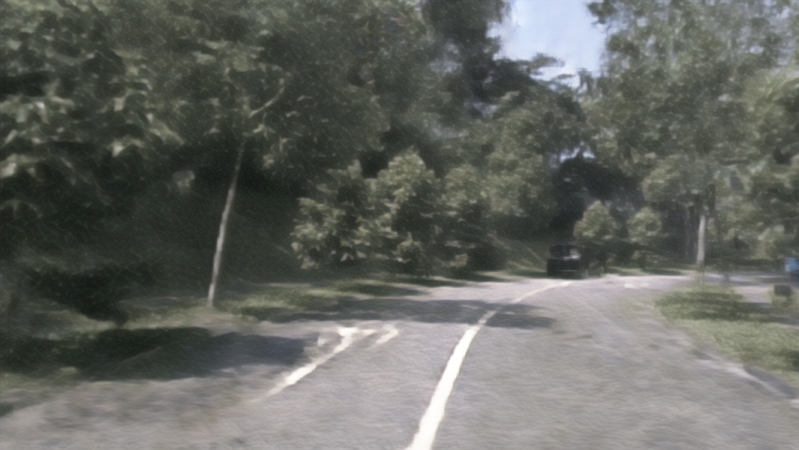} \\
     
     \rotatebox{90}{Ours} & \includegraphics[width=\mywidth]{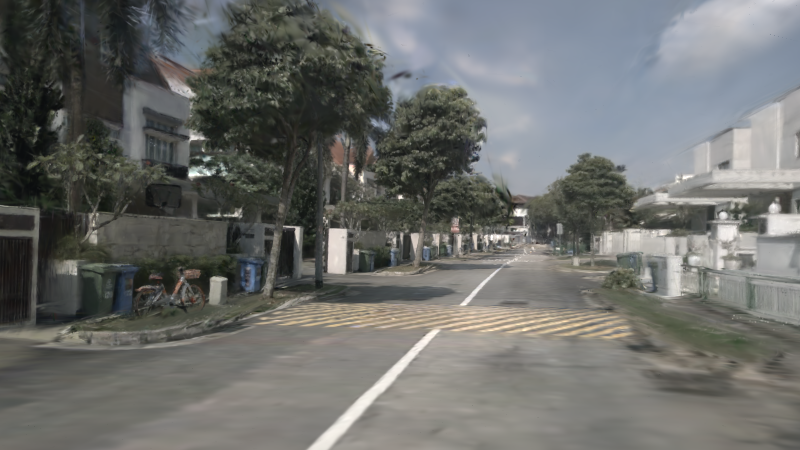}
     & \includegraphics[width=\mywidth]{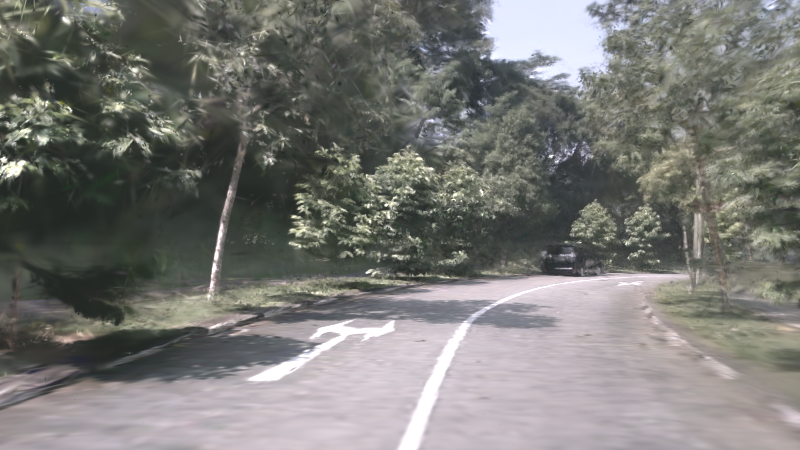} \\
     \end{tabular}
     \vspace{-0.3cm}
     \caption{\textbf{Extrapolated Views Qualitative Comparison} with NeuRAD \cite{yan2024street} on nuScenes.} 
     \vspace{-0.1cm}
\label{fig:compare_nusc}
\end{figure}

%% file: tables/comp_rogs.tex
\begin{table}
\centering
\small
\setlength{\tabcolsep}{2.1pt}
\begin{tabular}{@{\extracolsep{2pt}}lccccc@{}} 
\toprule
& PSNR$\uparrow$ & SSIM$\uparrow$  & LPIPS$\downarrow$ & $\text{KID}_{extrap}$ $\downarrow$ & \#GS$\downarrow$ \\
\cline{2-6}
ROGS \cite{feng2024rogs}
& 18.70 & 0.795 & 0.158 & 0.223 & 1.02M \\
Ours
& \textbf{27.44} & \textbf{0.924} & \textbf{0.061} & \textbf{0.196} & \textbf{0.71M}  \\
\bottomrule
\end{tabular}
\vspace{-0.2cm}
\caption{\textbf{Quantative Comparison} with RoGS. Note that we evaluate all the ground areas, which differs from the evaluation in the RoGS paper, where only sparse colored LiDAR points are considered.}
\vspace{-0.5cm}
\label{tab:comp_rogs}
\end{table}

%% file: figures/eval/ext_comparison/comp_rogs.tex
\begin{figure*}[t!]
     \centering
     \small 
     \setlength{\tabcolsep}{0pt}
     \def\mywidth{4.5cm}
     \def\myheight{2.0cm}
     \begin{tabular}{P{0.5cm}P{\mywidth}P{\mywidth}P{\mywidth}P{\mywidth}}
     \rotatebox{90}{RoGS}
     & \includegraphics[width=\mywidth, height=\myheight]{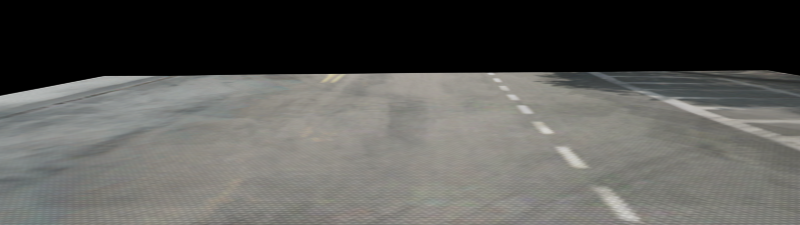}
     & \includegraphics[width=\mywidth, height=\myheight]{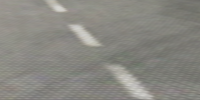}
     & \includegraphics[width=\mywidth, height=\myheight]{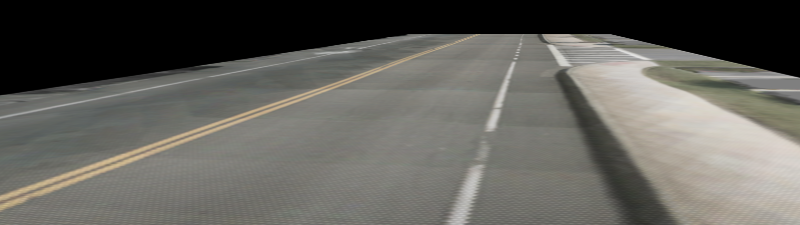}
     & \includegraphics[width=\mywidth, height=\myheight]{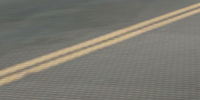} \\
     \rotatebox{90}{Ours}
     & \includegraphics[width=\mywidth, height=\myheight]{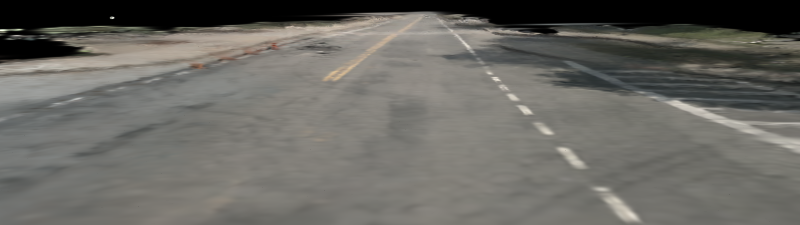}
     & \includegraphics[width=\mywidth, height=\myheight]{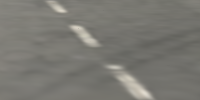}
     & \includegraphics[width=\mywidth, height=\myheight]{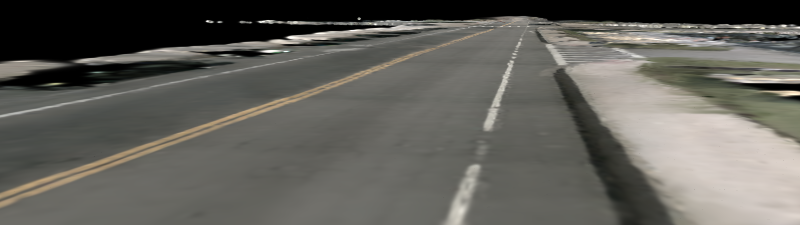}
     & \includegraphics[width=\mywidth, height=\myheight]{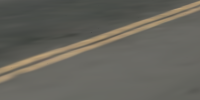} \\
     & Interpolated Views & Interp Zoom In & Extrapolated Views & Extrap Zoom In \\
     \end{tabular}
     \vspace{-0.1cm}
     \caption{\textbf{Qualitative Comparison} with RoGS. Compared to our results, RoGS exhibits grid-style aliasing and limited rendering areas. Both methods show no obvious distortion in extrapolated views.}
\label{fig:comp_rogs}
\end{figure*}

%% file: tables/ablation/ablation_dynamic.tex
\begin{table*}
\centering
\small
\setlength{\tabcolsep}{1.2pt}
\begin{tabular}{@{\extracolsep{2pt}}lccccccccccccccc@{}} 
\toprule
\multicolumn{1}{c}{}                & \multicolumn{5}{c}{KITTI (5\% noise)} & \multicolumn{5}{c}{KITTI (10\% noise)} & \multicolumn{5}{c}{KITTI (20\% noise)}                             \\
& PSNR$\uparrow$  & SSIM$\uparrow$  & LPIPS$\downarrow$    & $e_\bR\downarrow$ & $e_\bt\downarrow$   & PSNR$\uparrow$  & SSIM$\uparrow$  & LPIPS$\downarrow$    & $e_\bR\downarrow$ & $e_\bt\downarrow$    &PSNR$\uparrow$  & SSIM$\uparrow$  & LPIPS$\downarrow$ & $e_\bR\downarrow$ & $e_\bt\downarrow$ \\
\cline{2-6} \cline{7-11} \cline{12-16} 
w/o opt., w/o uni.                    
& 23.83 & 0.878 & 0.062 & 0.031 & 0.027 & 22.16 & 0.861 & 0.079 & 0.063 & 0.106 & 20.28 & 0.835 & 0.101 & 0.125 & 0.425 \\
w/ opt.,  w/o uni.
& 24.80 & 0.897 & 0.038 & 0.022 & 0.051 & 22.75 & 0.879 & 0.056 & 0.054 & 0.130 & 20.56 & 0.855 & 0.081 & 0.135 & 0.612 \\
w/ opt.,  w/ uni. (Ours) 
& \textbf{28.78} & \textbf{0.928} & \textbf{0.023} & \textbf{0.017} & \textbf{0.022} & \textbf{26.66} & \textbf{0.908} & \textbf{0.032} & \textbf{0.037} & \textbf{0.035} & \textbf{23.59} & \textbf{0.875} & \textbf{0.061} & \textbf{0.081} & \textbf{0.176} \\
\bottomrule
\end{tabular}
\vspace{-0.2cm}
\caption{\textbf{Ablation Study of Dynamic Scenes} of KITTI.}
\vspace{-0.4cm}
\label{tab:ablation_dynamic}
\end{table*}

%% file: tables/ablation/ablation_static.tex
\begin{table}
\centering
\small
\setlength{\tabcolsep}{2.1pt}
\begin{tabular}{@{\extracolsep{2pt}}lcccc@{}} 
\toprule
& PSNR$\uparrow$  & SSIM$\uparrow$  & LPIPS$\downarrow$ & Depth $\downarrow$  \\
\cline{2-5}
w/o Affine transform
& 24.18 & 0.827 & 0.083 & --\\
w/o $\cL_\bS$ 
& 24.47 & 0.831 & 0.081 & 0.892 \\
Ours 
& \textbf{24.52} & \textbf{0.833} & 0.081 & \textbf{0.872} \\
\bottomrule
\end{tabular}
\vspace{-0.2cm}
\caption{\textbf{Ablation Study of Static Scenes} on KITTI-360.}
\vspace{-0.2cm}
\label{tab:ablation_static}
\end{table}

%% file: tables/ablation/ablation_ground.tex
\begin{table}
\centering
\small
\setlength{\tabcolsep}{2.1pt}
\begin{tabular}{@{\extracolsep{2pt}}lccccc@{}} 
\toprule
& PSNR$\uparrow$ & SSIM$\uparrow$  & LPIPS$\downarrow$ & $\text{KID}_{interp}$ $\downarrow$ & $\text{KID}_{extrap}$ $\downarrow$ \\
\cline{2-6}
- $\cL_{ground}$ & \textbf{28.03} & \textbf{0.928} & \textbf{0.059} & 0.041 & 0.249 \\
- lr $s_x,s_z$ & 25.84 & 0.909 & 0.079 & 0.050 & 0.266 \\
- lr $\mu$ & 26.91 & 0.916 & 0.065 & \textbf{0.039} & 0.242 \\
Ours & 27.44 & 0.924 & 0.061 & \textbf{0.039} & \textbf{0.196}  \\
\bottomrule
\end{tabular}
\vspace{-0.2cm}
\caption{\textbf{Quantitative Ablation of Ground Model.} While the model without $\cL_{ground}$ demonstrates better metrics in interpolated views, noticeable artifacts are rendered in extrapolated views.}
\vspace{-0.2cm}
\label{tab:ablation_ground}
\end{table}

%% file: figures/eval/ablation/unicycle/unicycle.tex
\begin{figure}[t!]
     \centering
     \small 
     \setlength{\tabcolsep}{0pt}
     \def\mywidth{2.85cm}
     \begin{tabular}{P{\mywidth}P{\mywidth}P{\mywidth}}
     \includegraphics[width=\mywidth]{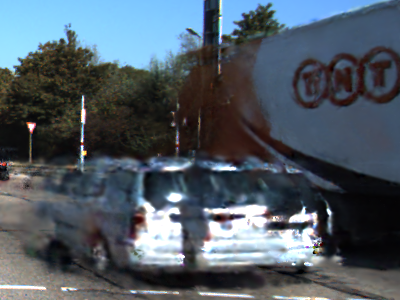}  & 
     \includegraphics[width=\mywidth]{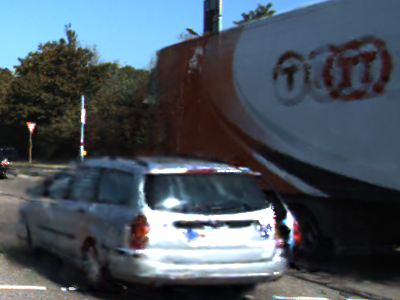} &
     \includegraphics[width=\mywidth]{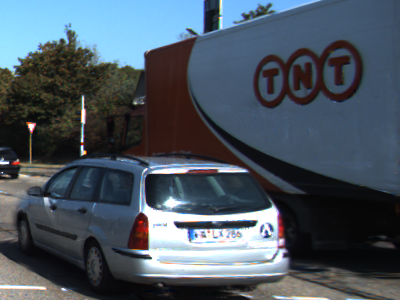} \\
     \includegraphics[width=\mywidth]{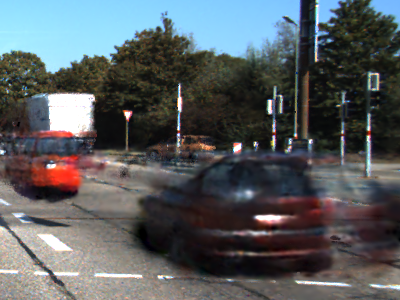}  & 
     \includegraphics[width=\mywidth]{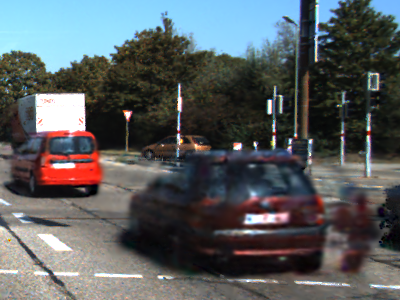} &
     \includegraphics[width=\mywidth]{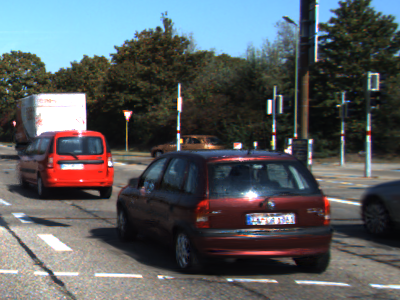} \\
     \rotatebox{0}{w/o opt., w/o uni.} & \rotatebox{0}{w/ opt., w/o uni.} & \rotatebox{0}{Ours} \\
     \end{tabular}
     \vspace{-0.1cm}
     \caption{\textbf{Detail Qualitative Comparison} on KITTI with Noisy Bounding Boxes. }
     \vspace{-0.2cm}
\label{fig:comp_noisy}
\end{figure}

%% file: figures/eval/ablation/ground/ground.tex
\begin{figure}[t!]
     \centering
     \small 
     \setlength{\tabcolsep}{0pt}
     \def\mywidth{4.2cm}
     \def\myheight{1.6cm}
     \begin{tabular}{P{0.5cm}P{\mywidth}P{\mywidth}}
     \rotatebox{90}{-$\cL_{ground}$}
     & \includegraphics[width=\mywidth, height=\myheight]{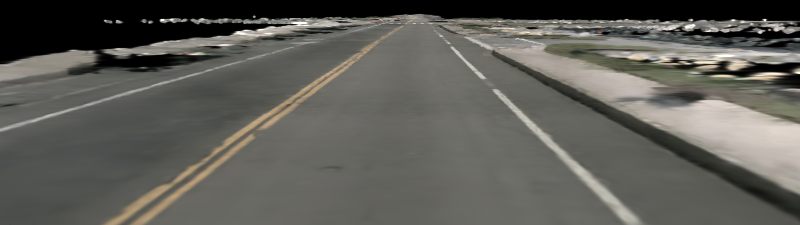}
     & \includegraphics[width=\mywidth, height=\myheight]{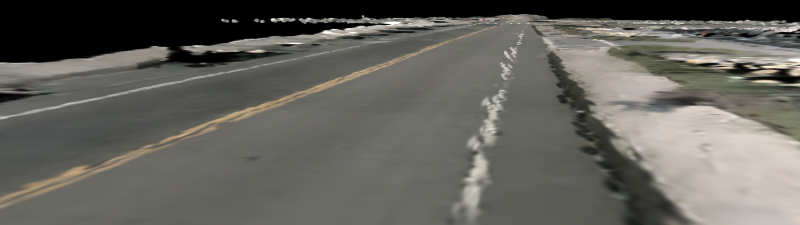} \\
     \rotatebox{90}{-lr $s_x,s_z$}
     & \includegraphics[width=\mywidth, height=\myheight]{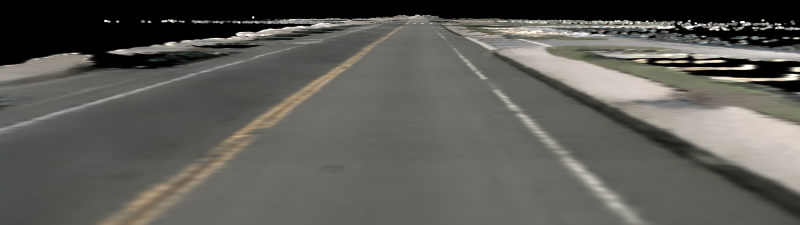}
     & \includegraphics[width=\mywidth, height=\myheight]{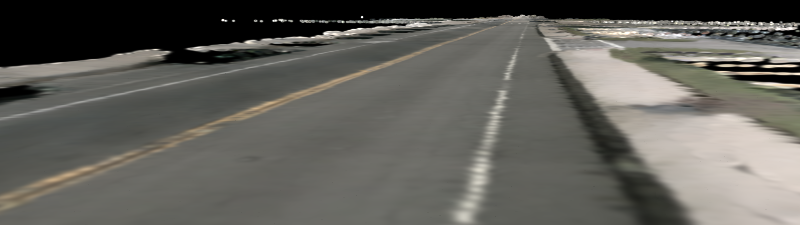} \\
     \rotatebox{90}{-lr $\mu$}
     & \includegraphics[width=\mywidth, height=\myheight]{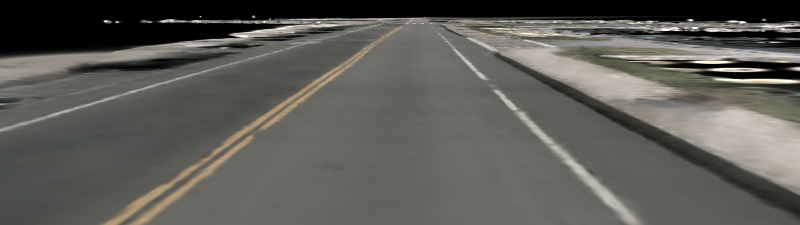}
     & \includegraphics[width=\mywidth, height=\myheight]{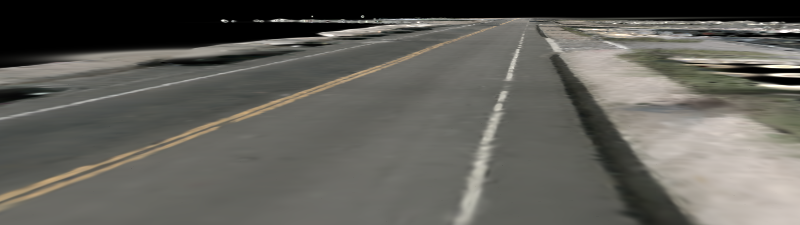} \\
     \rotatebox{90}{Ours}
     & \includegraphics[width=\mywidth, height=\myheight]{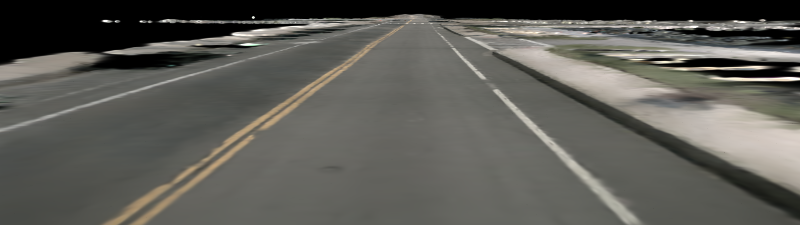}
     & \includegraphics[width=\mywidth, height=\myheight]{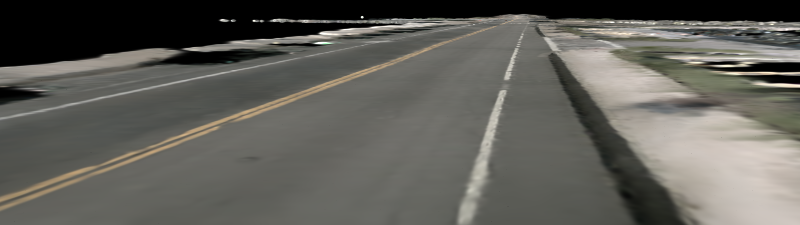} \\
     {} & \rotatebox{0}{Interpolated} & \rotatebox{0}{Extrapolated} \\
     \end{tabular}
     \vspace{-0.1cm}
     \caption{\textbf{Qualitatively Ablation of Ground Model}}
     \vspace{-0.2cm}
\label{fig:ablation_ground}
\end{figure}

%% file: sec_benchmark.tex
\section{Closed-Loop Benchmark}\label{sec:benchmark}
In this section, we first establish the HUGSIM benchmark, comprising a variety of scenes and scenarios with varying difficulty levels for evaluating AD algorithms. Next, we use the HUGSIM benchmark to evaluate several existing AD algorithms, providing a fair, closed-loop evaluation.
Furthermore, we provide a brief case study on the failure cases of existing AD algorithms.

\subsection{HUGSIM Benchmark}

We establish the HUGSIM benchmark based on KITTI-360 \cite{liao2022kitti}, Waymo \cite{Sun2020CVPR}, nuScenes \cite{caesar2020nuscenes} and PandaSet \cite{xiao2021pandaset}. Our benchmark contains more than 70 scenes and over 400 scenarios with varying difficulty levels for testing AD algorithms. 

\boldparagraph{Scenes}
The HUGSIM benchmark consists of scenes from KITTI-360, Waymo, nuScenes, and PandaSet. These scenes feature a variety of street layouts, lighting conditions, and imaging styles.

\input{figures/eval/ablation/insert/insert}

As AD algorithms like UniAD \cite{hu2023planning}, VAD \cite{jiang2023vad} and Latent-Transfuser (LTF) \cite{Chitta2023PAMI} demand for a relative large field on view, KITTI \cite{geiger2012we} is not included in HUGSIM, as it only provides two front-facing cameras. For scene reconstruction, we use the two front-facing perspective cameras and two side fisheye cameras converted to perspective \cite{mei2007single} in KITTI-360, the three front perspective cameras in Waymo, and all six perspective cameras in both nuScenes and PandaSet.

For our simulator rendering, we use the same camera configuration as nuScenes. For scenes from nuScenes and PandaSet, we provide rendered images from all six cameras as observations to the AD algorithms. For scenes from KITTI-360 and Waymo, we provide renderings from only the front three cameras, as these two datasets primarily focus on front and side views. Our experiments show that although UniAD and VAD are trained on a setup with six surrounding cameras, they exhibit robust generalizability even when only the front three camera views are provided.

\boldparagraph{Scenarios}
For each reconstructed scene, we design scenarios across various difficulty levels: easy, medium, hard and extreme. Each reconstructed scene consists of more than four scenarios, as each difficulty level may contain multiple scenarios. Altogether, the HUGSIM benchmark comprises over 400 scenarios, offering a diverse set of training and testing environments for AD algorithms with interactive actors.

Easy scenarios feature mostly static scenes or only include replayed driving actors modeled with a unicycle model. Medium scenarios include IDM and constant-speed driving actors with typical driving behaviors, where IDM actors yield to the ego vehicle. IDM is implemented only in nuScenes due to its HD map availability. Hard and extreme scenarios introduce aggressive actors that attempt to collide with the ego vehicle, with extreme scenarios increasing both the number of aggressive actors, a higher frequency of re-planned attack routes and a greater likelihood of selecting the most aggressive trajectories. We provide details of the scenario designs in the appendix.

\input{figures/benchmark/eval/evaluation}
\subsection{Evaluation on HUGSIM Benchmark}

\boldparagraph{Evaluated AD Algorithms}
The HUGSIM benchmark is well-suited for testing AD algorithms that require only RGB images and ego vehicle status as inputs. In this paper, we evaluate UniAD \cite{hu2023planning}, VAD \cite{jiang2023vad} and Latent-Transfuser (LTF) \cite{Chitta2023PAMI}. LTF is the LiDAR free version of Transfuser, which replaces the LiDAR input as a template positional encoding. It is worth noting that only LTF requires data from the front three cameras, while UniAD and VAD demands for input from six surrounding cameras. For all these three methods, we implement evaluation APIs that interface with HUGSIM Gymnasium environments \cite{brockman2016openai}. These APIs receive data, parse it into the required format for model inference, and send back the planned future waypoints to HUGSIM.

\input{figures/benchmark/case_study/case_study}

\boldparagraph{Closed-Loop Evaluation Results}
The evaluation results on our HUGSIM benchmark are presented in \figref{fig:eval_benchmark}. Detailed evaluation results can be found in the appendix. 
UniAD and LTF demonstrate similar performance on the HUGSIM benchmark. While LTF performs better at easier scenarios, UniAD exhibits more powerful capabilities in handling most of the complex scenarios. We attribute this advantage of UniAD to its trajectory post-processing strategy based on predicted occupancy. However, this post-processing strategy also introduces non-smooth trajectories, negatively impacting the $COM$ score.
Another notable point is that in KITTI-360 and Waymo scenes, only the front three camera views are provided, while UniAD is trained using surrounding six views. The similar performance on Waymo and nuScenes indicates the generalizability of UniAD.
VAD shows less satisfying performance compared to UniAD and LTF on the HUGSIM benchmark, particularly on KITTI-360, Waymo, and PandaSet. This could potentially be due to VAD overfitting to the scene and scenario styles in nuScenes, making it challenging to handle out-of-domain observations in our closed-loop simulator.
All these algorithms face challenges with KITTI-360 scenes due to narrow streets and numerous parked vehicles, along with an imaging style that differs significantly from nuScenes and nuPlan, which are training datasets of these algorithms.
Furthermore, the HUGSIM benchmark poses significant challenges for existing AD algorithms. Even the easy level, which consists of original scenes, can be difficult for these algorithms, highlighting the lack of generalization and the need for systematic analysis using a closed-loop simulator as presented in this paper. The greater challenges presented at the medium, hard, and extreme levels indicate that substantial efforts are still required to achieve the goal of full autonomy.

\boldparagraph{Case Study}
We outline some common failure cases encountered by AD algorithms in our closed-loop simulator as shown in \figref{fig:case_study}: (a) Since collisions never occur in open-loop datasets, models rarely encounter situations where there is no drivable area in front of the vehicle. As a result, they consistently predict a drivable area ahead, regardless of the RGB observations. (b) AD algorithms face difficulties with turning, often steering at inappropriate angles relative to the street structure. (c) Although a leading vehicle is detected, AD algorithms often do not take evasive action in advance; instead, they attempt to make a sharp turn at a very close distance, which can easily result in a collision. (d) The planned trajectory of AD algorithms can be unstable, which may lead to collisions in narrow streets and a reduction in $COM$ score.

%% file: figures/eval/ablation/insert/insert.tex
\begin{figure}[t!]
     \centering
     \small 
     \setlength{\tabcolsep}{0pt}
     \def\mywidth{4.4cm}
     \begin{tabular}{P{\mywidth}P{\mywidth}}
     \includegraphics[width=\mywidth]{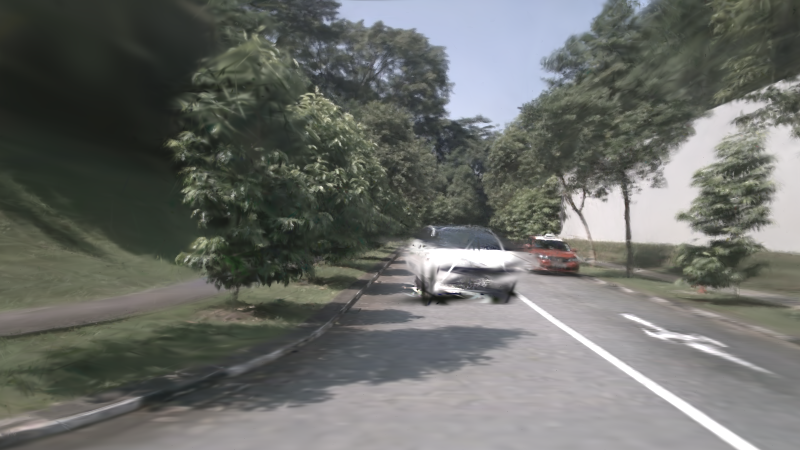}
     & \includegraphics[width=\mywidth]{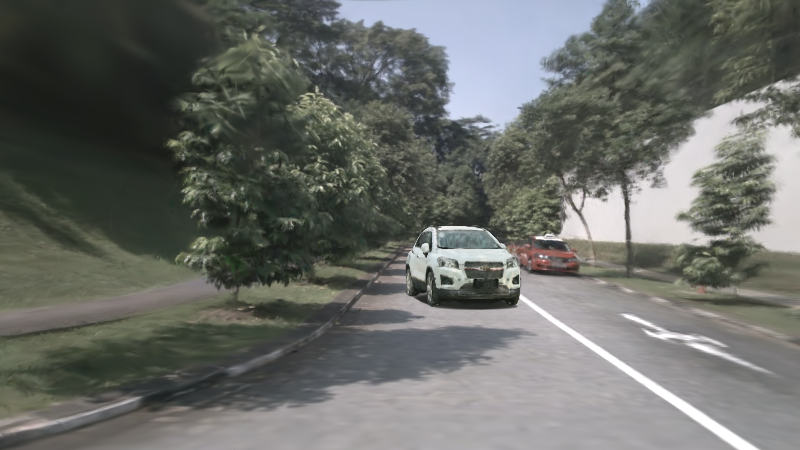} \\
     \includegraphics[width=\mywidth]{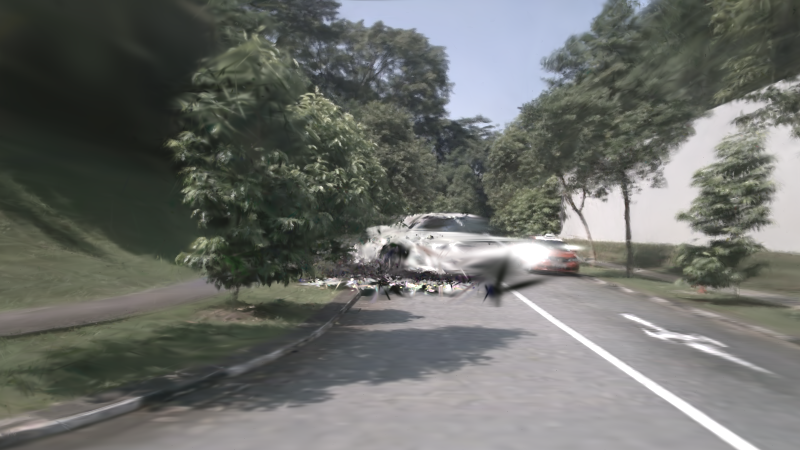}
     & \includegraphics[width=\mywidth]{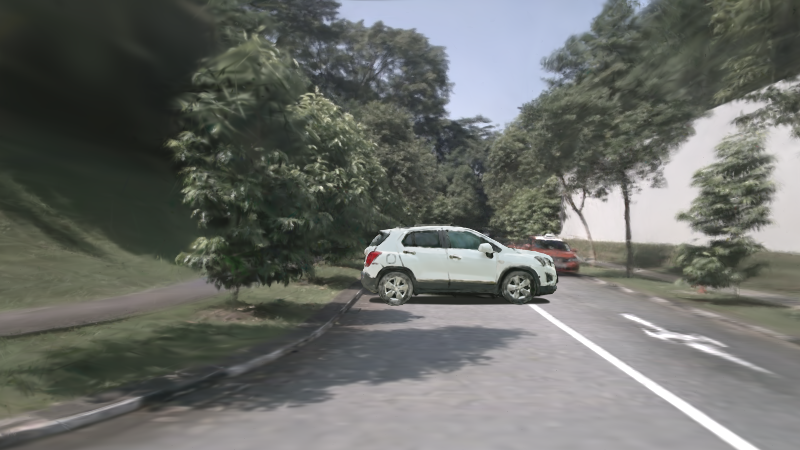} \\
     \rotatebox{0}{Native} & \rotatebox{0}{Non-Native} \\
     \end{tabular}
     \vspace{-0.1cm}
     \caption{\textbf{Qualitative Comparison} with native and non-native vehicles insertion.}
     \vspace{-0.4cm}
\label{fig:ablation_insert}
\end{figure}

%% file: figures/benchmark/eval/evaluation.tex
\begin{figure*}
    \centering
    \includegraphics[width=\textwidth]{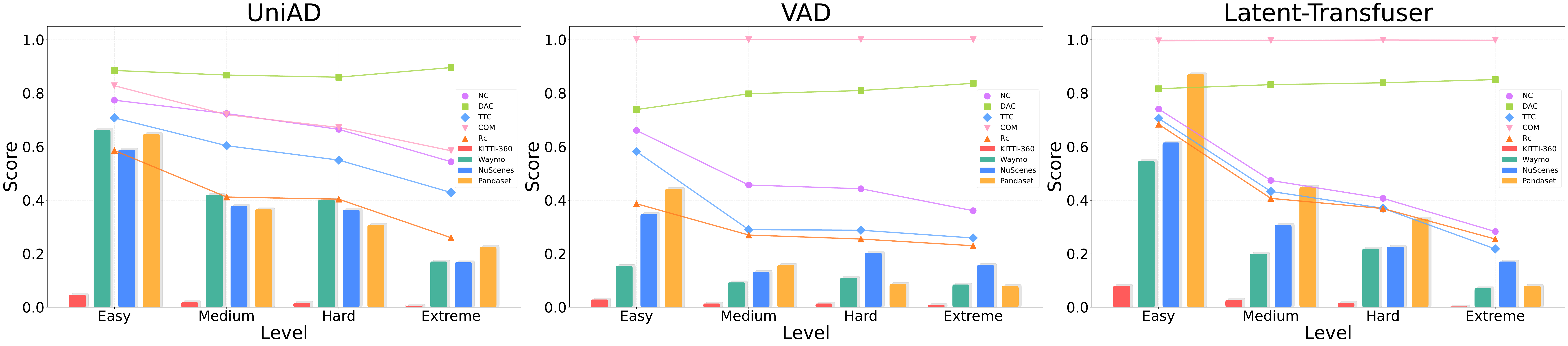}
    \vspace{-0.6cm}
    \caption{\textbf{Evaluation Results on HUGSIM Benchmark}. The histograms indicate performance of models on each type of dataset, while the lines represent the averaged sub-scores across datasets.}
\label{fig:eval_benchmark}
\vspace{-0.4cm}
\end{figure*} 

%% file: figures/benchmark/case_study/case_study.tex
\begin{figure}[t!]
     \centering
     \small 
     \setlength{\tabcolsep}{0pt}
     \def\mywidth{7.8cm}
     \begin{tabular}{P{0.8cm}P{\mywidth}}
     \rotatebox{0}{(a)} & 
     \includegraphics[width=\mywidth]{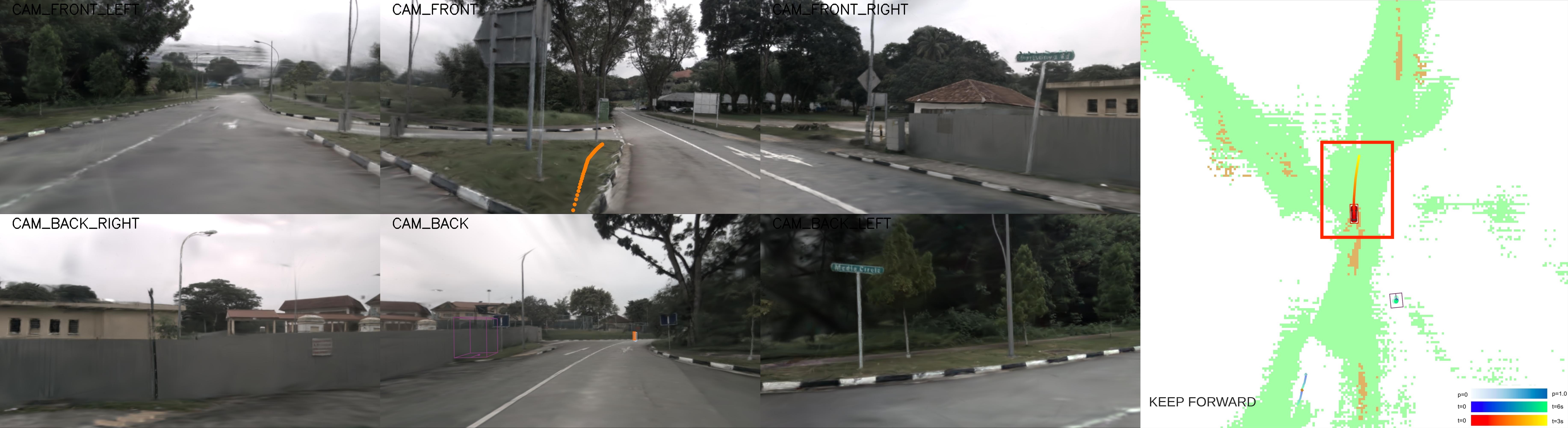} \\
     \rotatebox{0}{(b)} & 
     \includegraphics[width=\mywidth]{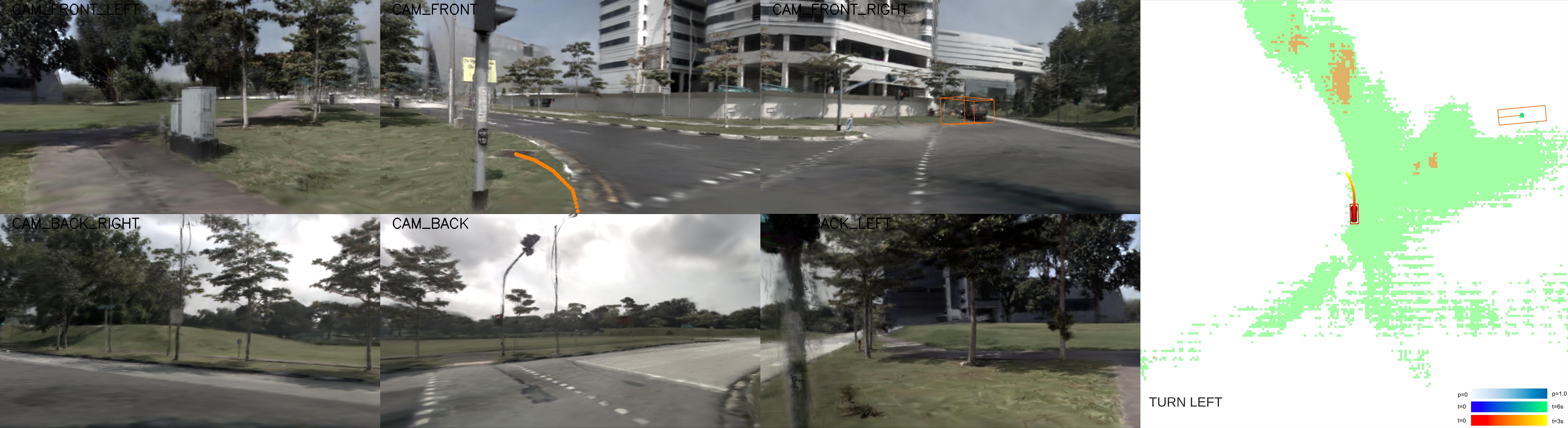} \\
     \rotatebox{0}{(c)} & 
     \includegraphics[width=\mywidth]{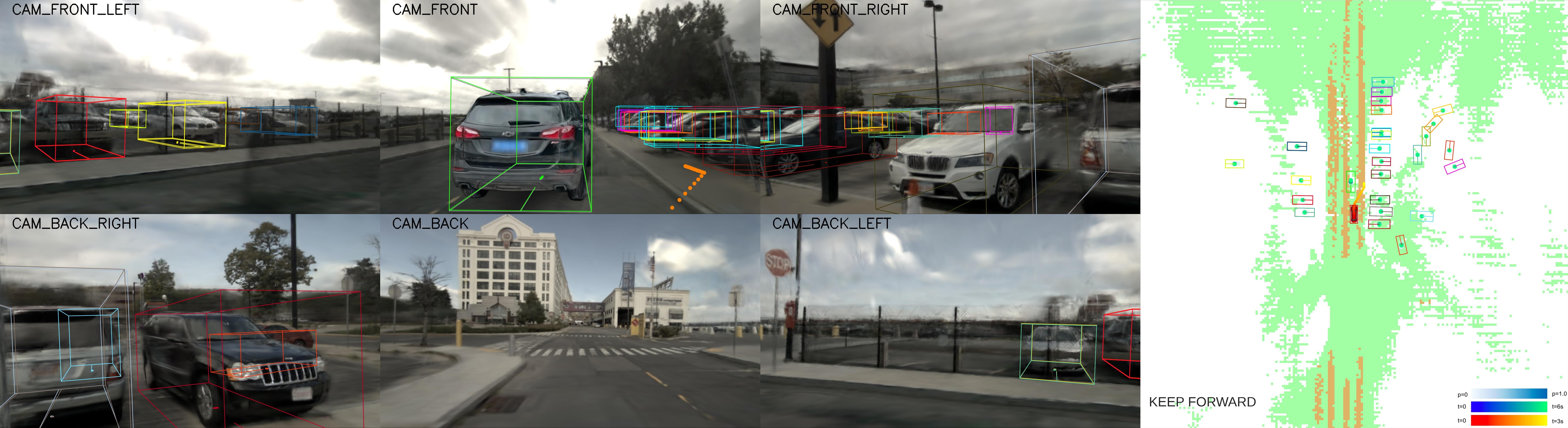} \\
     \rotatebox{0}{(d)} &
     \includegraphics[width=\mywidth]{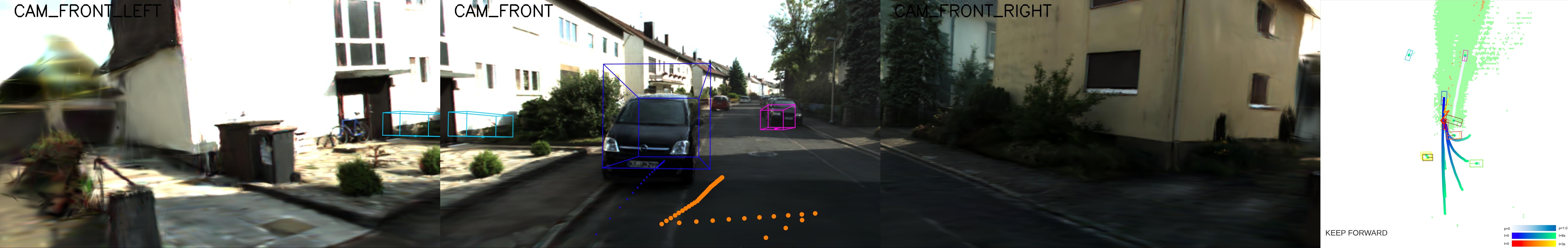} \\
     \end{tabular}
     \vspace{-0.2cm}
     \caption{\textbf{Case Study}. The left part shows the RGB observations, perception results, and planning trajectories, the right part displays the predicted BEV information of UniAD.}
 \vspace{-0.4cm}
\label{fig:case_study}
\end{figure}

%% file: sec_conclusion.tex
\section{Conclusion}

We present HUGSIM, a novel photorealistic closed-loop simulator for autonomous driving, featuring real-time, high-quality rendering in extrapolated views and efficiently generated actor behavior. Specifically, we reconstruct urban scenes using 3D Gaussians and introduce a ground model, along with single-vehicle reconstruction, to improve the rendering quality of extrapolated views. For actor behavior, we propose an attack-cost-based trajectory interactively search to simulate aggressive driving behaviors of actors.

Furthermore, we establish the HUGSIM benchmark across multiple datasets including variance sequences, designing more than 300 scenarios for evaluating and training AD algorithms. We evaluate several baselines on our benchmark. Our results show that the HUGSIM benchmark presents significant challenges for existing AD algorithms. This closed-loop benchmark reveals substantial room for improvement in autonomous driving performance.
We hope that our dataset and benchmark will fertilize new research across communities, fostering progress towards the ultimate goal of full autonomy.

For future work, HUGSIM can be enhanced in several ways. First, we assume all dynamic objects follow rigid motion, which may cause blurring for non-rigidly moving objects, such as pedestrians. This can be addressed by incorporating non-rigid dynamic reconstruction methods~\cite{yang2024deformable, chen2024omnire} into our framework. While our method improves rendering at extrapolated viewpoints, it struggles with high-fidelity rendering at views that are far from the input or very close to objects. These challenges could be mitigated by leveraging priors from 2D generative models~\cite{wang2024freevs, yu2024viewcrafter}. Furthermore, since our approach opens up the possibility for fine-tuning AD algorithms in a photorealistic closed-loop setting, this presents a promising avenue for future exploration.

\ifCLASSOPTIONcompsoc
  \section*{Acknowledgments}
\else
  \section*{Acknowledgment}
\fi

The authors thank Kashyap Chitta for providing details of Transfuser and NAVSIM, and Chaojie Ji, Yuanbo Yang, Xianxu Xiang, and Jiahao Shao for proofreading. This work is supported by NSFC under grant 62202418, U21B2004 and the National Key R\&D Program of China under Grant 2021ZD0114501. Yiyi Liao is with the Zhejiang Provincial Key Laboratory of Information Processing, Communication and Networking (IPCAN). Andreas Geiger was supported by the ERC Starting Grant LEGO-3D (850533) and the DFG EXC number 2064/1 - project number 390727645.

%% file: sec_appendix.tex
\clearpage
\begin{appendices}

\section{Implementation}
\label{sec:implementation}
In this section, we begin by discussing our 3D Gaussian details, encompassing semantic, opacity and depth implementation (\ref{sec: 3d_gs}). Subsequently, we discuss the difference between 3D softmax and 2D softmax in 3D Semantic Scene Reconstruction (\ref{sec: 3d_semantic}). Finally, we elucidate the evaluation metrics we utilize (\ref{sec: metrics}).

\subsection{3D Gaussian Details}
\label{sec: 3d_gs}

Following \cite{kerbl20233d}, each Gaussian has the following attributes: rotation ($\bR_g \in \nR^{3\times3}$), scale ($\bS_g \in \nR^{3\times1}$), opacity ($\alpha$) and spherical harmonics ($SH$). The corresponding 3D covariance matrix $\bSigma \in \nR^{3\times3}$ can be calculated using the following formula:
\begin{equation}
\bSigma = \bR_g\bS_g\bS_g^T\bR_g^T
\end{equation}
When provided with a viewing transformation $\bW \in \nR^{3\times3}$ and the Jacobian of the affine approximation of the projective transformation $\bJ \in \nR^{3\times3}$, the covariance matrix $\bSigma^{\prime} \in \nR^{3\times3}$ in camera coordinates can be expressed as:
\begin{equation}
\bSigma^{\prime} = \bJ\bW\bSigma \bW^T\bJ^T
\end{equation}
Following EWA splatting \cite{zwicker2002ewa}, we can skip the third row and column of $\bSigma^{\prime}$ to obtain a $2\times2$ covariance matrix with the same structure and properties. For brevity, we still use the notation $\bSigma^{\prime} \in \nR^{2 \times 2}$ to denote the 2D covariance matrix.

By considering the projected 3D Gaussian center $\bmu \in \nR^{2\times1}$ and an arbitrary point $\bx \in \nR^{2\times1}$ on camera coordinates, the opacity $\alpha^{\prime}$ of $\bx$ contributed by this 3D Gaussian can be computed as follows:
\begin{equation}
\alpha^{\prime} = \alpha \exp \left( -\frac{1}{2}(\bx-\bmu)^T (\bSigma^{\prime})^{-1} (\bx-\bmu)\right)
\end{equation}
The color $\bc$ of each Gaussian can be computed based on the view direction and its corresponding spherical harmonics ($SH$). Given a set of sorted 3D Gaussians $\cN$ along the ray, we obtain the accumulated color via volume rendering:
\begin{equation}
\pi: \quad \bC = \sum_{i \in \mathcal{N}} \bc_i \alpha'_i \prod_{j=1}^{i-1}(1-\alpha'_j)
\end{equation}
The same volume rendering technique can be applied to obtain semantic $\bS$, depth $\bD$ and optical flow $\bF$. With the given semantic feature $\bs_i$, depth value $d_i$, and Gaussian motion $\mathbf{f}_i$ relative to the camera pose, we can define the semantic rendering, depth rendering, and flow rendering as follows:
\begin{align}
    \label{eq:3d_smt_render}
    \quad \bS &= \sum_{i \in \mathcal{N}} \text{softmax}(\bs_i) \alpha'_i \prod_{j=1}^{i-1}(1-\alpha'_j) \\ 
    \quad \bD &= \sum_{i \in \mathcal{N}} d_i \alpha'_i \prod_{j=1}^{i-1}(1-\alpha'_j)  \\ 
    \quad \bF &= \sum_{i \in \mathcal{N}} \mathbf{f}_i \alpha'_i \prod_{j=1}^{i-1}(1-\alpha'_j)
\end{align}
Note that all the projections and volume rendering techniques mentioned are implemented in CUDA. Calculating the projected 2D opacity $\alpha^{\prime}$ on each pixel and sorting Gaussians based on their distances from the camera takes the majority of computations in the rendering process. These computations need to be performed only once for rendering all modalities, thus maintaining the real-time rendering property of the original 3D Gaussian Splatting.

\subsection{3D Semantic Scene Reconstruction}
\label{sec: 3d_semantic}
We utilize \eqref{eq:3d_smt_render}, referred to as 3D softmax, to render semantic maps. This is in contrast to most existing NeRF-based semantic reconstruction methods that perform softmax to the accumulated 2D logits~\cite{fu2022panoptic
, zhi2021place}, described in \eqref{eq:2d_smt_render}, referred to as 2D softmax. The fundamental difference between these two rendering techniques lies in the fact that 3D softmax normalizes the logits of each 3D point. This normalization process helps prevent a single point with a significantly high logit value from imposing an overwhelming influence on the overall volume rendering outcome. On the other hand, it also prevents placing 3D points of low logit values in empty space. As a result, 3D softmax is effective in reducing floaters and enhancing the geometry of the reconstruction results. In \eqref{sec:softmax_ablation}, we present a comprehensive analysis of the qualitative and quantitative comparison results between these two rendering methods.

\begin{equation}
 \quad \bS_{\text{2D\_norm}} = \text{softmax} \left(\sum_{i \in \mathcal{N}} \bs_i \alpha'_i \prod_{j=1}^{i-1}(1-\alpha'_j)\right)
\label{eq:2d_smt_render}
\end{equation}
In the following sections, we refer to our default setting obtained by \eqref{eq:3d_smt_render} as $\bS_{\text{3D\_norm}}$.

\input{supplement/figures/scenarios/scenarios}

\subsection{Metrics}
\label{sec: metrics}

\boldparagraph{Novel View Appearance Synthesis}To assess the quality of novel view appearance synthesis, we utilize the Peak Signal-to-Noise Ratio (PSNR), Structural Similarity Index (SSIM), and Learned Perceptual Image Patch Similarity (LPIPS)~\cite{zhang2018unreasonable} following the common practice. 

\boldparagraph{Novel View Semantic Synthesis} Following KITTI-360~\cite{liao2022kitti}, we evaluate the quality of novel view semantic synthesis via the mean Intersection over Union (mIoU) metric.

\boldparagraph{3D Semantic Reconstruction} 
We evaluate 3D semantic reconstruction quality by extracting a 3D semantic point cloud and comparing it with the ground truth LiDAR points. We evaluate both geometric and semantic metrics in the 3D space. Specifically, we evaluate geometric reconstruction quality by measuring the accuracy ($acc.$) and completeness ($comp.$). Accuracy measures the average distance from reconstructed points to the nearest LiDAR point, while completeness measures the average distance from LiDAR points to the nearest reconstructed points. In order to measure the semantic quality of the reconstructed point cloud, we map the predicted 3D semantics to the LiDAR points. Concretely, for each point in the LiDAR point cloud, we identify its closest counterpart in the predicted semantic point cloud and allocate a semantic label based on this nearest neighbor. The assigned semantic labels of all LiDAR points are then compared with the 3D semantic segmentation ground truth provided by KITTI-360, evaluated via the mIoU metric. Note that we only use the LiDAR point clouds for evaluation.

\boldparagraph{3D Tracking} To demonstrate the effectiveness of our model in rectifying noisy 3D tracking results, we evaluate the accuracy of predicted poses compared to ground truth poses in our ablation study. 
Considering the rotation and translation parameters of a ground truth bounding box denoted as $\hat{\bR}$ and $\hat{\bt}$, respectively, and the corresponding parameters of predicted poses, represented as $\bR$ and $\bt$, we employ two metrics for this evaluation following \cite{chen2020category}: $e_\bR$ quantifies the rotation accuracy, while $e_\bt$ assesses the translation accuracy as follows
\begin{align}
e_\bR &= \arccos{\frac{Tr(\hat{\bR} \cdot \bR^{-1}) - 1} {2}} \\
e_\bt &= \Vert \hat{\bt} - \bt \Vert_2
\end{align}
where $Tr$ represents the trace of a matrix. 

\boldparagraph{Depth Estimation} In our ablation study, we evaluate the depth estimation quality of our different variants. This is achieved by first projecting the LiDAR points acquired at the same frame to the 2D image space, followed by measuring the L2 distance between the projected LiDAR depth and our method. Considering the projected LiDAR depth is sparse, our assessment focuses solely on pixels with valid LiDAR projections when calculating the L2 distance.

\input{supplement/tabs/speed_breakdown}
\input{supplement/tabs/comparison_dynamic_gtpose}

\input{supplement/figures/softmax_ablation/softmax_ablation}

\input{supplement/figures/nerfacto/nerfacto}

\section{Data}
\label{sec:data}
In this section, we present details of datasets on which we conducted our experiments, including KITTI \cite{geiger2012we}, Virtual KITTI 2 (vKITTI) \cite{cabon2020virtual}, KITTI-360 \cite{liao2022kitti}, Waymo \cite{Sun2020CVPR}, nuScenes \cite{caesar2020nuscenes} and PandaSet \cite{xiao2021pandaset}.

\boldparagraph{KITTI}
Following NSG \cite{ost2021neural} and MARS \cite{wu2023mars}, we select frames 140 to 224 from Scene02 and frames 65 to 120 from Scene06 on KITTI for conducting our experiments.

\boldparagraph{vKITTI}
Virtual KITTI 2 is a synthetic dataset that closely resembles the scenes present in KITTI. In line with the settings outlined in NSG and MARS, we conduct experiments on exactly the same frames from Scene02 and Scene06.

\boldparagraph{KITTI-360}
We perform experiments on KITTI-360, encompassing both static and dynamic scenes. For the tasks of novel view synthesis and novel semantic synthesis on the leaderboard, we conduct experiments on the sequences provided by the official dataset. We also explore dynamic scenes, such as frames 11322 to 11381 from sequence 00, as showcased in our teaser. Additionally, We construct our closed-loop benchmark based on KITTI-360, using both the front two stereo cameras and the side two fisheye cameras. We convert images from the fisheye cameras to perspective images using the model of \cite{mei2007single} as the projection model.

\input{supplement/figures/optimize/optimize}

\input{supplement/tabs/compare_qd3dt}

\begin{figure}[t]
\centering
\includegraphics[width=0.55\linewidth]{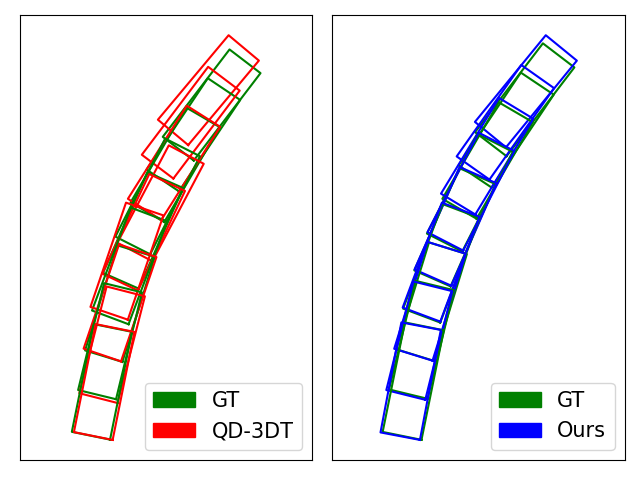} \\
\vspace{-0.3cm}
\captionof{figure}{\textbf{Pose comparison} with QD-3DT.}
\label{fig:traj}
\vspace{-0.2cm}
\end{figure}

\boldparagraph{Waymo}
We perform experiments on Waymo to compare with StreetGaussian \cite{yan2024street} and NeuRAD \cite{tonderski2024neurad} in extrapolated views. We select all frames from sequences segment-10676267326664322837 and segment-9385013624094020582 for the experiments. Every 5th frame is dropped as the test frame, and use the front, front-left and front-right cameras as inputs. Additionally, we construct HUGSIM based on Waymo using the same settings.

\boldparagraph{Nuscenes}
NuScenes is used to compare our method with StreetGaussian, NeuRAD, and RoGS \cite{feng2024rogs} in extrapolated views. We select all frames from sequences 0051, 0411, and 0655. Every 5th frame is dropped as the test frame, and we use all 6 surrounding cameras as inputs. HUGSIM benchmark is also constructed based on nuScenes. It is important to note that the camera extrinsics of nuScenes are not well refined, so we apply a rigid bundle adjustment to obtain more accurate camera extrinsics.

\boldparagraph{PandaSet}
Scenes from PandaSet are also part of the HUGSIM benchmark. We use all 6 surrounding cameras in PandaSet and drop every 5th frame as the test frame during scene reconstruction training.

\section{HUGSIM Benchmark}
\label{sec:appendix_benchmark}

\subsection{Scenario Design}
We design easy, medium, hard and extreme levels for our HUGSIM benchmark. The easy scenarios consist of static scenes or replayed dynamic vehicle behaviors. The design of scenarios for the medium, hard, and extreme levels is shown in \figref{fig:scenarios}.

\boldparagraph{Medium}
Scenario (a) simulates a stationary vehicle in the front of ego vehicle. Scenario (b) simulates a vehicle driving ahead the ego but may slower than ego vehicle. Scenario (c) simulates a vehicle driving ahead but suddenly stops.

\boldparagraph{Hard}
Scenario (a) simulates a vehicle driving in reverse. Scenario (b) simulates a stationary vehicle ahead and a vehicle reversing in the side lane. Scenario (c) simulates an oncoming aggressive vehicle attempting to attack the ego vehicle.

\boldparagraph{Extreme}
Scenario (a) simulates a situation similar to Hard scenario (c), but with the aggressive vehicle having a larger attack frequency and being more likely to select the most aggressive trajectory. Scenario (b) simulates an aggressive vehicle suddenly changing lanes in front of the ego vehicle. Scenario (c) simulates multiple reverse-driving vehicles.

\subsection{Implementation of Tested AD Algorithms}

We implement inference scripts for UniAD \cite{hu2023planning}, VAD \cite{jiang2023vad}, and LTF \cite{Chitta2023PAMI} to integrate these AD algorithms with HUGSIM. We build communication bridges based on named pipelines in memory, which transfer data at ultra-high speed, with nearly no additional time consumption.  

\boldparagraph{UniAD}
UniAD proposes an end-to-end AD algorithm, enabling flexible intermediate representations and exchanging multi-task knowledge toward planning. The inference code of the official UniAD implementation couples observations, CAN-bus data, and ground-truth information together. We re-implement the inference script, which only requires RGB image observations and sensor poses, based on the official code.

\boldparagraph{VAD}
VAD proposes an end-to-end vectorized paradigm for AD and is developed on the UniAD codebase. We re-implemented the inference script for VAD in a similar way to UniAD.

\boldparagraph{LTF}
Latent TransFuser (LTF) is an image-only version of Transfuser \cite{Chitta2023PAMI}, replacing the LiDAR BEV histogram input with a positional encoding. The original version of LTF is trained on CARLA \cite{dosovitskiy2017carla}, we use the version implemented in NAVSIM \cite{dauner2024navsim} which is trained on nuplan \cite{caesar2021nuplan}, offering better alignment with the real world. 

\input{tables/adtest/ALL}

\section{Baselines}
\label{sec:baseline}
In this section, we discuss the baselines against which we compare our approach, including  NSG~\cite{ost2021neural}, MARS~\cite{wu2023mars}, PNF~\cite{kundu2022panoptic}, Semantic Nerfacto~\cite{tancik2023nerfstudio}, StreetGaussian \cite{yan2024street}, NeuRAD \cite{tonderski2024neurad} and RoGS \cite{feng2024rogs}.

\boldparagraph{NSG}
NSG is the pioneering method that introduces the decomposition of dynamic scenes into static background and dynamic foreground components. They propose a learned scene graph representation that enables efficient rendering of novel scene arrangements and viewpoints. However, the official source code provided by NSG often encounters issues when training on KITTI Scene02. Therefore, we utilize the version implemented by the authors of MARS, which is more stable and yields slightly improved results compared to the original version.

\boldparagraph{MARS}
We utilize the latest version of the code provided by the official MARS repository. This latest version incorporates bug fixes and includes additional training iterations, resulting in improved performance. In fact, the updated version achieves a notable improvement of 3 to 4 dB on PSNR compared to the numbers reported in the original paper.

\boldparagraph{PNF}
Since PNF is not open-source, we directly compare our method to their submission on the KITTI-360 leaderboard regarding novel view appearance \& semantic synthesis. To the best of our knowledge, PNF is the only work that considers the optimization of noisy 3D bounding boxes of dynamic objects. In our ablation study, we conduct a na\"ive baseline that optimizes the 3D bounding boxes of each frame independently, which can be considered as a re-implementation of PNF's bounding box optimization in our framework.

\boldparagraph{Semantic Nerfacto}
For the evaluation of 3D semantic point cloud geometry, we compare our results with Semantic Nerfacto \cite{tancik2023nerfstudio} as an alternative to PNF \cite{kundu2022panoptic}.
Nerfacto \cite{tancik2023nerfstudio} is an integration of several successful methods that demonstrate strong performance on real data. It incorporates camera pose refinement, per-image appearance embedding, proposal sampling, scene contraction, and hash encoding within its pipeline. Additionally, Nerfacto includes a semantic head in its framework, enabling the generation of meaningful semantic maps, as demonstrated in \ref{sec:smt_nerfacto}.

\input{supplement/tabs/ablation_2dsemantic}

\boldparagraph{StreetGaussian}
StreetGaussian is a concurrent work to HUGS \cite{Zhou2024CVPR}, proposing a dynamic urban scene reconstruction algorithm also based on 3D Gaussian Splatting \cite{kerbl20233d}, with primary experiments conducted on Waymo. StreetGaussian requires both LiDAR and RGB data as inputs.

\boldparagraph{NeuRAD}
NeuRAD is also a concurrent work to HUGS, proposing a NeRF-based dynamic urban scene reconstruction method. NeuRAD integrates various strategies, achieving state-of-the-art performance across many AD datasets. Like StreetGaussian, NeuRAD requires both LiDAR and RGB data as inputs. It is worth noting that NeuRAD is the reconstruction rendering method used in NeuroNCAP \cite{ljungbergh2024neuroncap}.

\boldparagraph{RoGS}
RoGS is a concurrent work to HUGSIM, proposing an efficient method tailored to ground reconstruction based on 2D Gaussian Splatting \cite{huang20242d} in nuScenes. It requires only RGB images for ground reconstruction and uniformly tiles Gaussians on the ground plane.

\section{Additional Experiment Results}
\label{sec:exp}

\subsection{Time Consumption Breakdown}
\tabref{tab:breakdown} shows our detailed runtime breakdown as various components are incrementally enabled. Preparation (Pre.) contains operations like tile partition and Gaussian sorting. $\pi$ denotes volume rendering, and affine denotes affine transform. Other components like unicycle model, dynamic decomposition, and depth rendering are excluded as they hardly consume any additional time.

\input{supplement/figures/depth}

\subsection{Additional Comparison Experiments}
\boldparagraph{Dynamic Scene with GT 3D Bounding Boxes}
Despite not being our primary focus, we additionally provide a comparison with NSG and MARS using ground truth 3D trackings. In this setting, our approach demonstrates superior performance across all test scenes, see \tabref{tab:comp_gt}.

\boldparagraph{Details of Comparison with Semantic Nerfacto}
\label{sec:smt_nerfacto}
While Semantic Nerfacto excels at rendering meaningful novel view semantic images (as seen in \figref{fig:compare_nerfacto_2d}), \figref{fig:compare_nerfacto_3d} shows it struggling to accurately reconstruct correct geometry. Following the common practice of NeRF-based semantic reconstruction methods \cite{tancik2023nerfstudio}, we apply 2D softmax to Semantic Nerfacto. when we attempted to apply the 3D Softmax technique to Nerfacto, it did not yield better results compared to using 2D softmax. The results can be attributed to the incorrect of Nerfacto's 3D geometry. Instead of adjusting 2D logits with large-scale logits in 3D, the use of 3D softmax prevents the ''cheating'' approach by normalizing logits in 3D space. However, this normalization requirement necessitates sufficiently accurate geometry for satisfactory results.

\boldparagraph{Comparisons with Tracking Methods}
To further compare with off-the-shelf tracking methods, we show the performance of QD-3DT \cite{hu2022monocular} and our optimized pose initialized with \cite{hu2022monocular} in \tabref{tab:traj} and qualitatively illustrate the poses of one vehicle in \figref{fig:traj}. Our method consistently improves \cite{hu2022monocular} across two KITTI scenes.

\subsection{Additional Ablation Experiments}
\boldparagraph{3D and 2D Semantic Softmax}
\label{sec:softmax_ablation}
We provide more 3D and 2D semantic logits softmax comparison in \figref{fig:softmax_ablation} and \figref{tab:semantic2d_3d}. As can be seen, normalizing semantic logits in 3D space leads to notable qualitative and quantitative improvement compared to 2D space normalization.

\boldparagraph{Improvements on Geometry}
We now qualitatively examine how the semantic loss $\cL_\bS$ impact the geometry, as shown in \figref{fig:depth_comp}. Th figure reveals that incorporating the semantic loss improves the underlying geometry. It's important to note that when the semantic loss $\cL_\bS$ is active, the sky region of the depth maps in \figref{fig:depth_comp} is set to infinite.

\subsection{Visualization of Optimization Progress}
We present the visualization of the optimization progress for both the noisy bounding boxes and the background semantic point cloud in \figref{fig:vis_optim}. 
Using noisy 3D bounding boxes as input, our approach optimizes both the background and the poses of the bounding boxes simultaneously. As evident, the application of physical constraints derived from the unicycle model results in a smooth trajectory for the bounding boxes.

\subsection{HUGSIM Benchmark evaluation results}
We present the results of HD-Score and sub-scores $NC$, $DAC$, $TTC$, $COM$, and $R_c$ for evaluating UniAD \cite{hu2023planning}, VAD \cite{jiang2023vad}, and LTF \cite{Chitta2023PAMI} on our HUGSIM benchmark in \tabref{tab:adtest_all}. These results are identical to those in \figref{fig:eval_benchmark}, but presented in a more detailed format.

\end{appendices}

%% file: supplement/figures/scenarios/scenarios.tex
\begin{figure*}[t!]
     \centering
     \small 
     \setlength{\tabcolsep}{0pt}
     \def\mywidth{3.8cm}
     \def\myheight{3.3cm}
     \setlength{\tabcolsep}{8pt}
     \begin{tabular}{P{0.5cm}P{\mywidth}P{\mywidth}P{\mywidth}}

     \rotatebox{90}{Medium} &
     \includegraphics[width=\mywidth, height=\myheight]{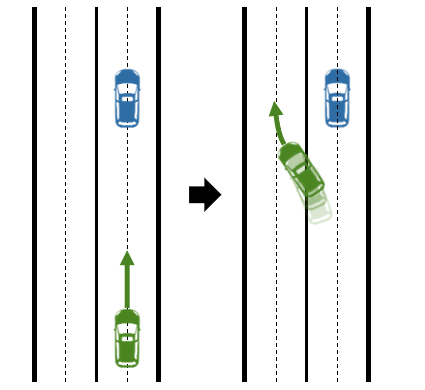}  &
     \includegraphics[width=\mywidth, height=\myheight]{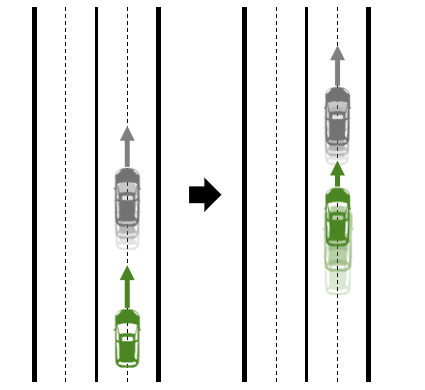}  & \includegraphics[width=\mywidth, height=\myheight]{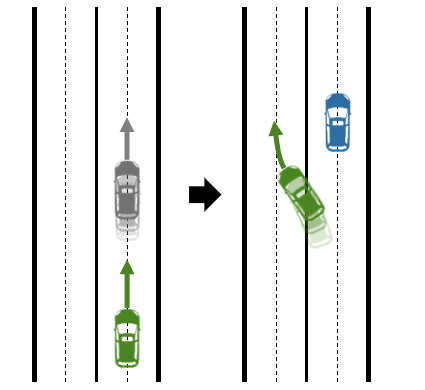} \\

     & (a) & (b) & (c) \\

     \rotatebox{90}{Hard} &
     \includegraphics[width=\mywidth, height=\myheight]{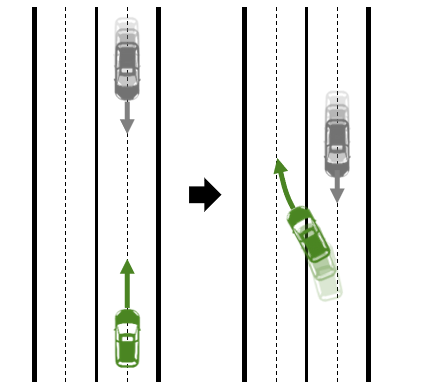}  &
     \includegraphics[width=\mywidth, height=\myheight]{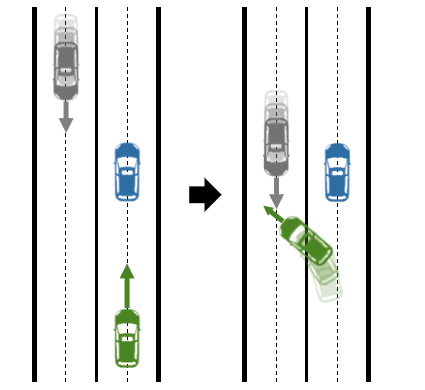}  & \includegraphics[width=\mywidth, height=\myheight]{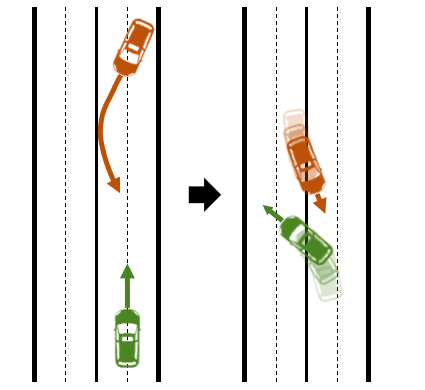} \\

     & (a) & (b) & (c) \\

     \rotatebox{90}{Extreme} &
     \includegraphics[width=\mywidth, height=\myheight]{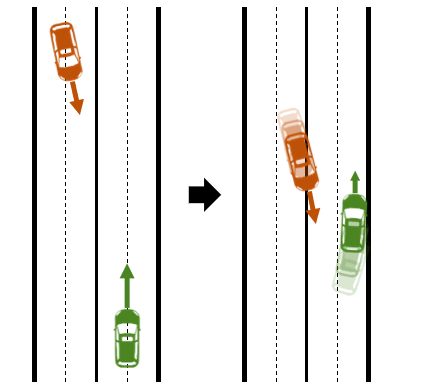}  &
     \includegraphics[width=\mywidth, height=\myheight]{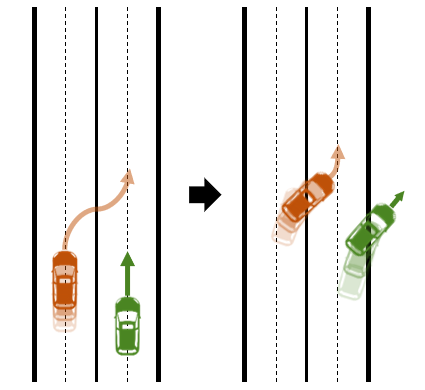}  & \includegraphics[width=\mywidth, height=\myheight]{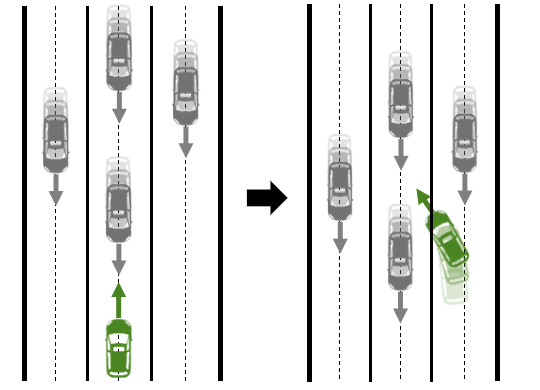} \\

     & (a) & (b) & (c) \\
     
     \end{tabular}
     \caption{\textbf{Scenarios Design}. Green, blue, gray, and orange represent the ego vehicle, static actors, normal actors, and aggressive actors, respectively.}
\label{fig:scenarios}
\vspace{-0.4cm}
\end{figure*}

%% file: supplement/tabs/speed_breakdown.tex
\begin{table*}[t]
\centering
\small
{
\setlength{\tabcolsep}{5pt}
\begin{tabular}{lcccccc}
\toprule 
& Pre. & + $\pi$ RGB & + Affine & + $\pi$ Semantic & + $\pi$ Flow  \\
\cline{2-6}
Speed (ms) & 6.25 & 8.13 (+1.88) & 8.54 (+0.41) & 9.70 (+1.16) & 10.17 (+0.47)  \\
\bottomrule
\end{tabular}
}
\caption{\textbf{Time consumption breakdown} of our method.}
\label{tab:breakdown}
\end{table*}

%% file: supplement/tabs/comparison_dynamic_gtpose.tex
\begin{table*}
\centering
\small
\setlength{\tabcolsep}{4pt}
\begin{tabular}{@{\extracolsep{2pt}}lcccccccccccc@{}} 
\toprule
\multicolumn{1}{c}{} & \multicolumn{3}{c}{KITTI Scene02} & \multicolumn{3}{c}{KITTI Scene06} & \multicolumn{3}{c}{vKITTI Scene02} & \multicolumn{3}{c}{vKITTI Scene06} \\
& PSNR$\uparrow$  & SSIM$\uparrow$  & LPIPS$\downarrow$  & PSNR$\uparrow$  & SSIM$\uparrow$  & LPIPS$\downarrow$ & PSNR$\uparrow$  & SSIM$\uparrow$  & LPIPS$\downarrow$  & PSNR$\uparrow$  & SSIM$\uparrow$  & LPIPS$\downarrow$  \\
\cline{2-4} \cline{5-7} \cline{8-10}  \cline{11-13} 
NSG \cite{ost2021neural}  & 22.51 & 0.653 & 0.397 & 23.38 & 0.717 & 0.243 & 23.50 & 0.718 & 0.352 & 26.42 & 0.811 & 0.170 \\
MARS \cite{wu2023mars} & 22.95 & 0.728 & 0.145 & 27.01 & 0.883 & 0.062 & 29.80 & 0.950 & 0.034 & 32.71 & 0.959 & 0.023 \\
Ours & \textbf{25.89} & \textbf{0.829} & \textbf{0.092} & \textbf{28.90} & \textbf{0.925} & \textbf{0.016} & \textbf{30.73} & \textbf{0.955} & \textbf{0.018} & \textbf{33.31} & \textbf{0.963} & \textbf{0.010} \\
\bottomrule
\end{tabular}
\caption{\textbf{Novel View Appearance on Dynamic Scenes} with ground truth 3D trackings.}
\label{tab:comp_gt}
\end{table*}

%% file: supplement/figures/softmax_ablation/softmax_ablation.tex
\begin{figure}[t!]
     \centering
     \small 
     \setlength{\tabcolsep}{0pt}
     \def\mywidth{4.5cm}
     \begin{tabular}{P{\mywidth}P{\mywidth}}
 
     \includegraphics[width=\mywidth]{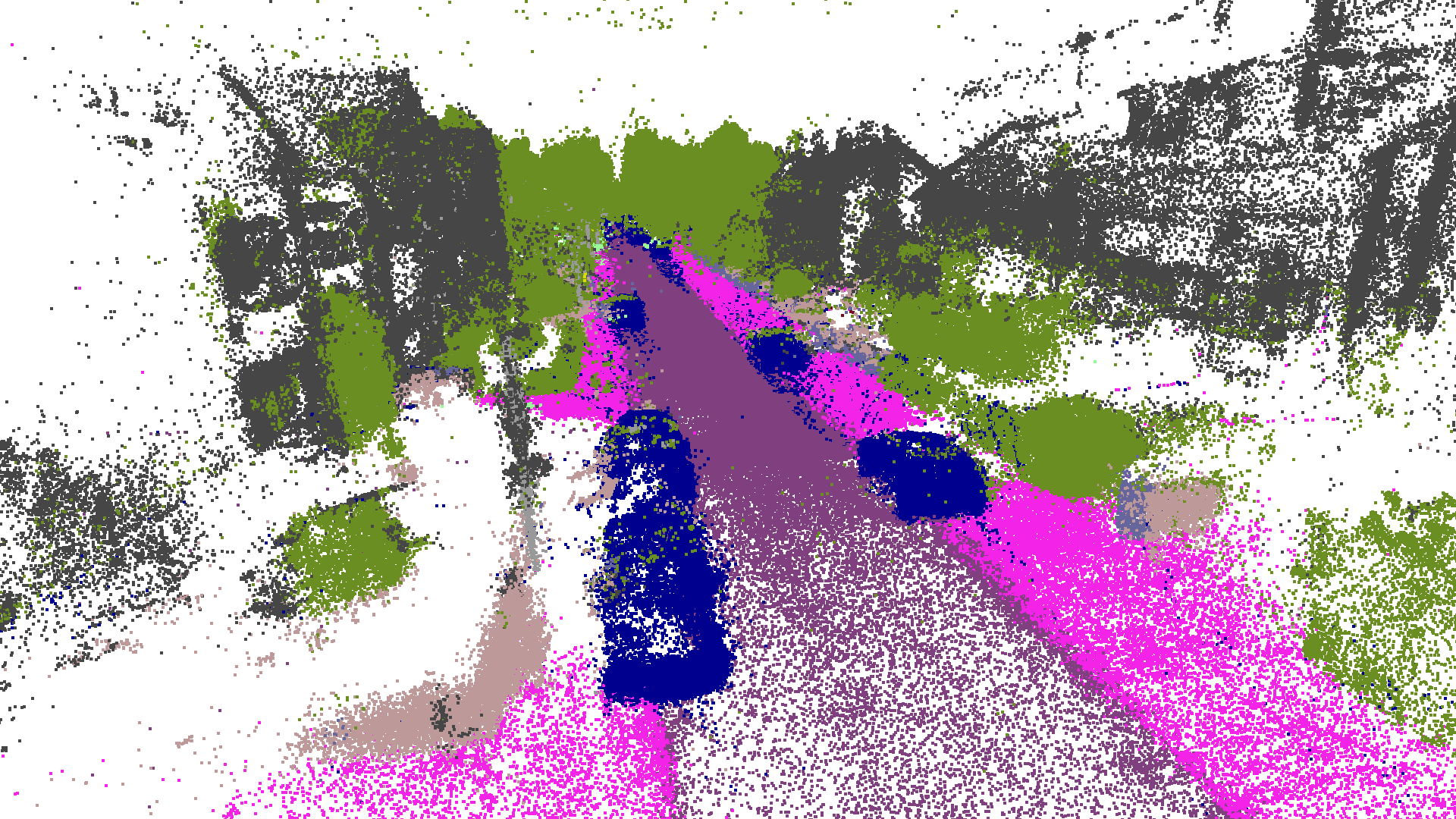}  & \includegraphics[width=\mywidth]{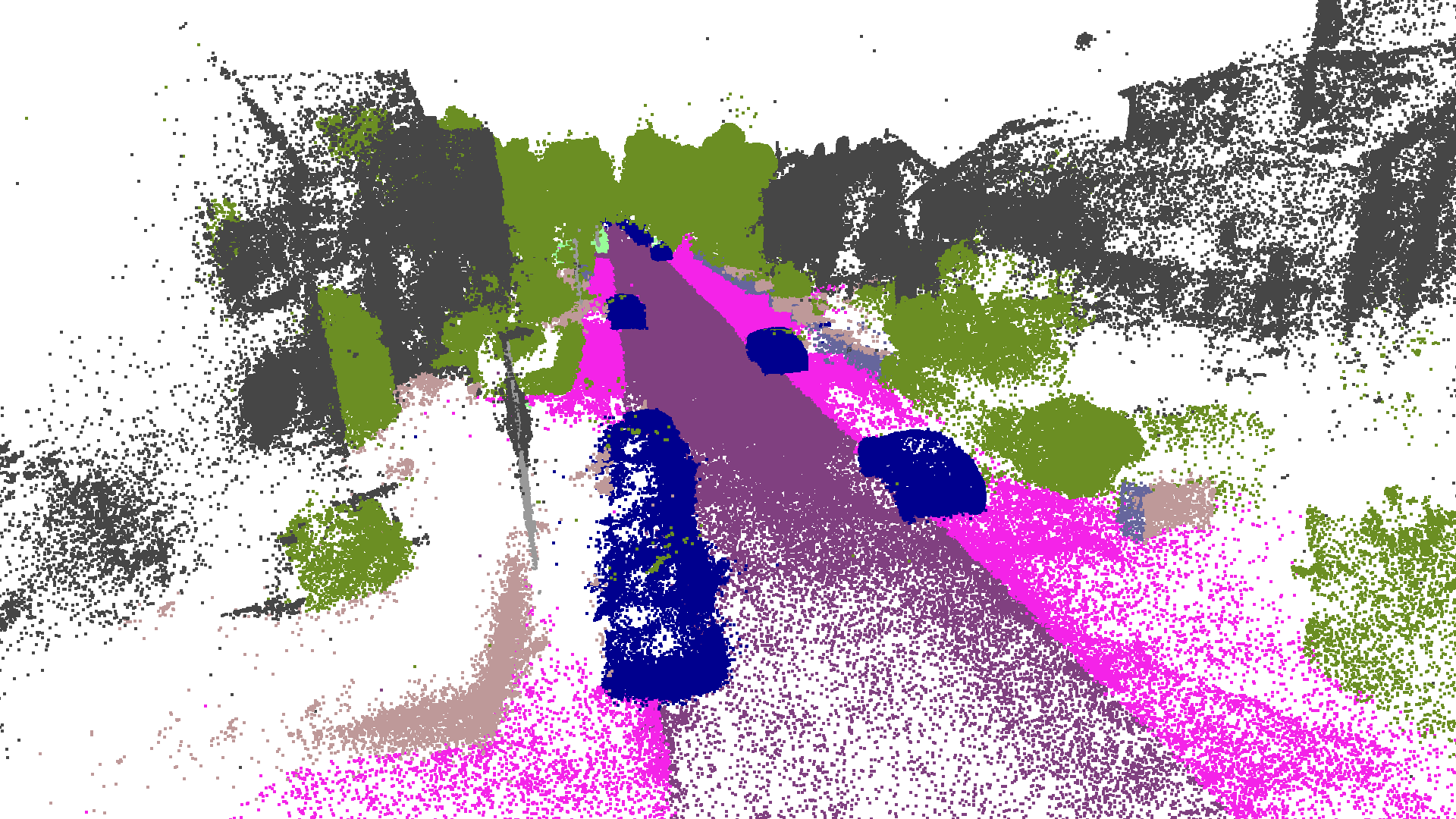} \\

     \includegraphics[width=\mywidth]{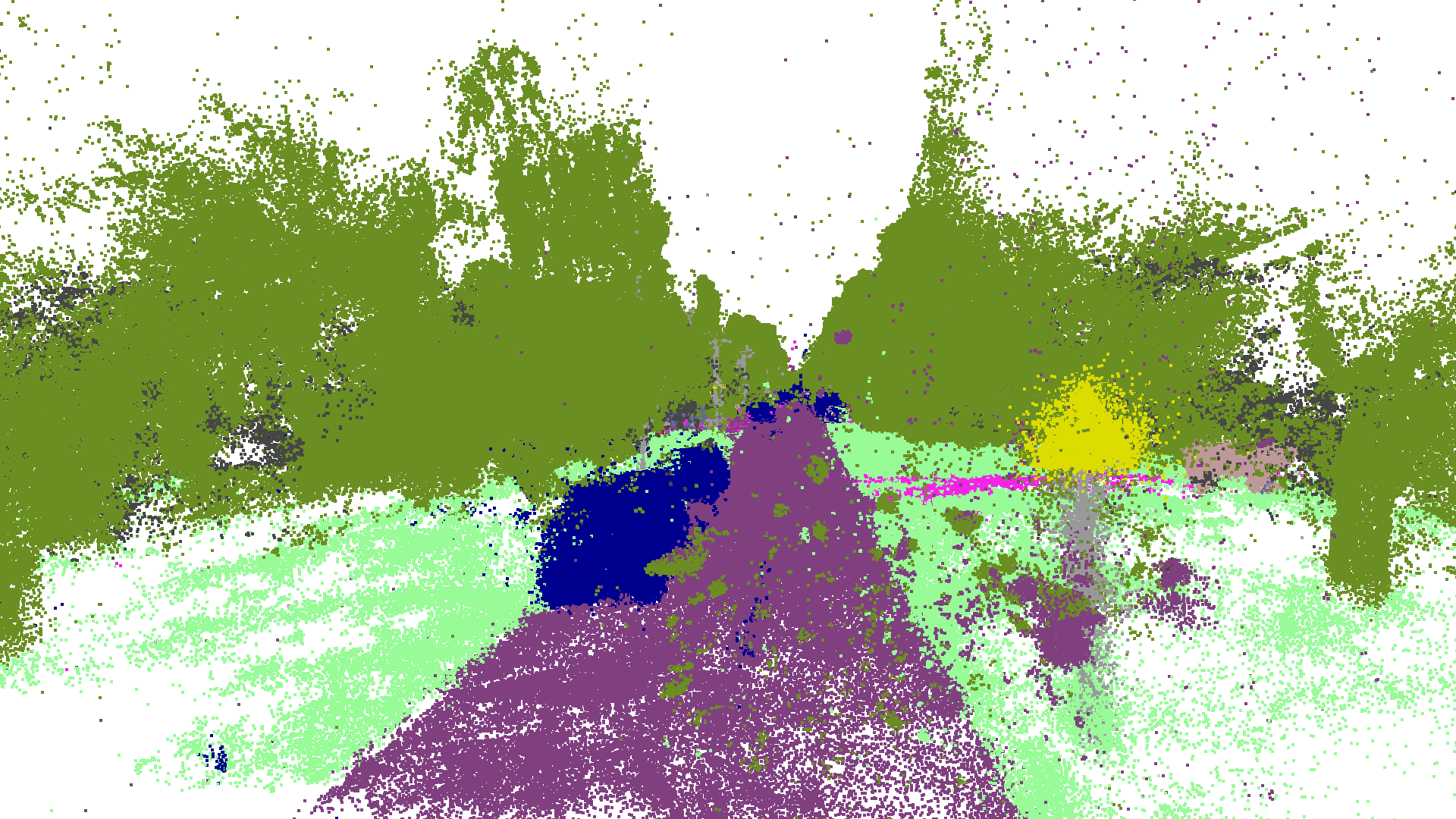}  &
     \includegraphics[width=\mywidth]{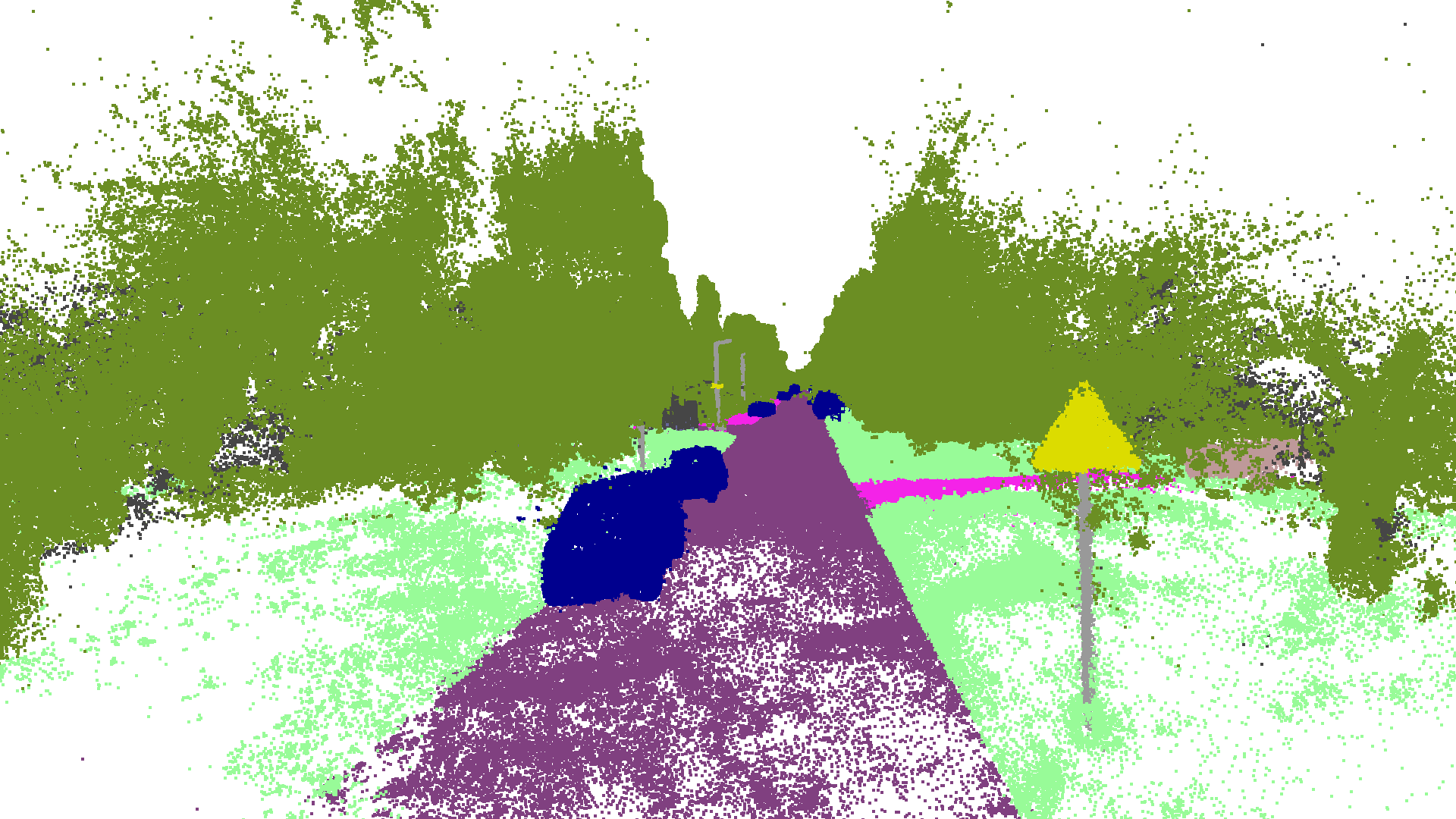} \\

     \includegraphics[width=\mywidth]{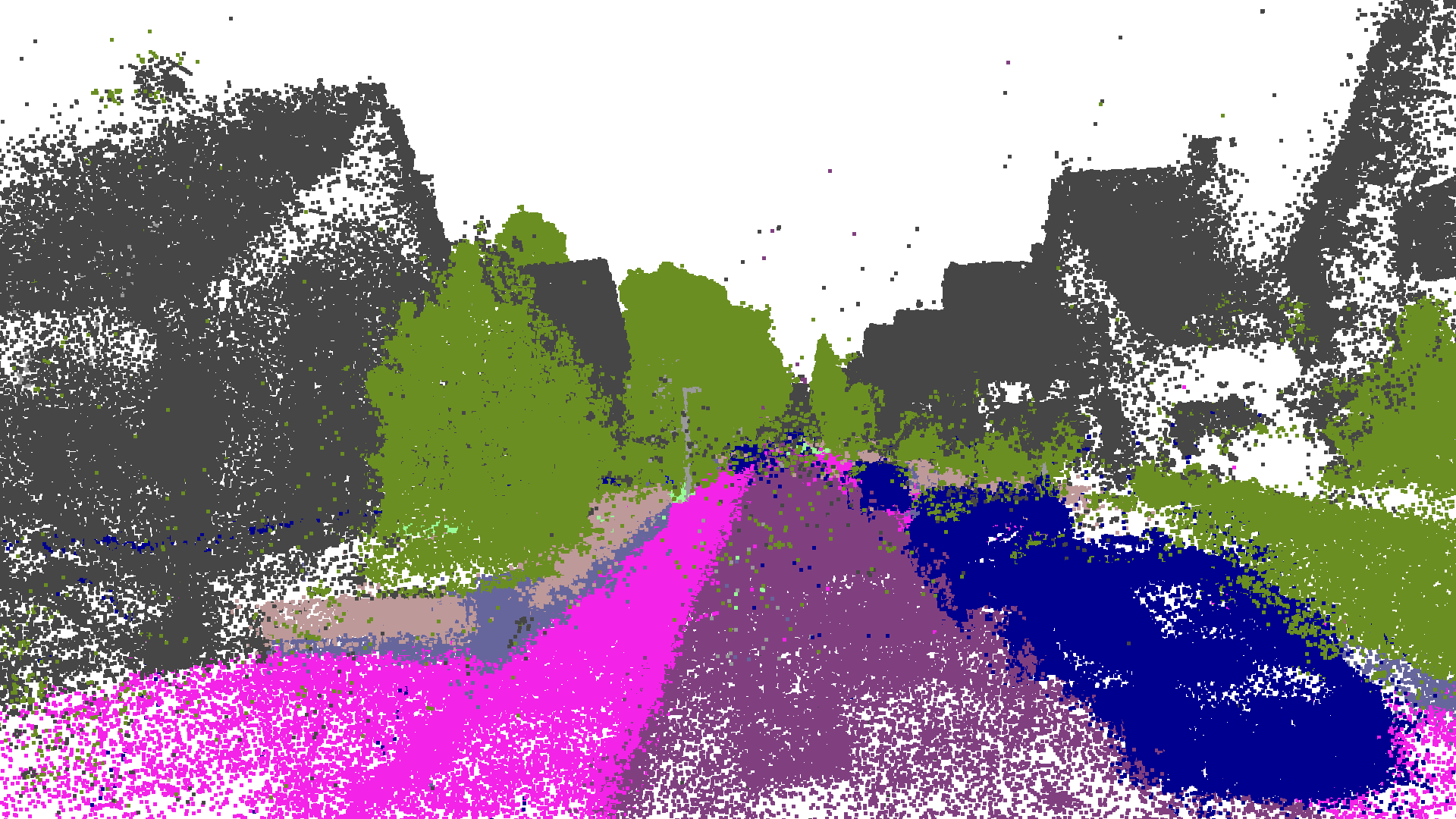}  &
     \includegraphics[width=\mywidth]{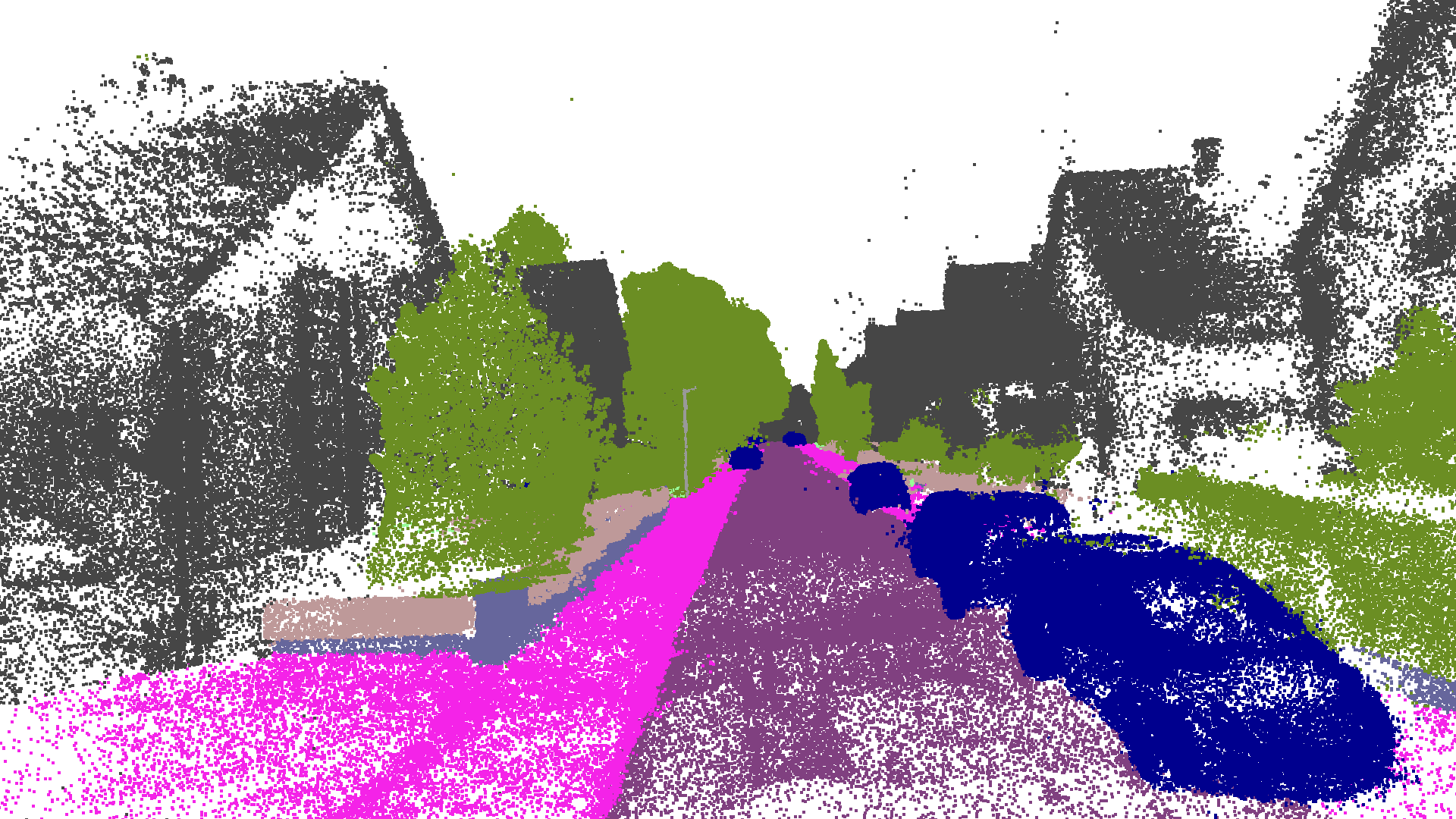} \\
    \rotatebox{0}{Ours w/ $\bS_\text{2D\_norm}$} & \rotatebox{0}{Ours w/ $\bS_\text{3D\_norm}$}
     \end{tabular}
     \caption{\textbf{Qualitative Comparison} of 3D and 2D softmax results. Note that normalizing semantic logits in 3D space (Ours w/ $\bS_\text{3D\_norm}$) clearly reduces floaters and yields better 3D semantic reconstruction than the 2D normalization counterpart (Ours w/ $\bS_\text{2D\_norm}$).}
     \vspace{-0.2cm}
\label{fig:softmax_ablation}
\vspace{-0.5cm}
\end{figure}

%% file: supplement/figures/nerfacto/nerfacto.tex
\begin{figure*}[t!]
     \centering
     \small 
     \setlength{\tabcolsep}{0pt}
     \def\mywidth{5.8cm}
     \begin{tabular}{P{\mywidth}P{\mywidth}P{\mywidth}}
 
     \includegraphics[width=\mywidth]{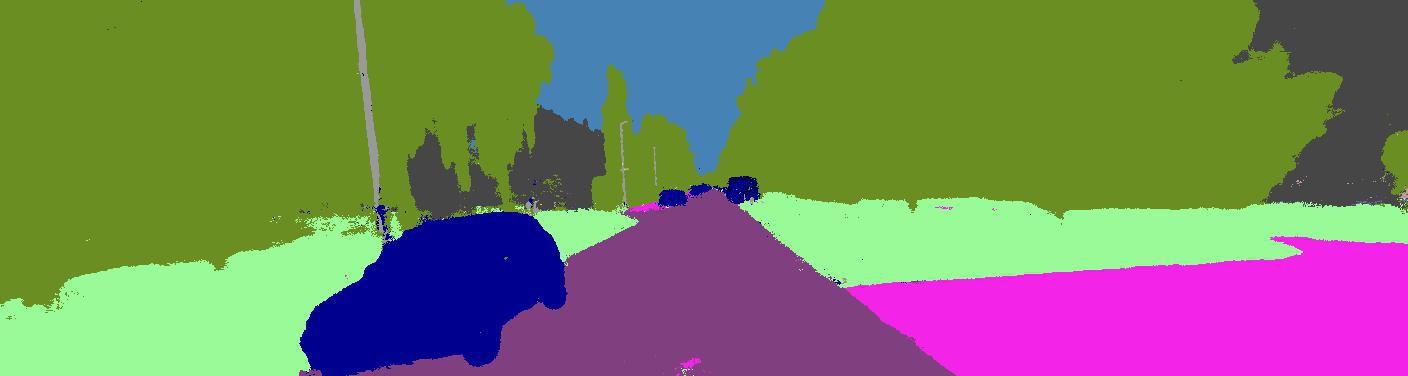}  &
     \includegraphics[width=\mywidth]{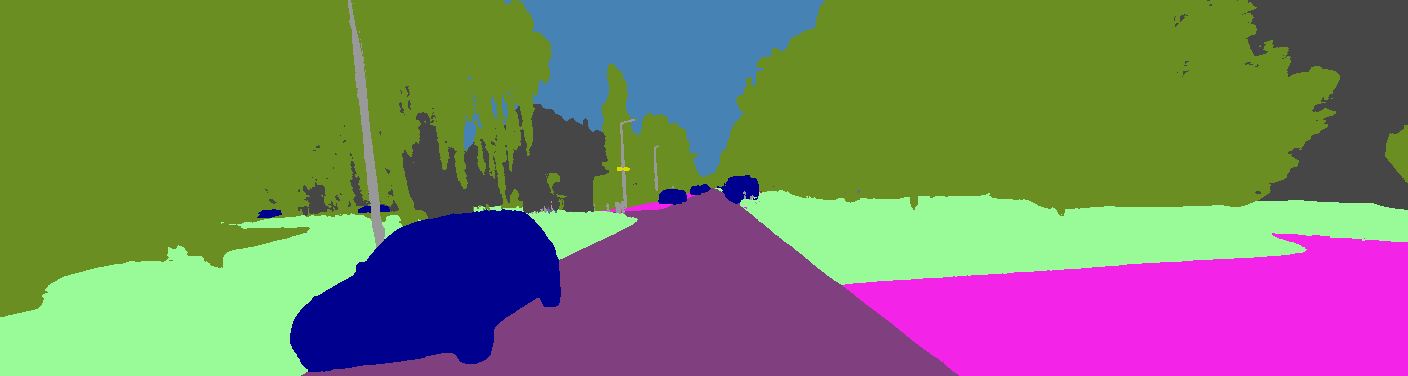}  & \includegraphics[width=\mywidth]{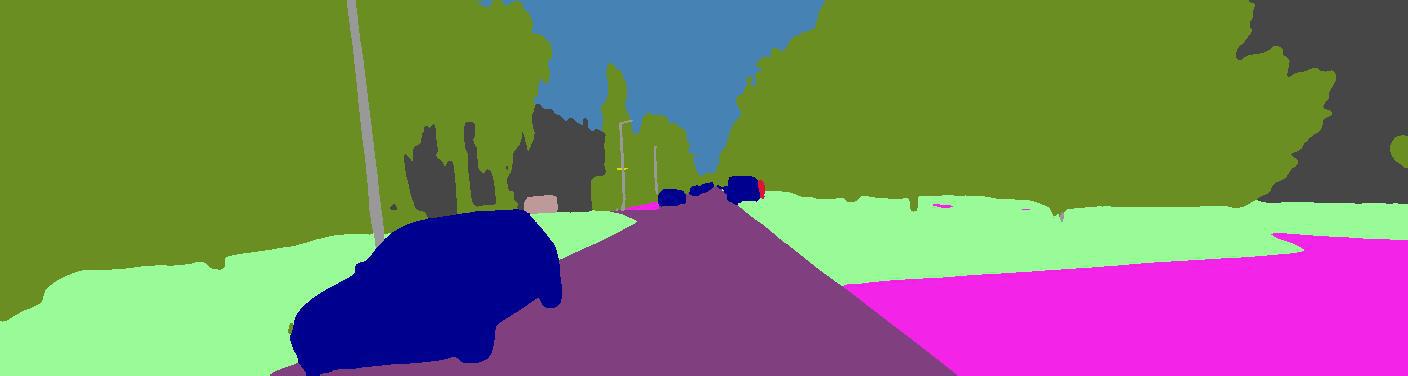} \\

     \includegraphics[width=\mywidth]{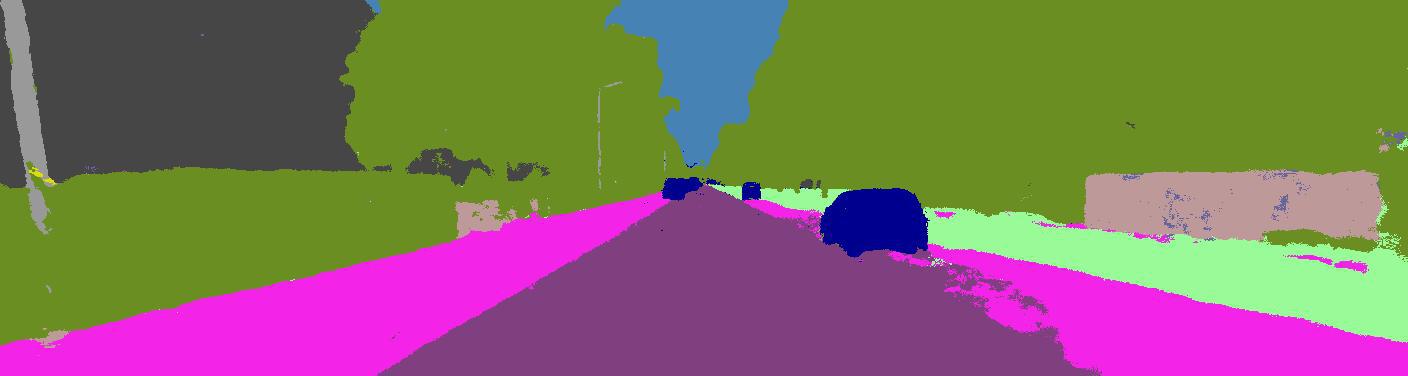}  &
     \includegraphics[width=\mywidth]{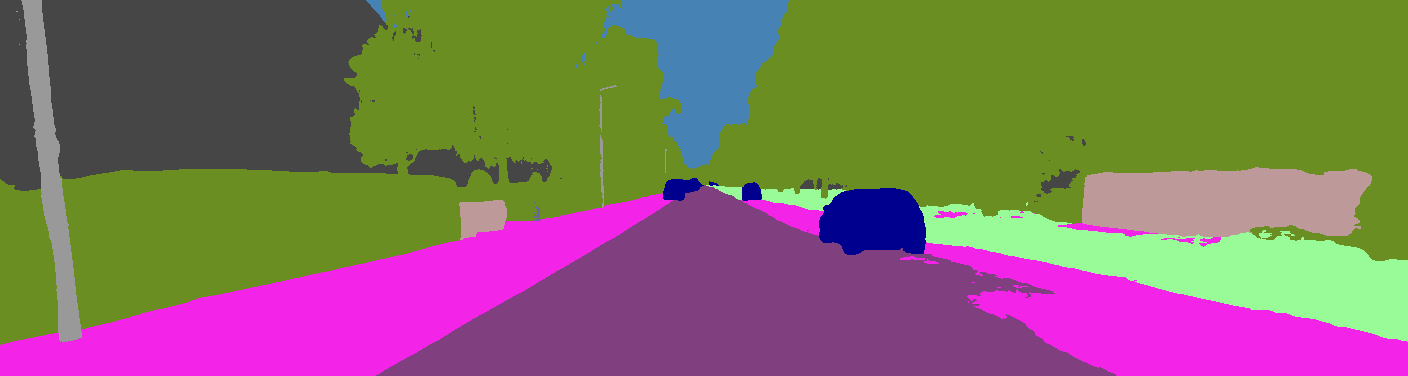}  & \includegraphics[width=\mywidth]{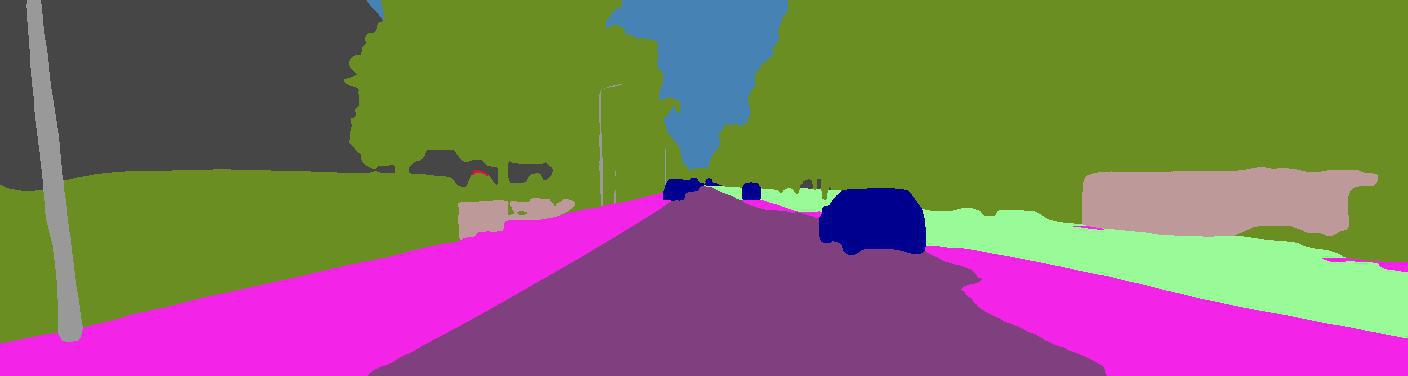} \\
     
     \includegraphics[width=\mywidth]{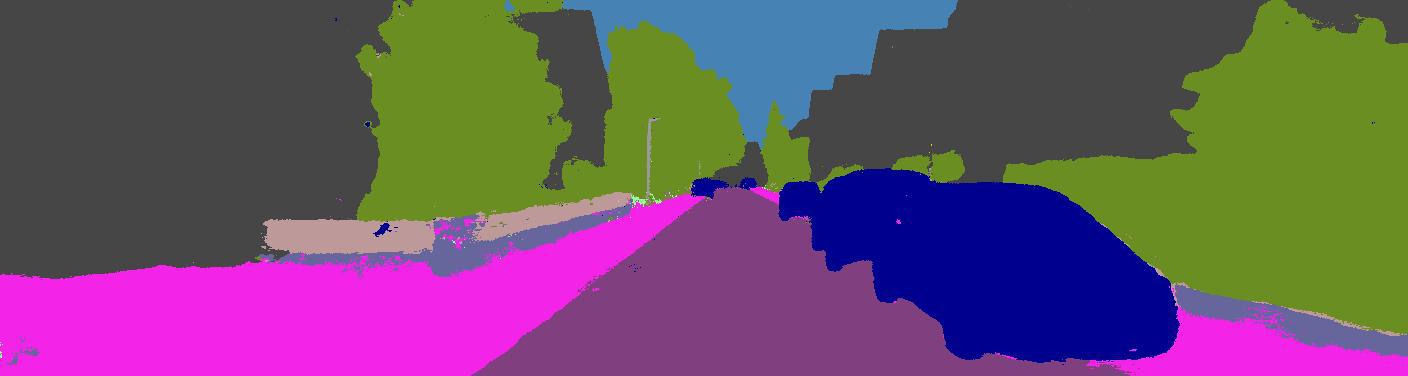}  &
     \includegraphics[width=\mywidth]{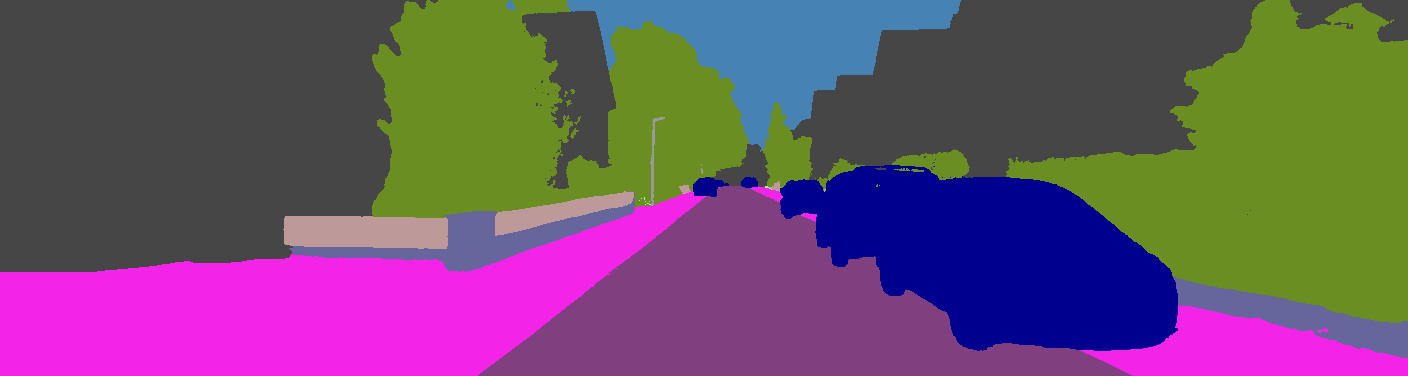}  & \includegraphics[width=\mywidth]{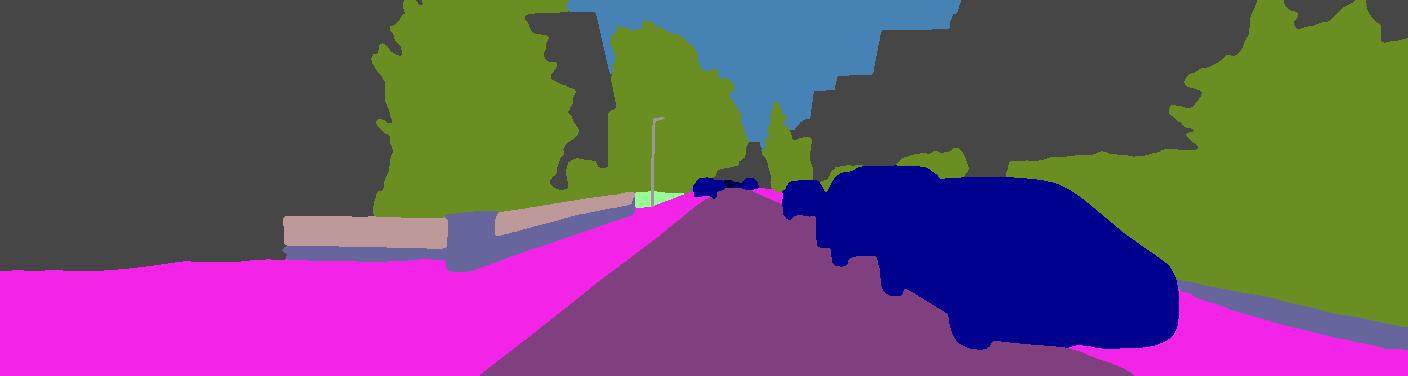} \\

     \includegraphics[width=\mywidth]{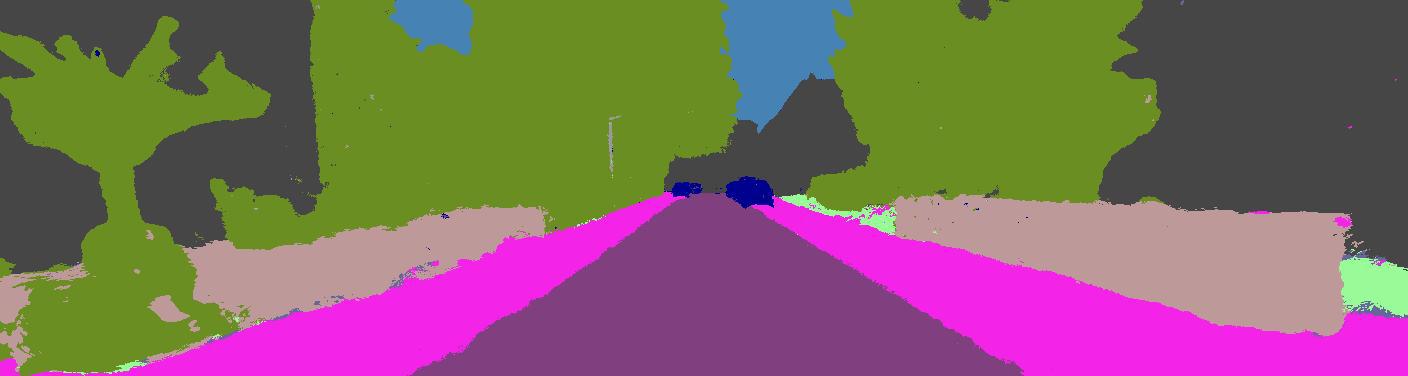}  &
     \includegraphics[width=\mywidth]{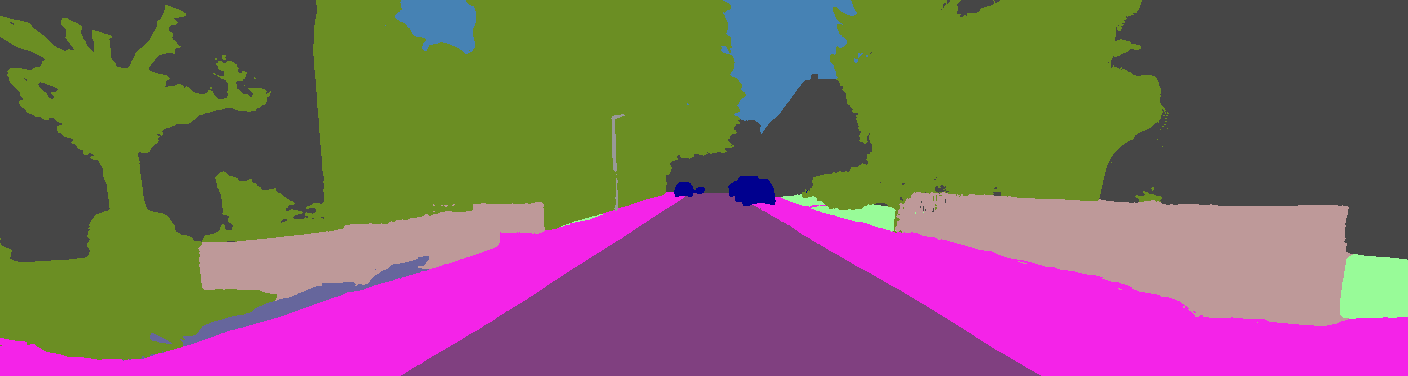}  & \includegraphics[width=\mywidth]{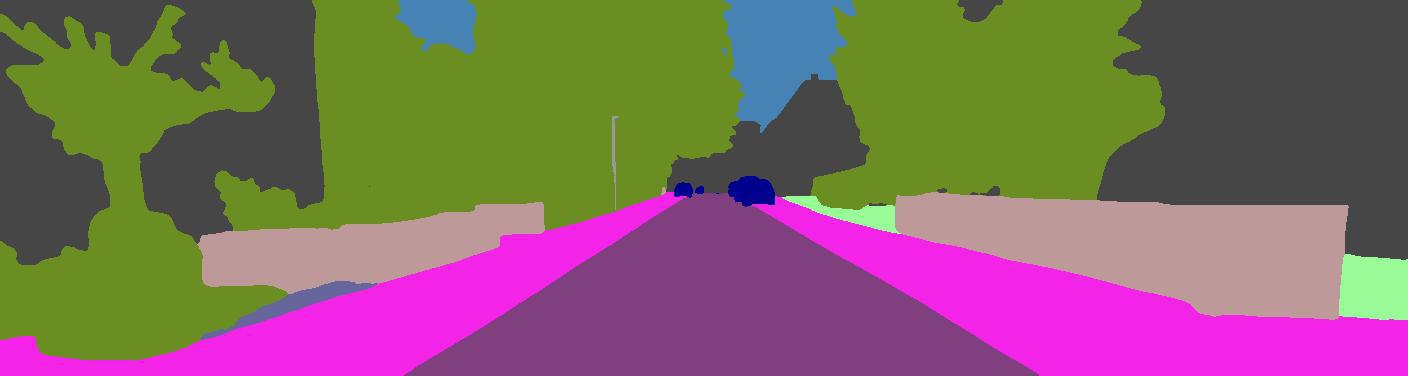} \\
    \rotatebox{0}{Semantic Nerfacto} & \rotatebox{0}{Ours} & \rotatebox{0}{Pseudo GT}
     \end{tabular}
     \caption{\textbf{Qualitative Comparison} with Nerfacto on 2D space. The Pseudo GT column represents the semantic maps that are predicted by \cite{borse2021inverseform} on GT RGB images.}
\label{fig:compare_nerfacto_2d}
\end{figure*}

\begin{figure}[t!]
     \centering
     \small 
     \setlength{\tabcolsep}{0pt}
     \def\mywidth{4.5cm}
     \begin{tabular}{P{\mywidth}P{\mywidth}}
 
     \includegraphics[width=\mywidth]{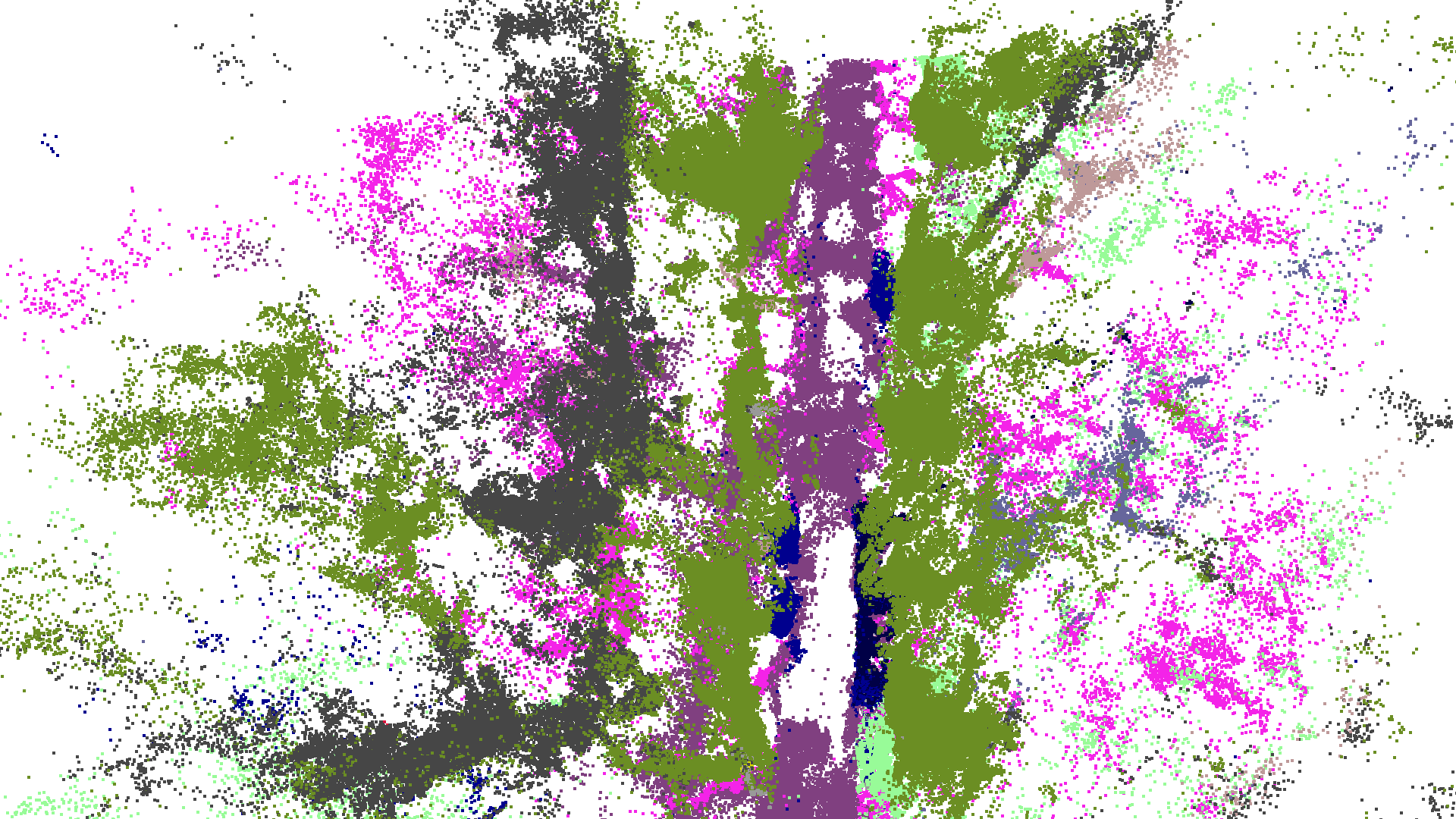}  & \includegraphics[width=\mywidth]{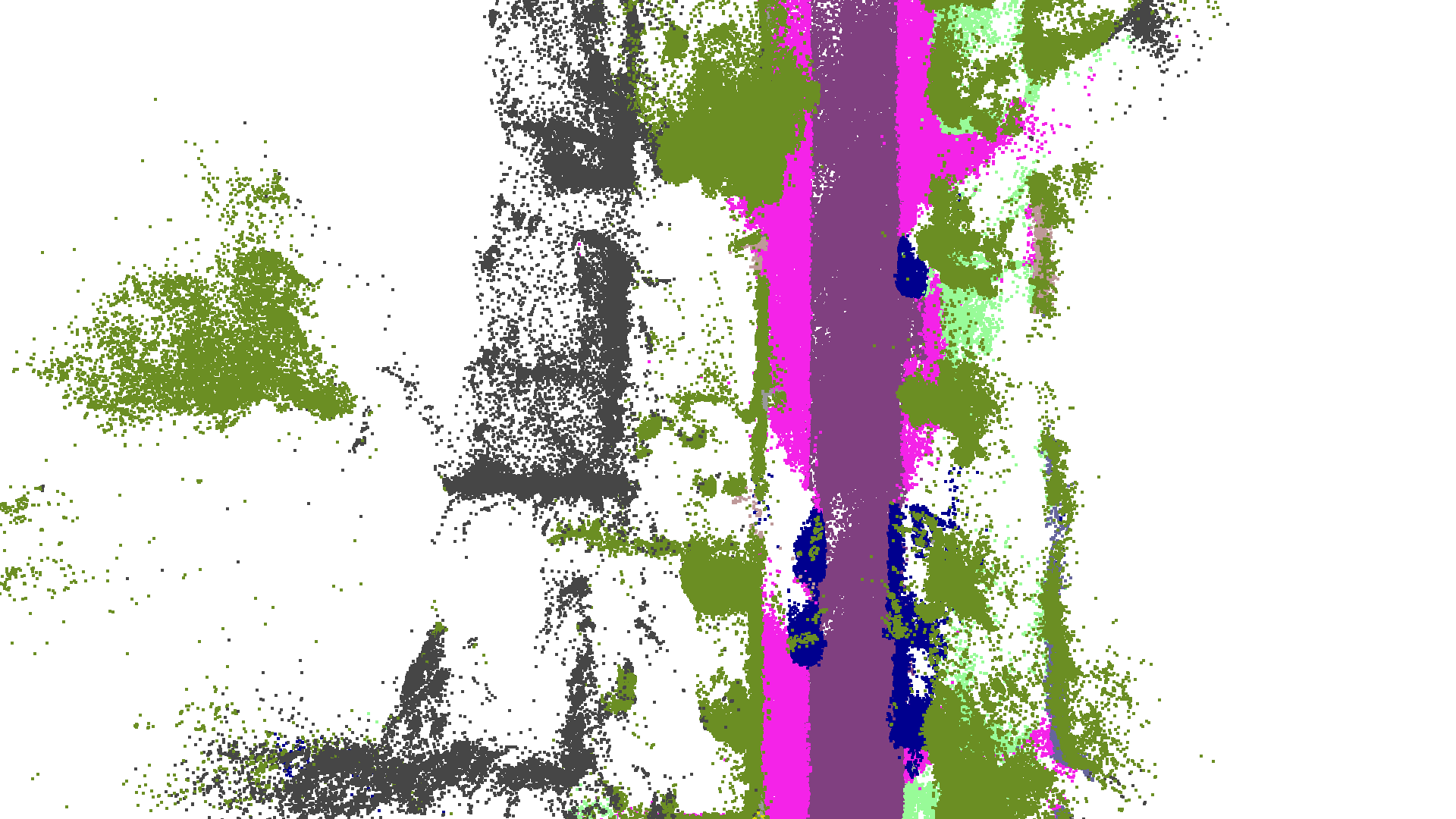} \\

     \includegraphics[width=\mywidth]{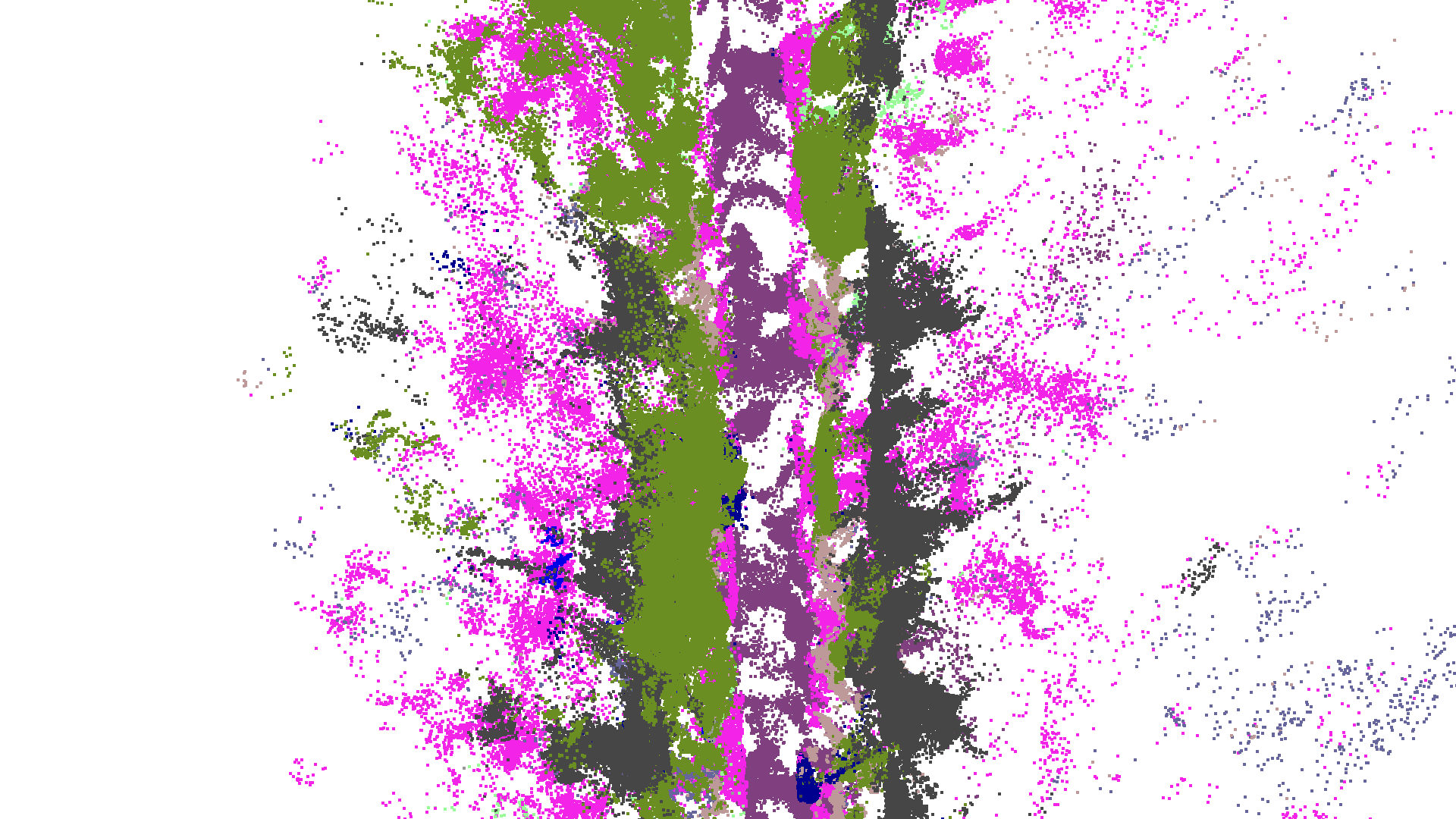}  &
     \includegraphics[width=\mywidth]{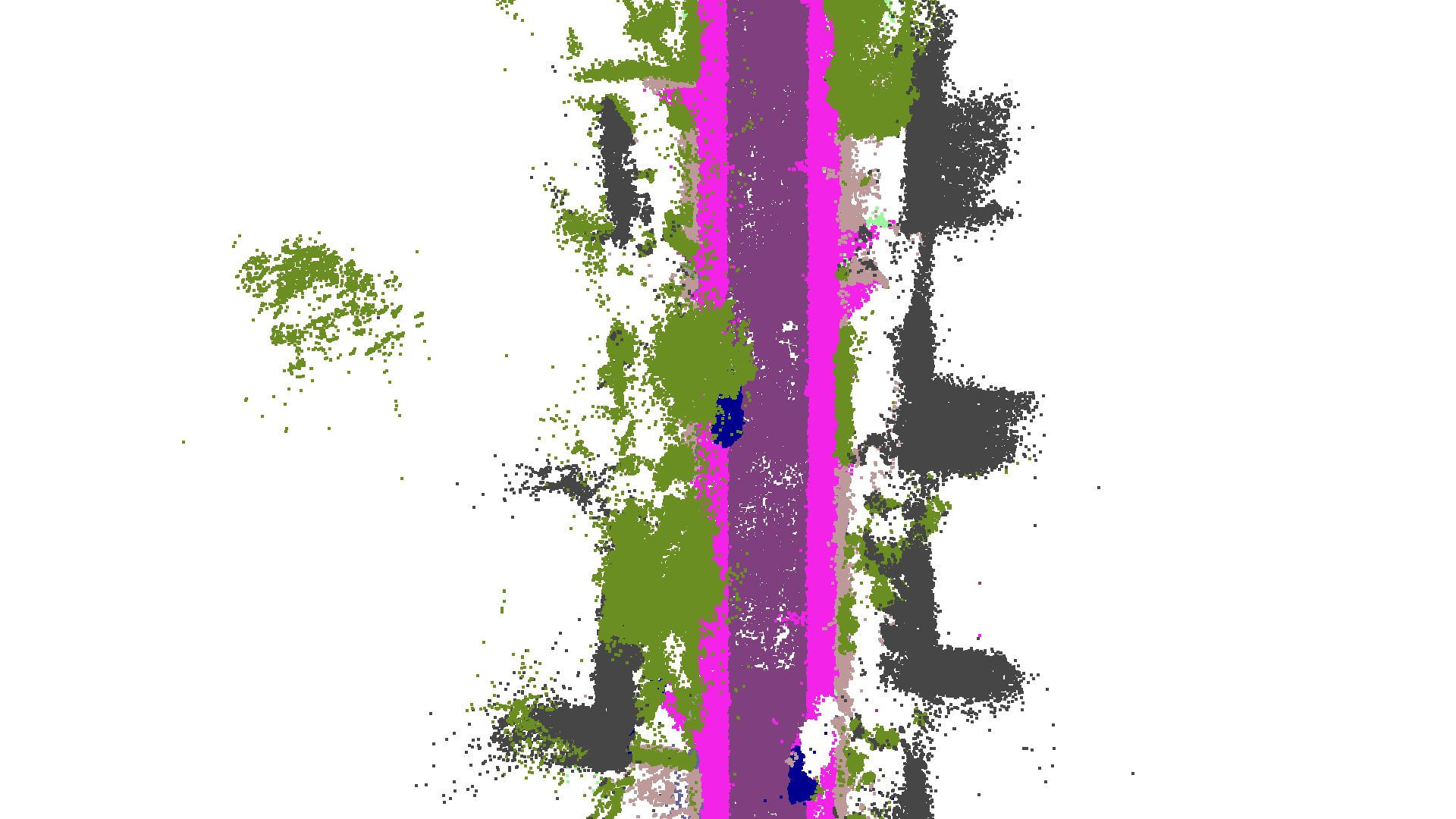} \\
    \rotatebox{0}{Semantic Nerfacto} & \rotatebox{0}{Ours}
     \end{tabular}
     \caption{\textbf{Qualitative Comparison} with Nerfacto on 3D space.
     The semantic point cloud extracted from Semantic Nerfacto struggles to faithfully represent the geometry.
     }
\label{fig:compare_nerfacto_3d}
\vspace{-0.2cm}
\end{figure}

%% file: supplement/figures/optimize/optimize.tex
\begin{figure*}[t!]
     \centering
     \small 
     \setlength{\tabcolsep}{0pt}
     \def\mywidth{6cm}
     \begin{tabular}{P{\mywidth}P{\mywidth}P{\mywidth}}
 
     \includegraphics[width=\mywidth]{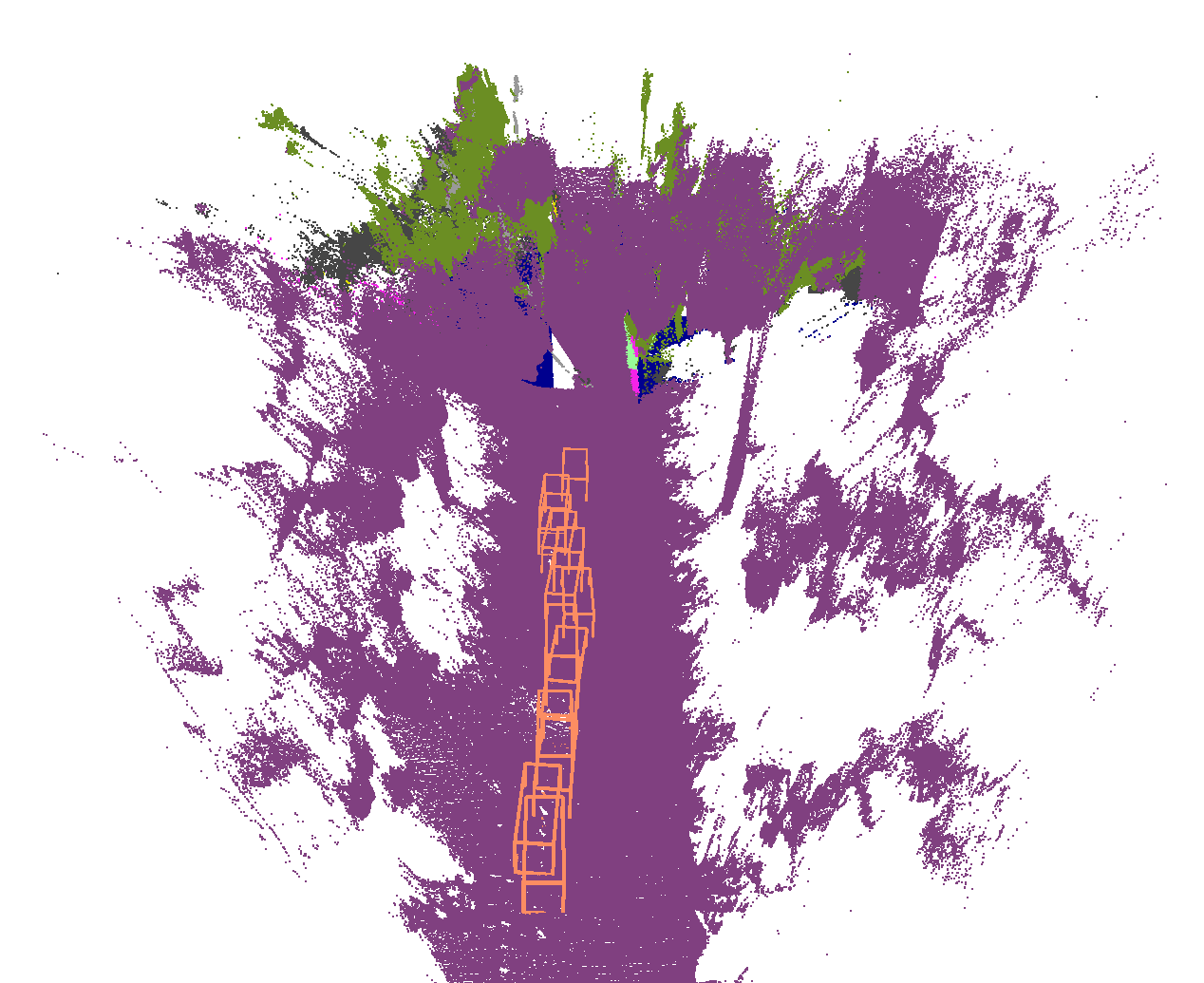}  & 
     \includegraphics[width=\mywidth]{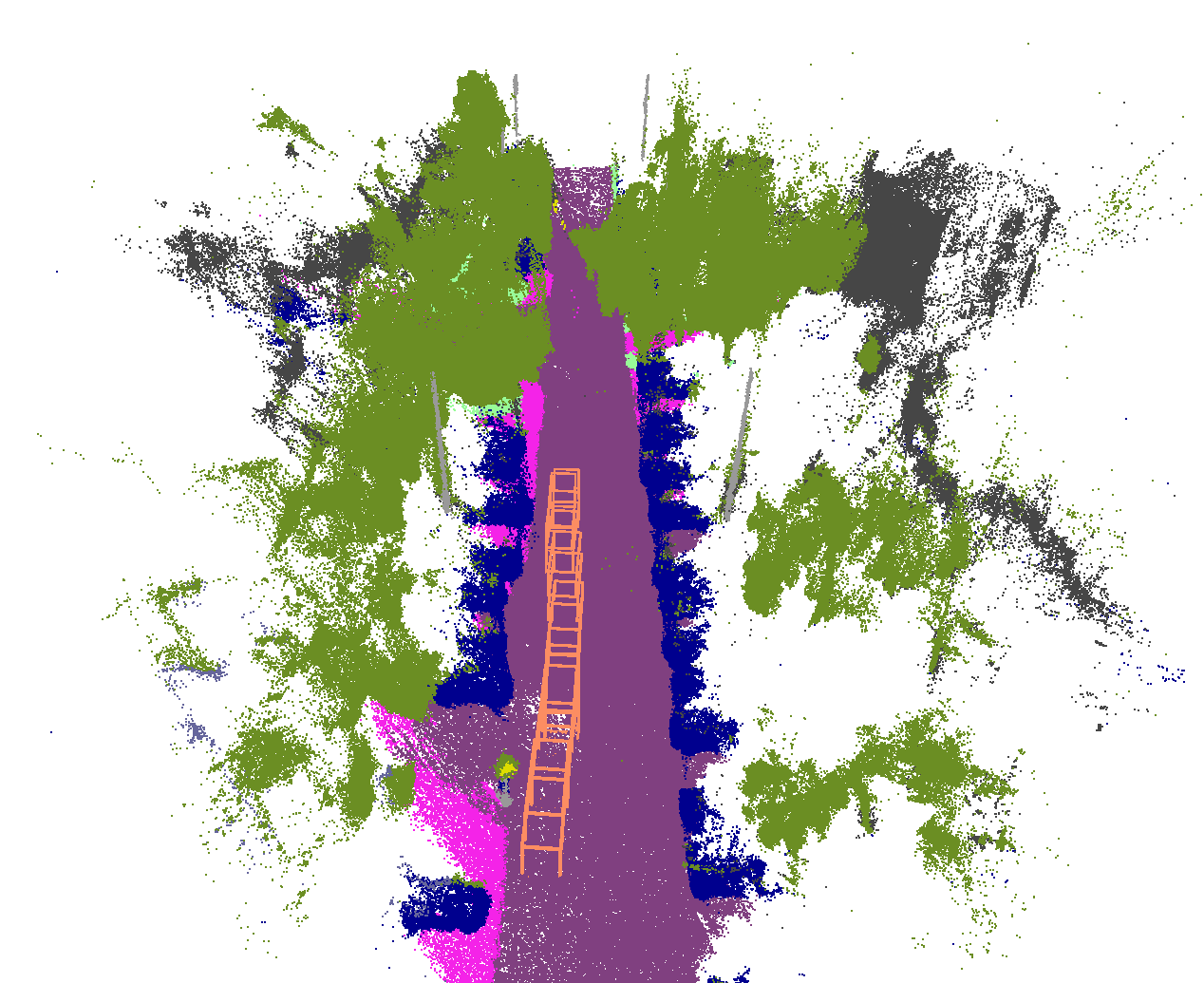}  &
     \includegraphics[width=\mywidth]{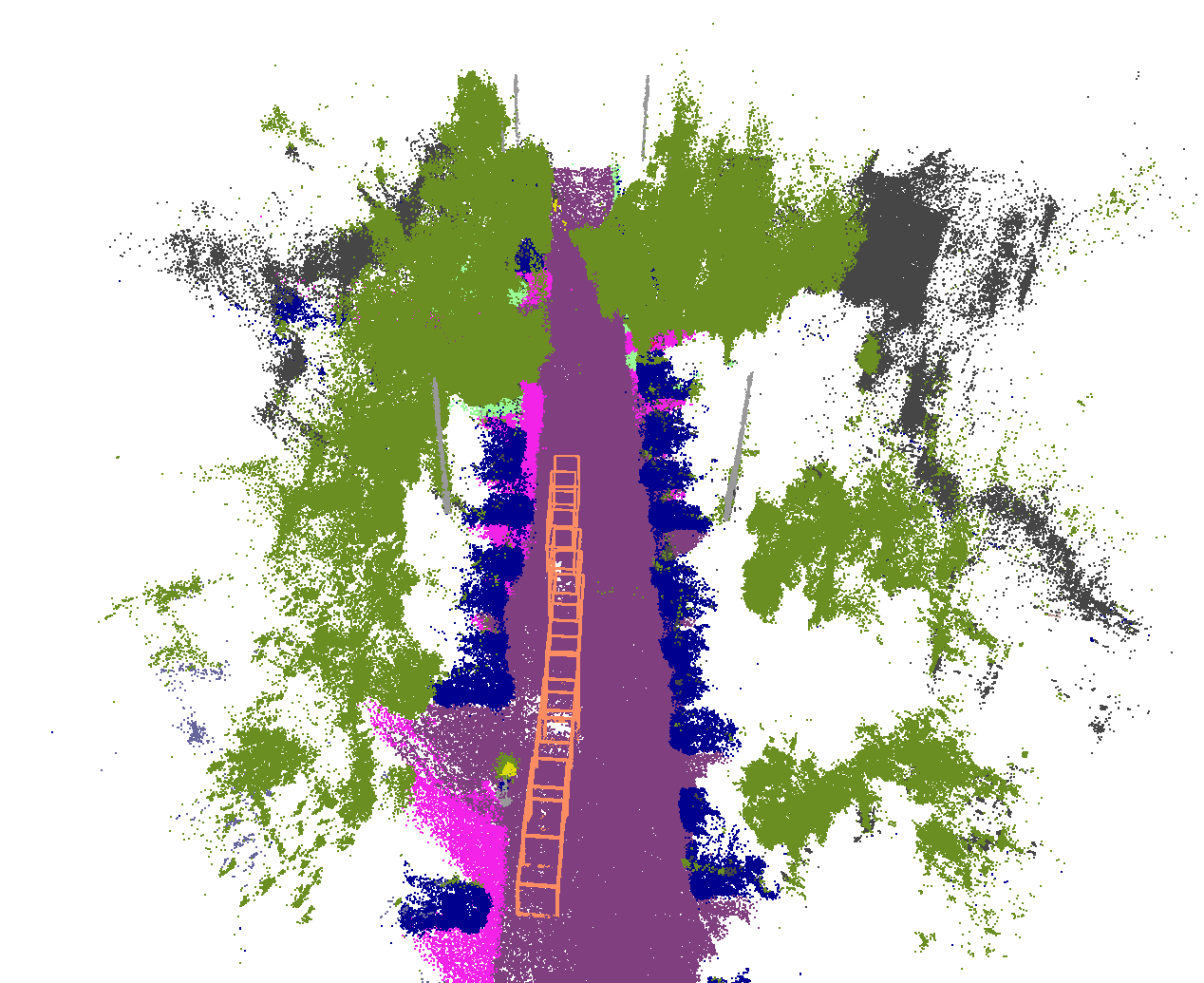} \\

    \rotatebox{0}{10 steps} & \rotatebox{0}{2000 steps} & \rotatebox{0}{5000 steps}
     \end{tabular}
     \caption{\textbf{Visualization of Optimization Progress}. Our method jointly optimizes the static background and the trajectory of the dynamic foreground objects. By integrating physical constraints using the unicycle model, our method allows for recovering a smooth trajectory from noisy 3D bounding boxes. To prevent visual clutter, we exclude point clouds of the dynamic object and only visualize the bounding boxes.}
     \vspace{-0.2cm}
\label{fig:vis_optim}
\end{figure*}

%% file: supplement/tabs/compare_qd3dt.tex
\begin{table}[t]
\centering
\small
{
\begin{tabular}{llll}\hline
  \cline{1-4}
                    &       & $e_\mathbf{R}\downarrow$   & $e_\mathbf{t}\downarrow$  \\ 
\cline{2-4}
\multirow{2}{*}{02} & QD-3DT & 0.027             & 0.215                \\
                    & Ours  & \textbf{0.018}    & \textbf{0.108}       \\ 
\cline{2-4}
\multirow{2}{*}{06} & QD-3DT & 0.017             & 0.046                \\
                    & Ours  & \textbf{0.012}    & \textbf{0.033}       \\
\cline{1-4}
\end{tabular}
}
\caption{\textbf{Quantitative Pose Comparison} on two KITTI sequences.}
\label{tab:traj}
\vspace{-0.5cm}
\end{table}

%% file: tables/adtest/ALL.tex
\begin{table*}
\centering
\small
\setlength{\tabcolsep}{2.1pt}
\begin{tabular}{@{\extracolsep{2pt}}llcccc|cccc|cccc@{}} 
\toprule
{} & {} & \multicolumn{4}{c}{UniAD \cite{hu2023planning}} & \multicolumn{4}{c}{VAD \cite{jiang2023vad}}  & \multicolumn{4}{c}{LTF \cite{Chitta2023PAMI}} \\
\toprule
{} & {} & Easy & Medium & Hard & Extreme & Easy & Medium & Hard & Extreme & Easy & Medium & Hard & Extreme \\

\cline{2-14}

\multirow{5}{*}{KITTI-360 \cite{liao2022kitti}} 
& $NC \uparrow$ & \textbf{0.459}& \textbf{0.379}& \textbf{0.315}& \textbf{0.242}& 0.333 & 0.322 & 0.253 & 0.171 & 0.330 & 0.170 & 0.181 & 0.133 \\
& $DAC \uparrow$ & \textbf{0.712}& \textbf{0.711}& \textbf{0.754}& \textbf{0.750}& 0.603 & 0.604 & 0.609 & 0.657 & 0.646 & 0.627 & 0.622 & 0.633 \\
& $TTC \uparrow$ & 0.271 & \textbf{0.193}& 0.131 & \textbf{0.091}& 0.204 & 0.162 & 0.105 & 0.090 & \textbf{0.290}& 0.125 & \textbf{0.132}& 0.080 \\
& $COM \uparrow$ & 0.672 & 0.652 & 0.555 & 0.527 & \textbf{1.000}& \textbf{1.000}& \textbf{1.000}& \textbf{1.000}& \textbf{1.000}& 0.996 & \textbf{1.000}& \textbf{1.000} \\
& $R_c \uparrow$ & 0.166 & 0.117 & 0.122 & 0.094 & 0.138 & 0.125 & \textbf{0.122}& 0.110 & \textbf{0.245}& \textbf{0.138}& 0.121 & \textbf{0.111} \\
& HD-Score$\uparrow$ & 0.047 & 0.019 & \textbf{0.017}& 0.006 & 0.029 & 0.014 & 0.014 & \textbf{0.008}& \textbf{0.080}& \textbf{0.028}& 0.017 & 0.003 \\

\cline{2-14}

\multirow{5}{*}{Waymo \cite{Sun2020CVPR}} 
& $NC \uparrow$ & \textbf{0.900}& \textbf{0.903}& \textbf{0.792}& \textbf{0.512}& 0.698 & 0.493 & 0.526 & 0.416 & 0.855 & 0.544 & 0.451 & 0.333 \\
& $DAC \uparrow$ & \textbf{0.863}& \textbf{0.903}& 0.771 & \textbf{0.915}& 0.568 & 0.679 & 0.694 & 0.726 & 0.779 & 0.830 & \textbf{0.861}& 0.852 \\
& $TTC \uparrow$ & \textbf{0.862}& \textbf{0.778}& \textbf{0.711}& \textbf{0.461}& 0.641 & 0.363 & 0.437 & 0.299 & 0.833 & 0.518 & 0.428 & 0.258 \\
& $COM \uparrow$ & 0.890 & 0.767 & 0.712 & 0.483 & \textbf{1.000}& \textbf{1.000}& \textbf{1.000}& \textbf{1.000}& 0.992 & 0.995 & 0.995 & \textbf{1.000} \\
& $R_c \uparrow$ & \textbf{0.784}& \textbf{0.547}& \textbf{0.590}& \textbf{0.326}& 0.323 & 0.266 & 0.259 & 0.272 & 0.698 & 0.371 & 0.357 & 0.261 \\
& HD-Score$\uparrow$ & \textbf{0.664}& \textbf{0.419}& \textbf{0.401}& \textbf{0.171}& 0.154 & 0.093 & 0.110 & 0.085 & 0.546 & 0.200 & 0.219 & 0.071 \\

\cline{2-14}

\multirow{5}{*}{Nuscenes \cite{caesar2020nuscenes}} 
& $NC \uparrow$ & 0.823 & \textbf{0.761}& \textbf{0.781}& \textbf{0.688}& 0.765 & 0.486 & 0.516 & 0.467 & \textbf{0.835}& 0.587 & 0.513 & 0.401 \\
& $DAC \uparrow$ & \textbf{0.967}& \textbf{0.925}& 0.929 & \textbf{0.973}& 0.824 & 0.917 & \textbf{0.941}& 0.971 & 0.858 & 0.872 & 0.872 & 0.928 \\
& $TTC \uparrow$ & \textbf{0.787}& \textbf{0.652}& \textbf{0.673}& \textbf{0.519}& 0.692 & 0.365 & 0.376 & 0.396 & 0.781 & 0.527 & 0.453 & 0.353 \\
& $COM \uparrow$ & 0.880 & 0.728 & 0.729 & 0.642 & \textbf{1.000}& \textbf{1.000}& \textbf{1.000}& \textbf{1.000}& 0.994 & 0.995 & 1.000 & 0.993 \\
& $R_c \uparrow$ & 0.703 & 0.510 & \textbf{0.516}& 0.291 & 0.540 & 0.340 & 0.383 & 0.311 & \textbf{0.847}& \textbf{0.542}& 0.498 & \textbf{0.357} \\
& HD-Score$\uparrow$ & 0.589 & \textbf{0.378}& \textbf{0.365}& 0.168 & 0.348 & 0.132 & 0.204 & 0.158 & \textbf{0.616}& 0.307 & 0.226 & \textbf{0.171} \\

\cline{2-14}

\multirow{5}{*}{Pandaset \cite{xiao2021pandaset}} 
& $NC \uparrow$ & 0.914 & \textbf{0.851}& \textbf{0.771}& \textbf{0.732}& 0.847 & 0.529 & 0.474 & 0.388 & \textbf{0.945}& 0.597 & 0.481 & 0.265 \\
& $DAC \uparrow$ & \textbf{0.998}& 0.933 & 0.987 & 0.947 & 0.963 & 0.990 & 0.995 & \textbf{0.993}& 0.987 & \textbf{1.000}& \textbf{0.999}& 0.991 \\
& $TTC \uparrow$ & 0.913 & \textbf{0.792}& \textbf{0.685}& \textbf{0.644}& 0.793 & 0.270 & 0.235 & 0.251 & \textbf{0.919}& 0.560 & 0.467 & 0.182 \\
& $COM \uparrow$ & 0.871 & 0.734 & 0.692 & 0.689 & \textbf{1.000}& \textbf{1.000}& \textbf{1.000}& \textbf{1.000}& \textbf{1.000}& \textbf{1.000}& \textbf{1.000}& \textbf{1.000}\\
& $R_c \uparrow$ & 0.691 & 0.473 & 0.389 & \textbf{0.329}& 0.548 & 0.349 & 0.257 & 0.229 & \textbf{0.947}& \textbf{0.579}& \textbf{0.498}& 0.293 \\
& HD-Score$\uparrow$ & 0.647 & 0.366 & 0.308 & \textbf{0.226}& 0.442 & 0.158 & 0.087 & 0.079 & \textbf{0.871}& \textbf{0.450}& \textbf{0.331}& 0.080 \\

\cline{2-14}

\multirow{5}{*}{Average} 
& $NC \uparrow$ & \textbf{0.774}& \textbf{0.724}& \textbf{0.665}& \textbf{0.544}& 0.661 & 0.457 & 0.443 & 0.361 & 0.741 & 0.474 & 0.407 & 0.283 \\
& $DAC \uparrow$ & \textbf{0.885}& \textbf{0.868}& \textbf{0.860}& \textbf{0.896}& 0.739 & 0.798 & 0.810 & 0.837 & 0.817 & 0.832 & 0.839 & 0.851 \\
& $TTC \uparrow$ & \textbf{0.708}& \textbf{0.604}& \textbf{0.550}& \textbf{0.429}& 0.582 & 0.290 & 0.288 & 0.259 & 0.706 & 0.433 & 0.370 & 0.218 \\
& $COM \uparrow$ & 0.828 & 0.720 & 0.672 & 0.585 & \textbf{1.000}& \textbf{1.000}& \textbf{1.000}& \textbf{1.000}& 0.996 & 0.997 & 0.999 & 0.998 \\
& $R_c \uparrow$ & 0.586 & \textbf{0.412}& \textbf{0.404}& \textbf{0.260}& 0.387 & 0.270 & 0.255 & 0.230 & \textbf{0.684}& 0.407 & 0.369 & 0.255 \\
& HD-Score$\uparrow$ & 0.487 & \textbf{0.295}& \textbf{0.273}& \textbf{0.143}& 0.243 & 0.099 & 0.104 & 0.082 & \textbf{0.528}& 0.246 & 0.198 & 0.081 \\

\bottomrule
\end{tabular}

\caption{\textbf{AD Algorithms Evaluating} on the HUGSIM Benchmark}

\label{tab:adtest_all}
\vspace{-0.2cm}
\end{table*}

%% file: supplement/tabs/ablation_2dsemantic.tex
\begin{table*}
\centering
\small
\setlength{\tabcolsep}{14pt}
\begin{tabular}{@{\extracolsep{4pt}}lccc|c@{}} 
\toprule
\multicolumn{1}{c}{} & Seq01 mIoU$_\text{cls}\uparrow$ & Seq02 mIoU$_\text{cls}\uparrow$& Seq03 mIoU$_\text{cls}\uparrow$& Average mIoU$_\text{cls}\uparrow$\\
\hline
Ours w/ $\bS_{\text{2D\_norm}}$ & 0.427 & 0.363 & 0.416 & 0.402 \\
Ours w/ $\bS_{\text{3D\_norm}}$ & \textbf{0.544} &\textbf{0.452} &\textbf{0.520} & \textbf{0.505} \\
\bottomrule
\end{tabular}
\caption{\textbf{Comparison on 3D and 2D Semantic Softmax} on KITTI-360.}
\label{tab:semantic2d_3d}
\vspace{-0.5cm}
\end{table*}

%% file: supplement/figures/depth.tex
\begin{figure}[t!]
     \centering
     \small 
     \setlength{\tabcolsep}{0pt}
     \def\mywidth{4.5cm}
     \begin{tabular}{P{\mywidth}P{\mywidth}}
 
     \includegraphics[width=\mywidth]{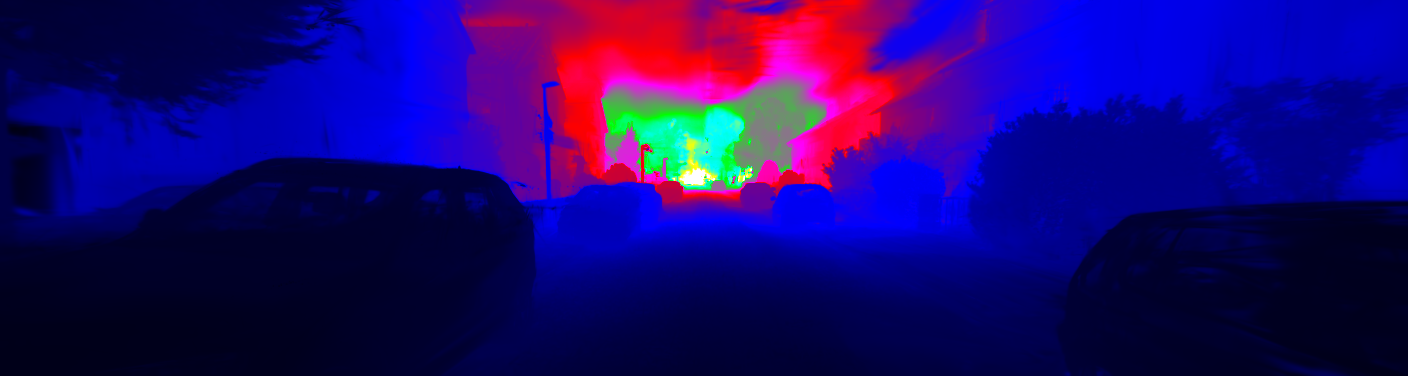}  & \includegraphics[width=\mywidth]{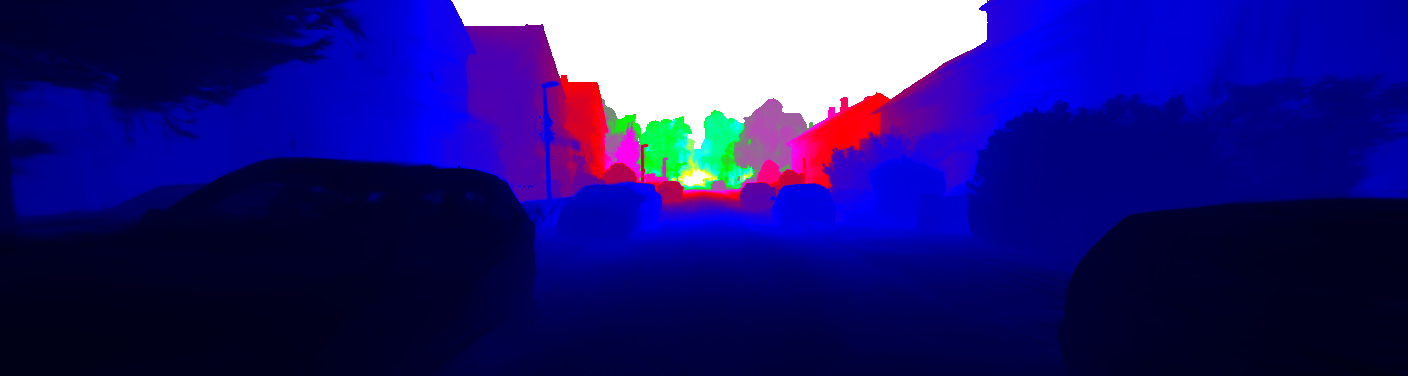}  \\

     \includegraphics[width=\mywidth]{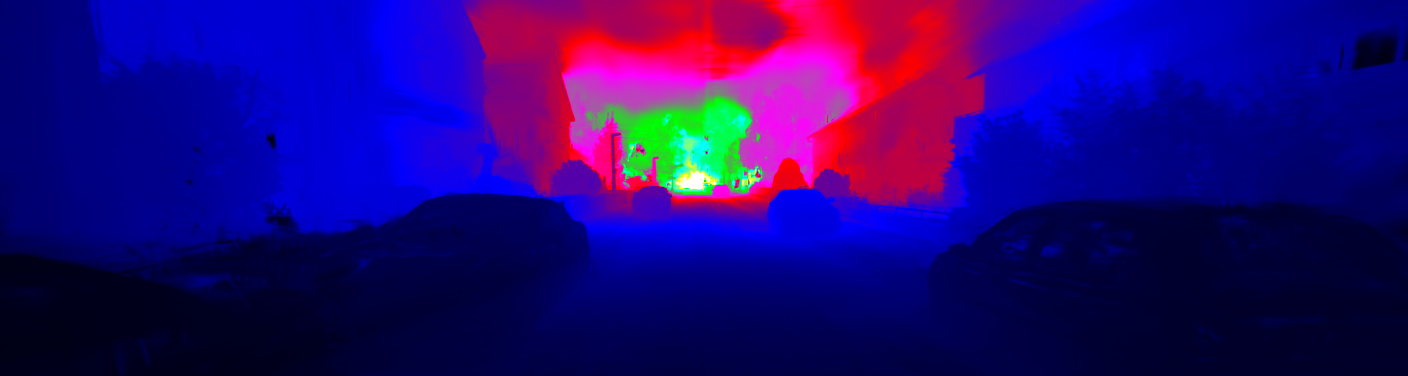}  & \includegraphics[width=\mywidth]{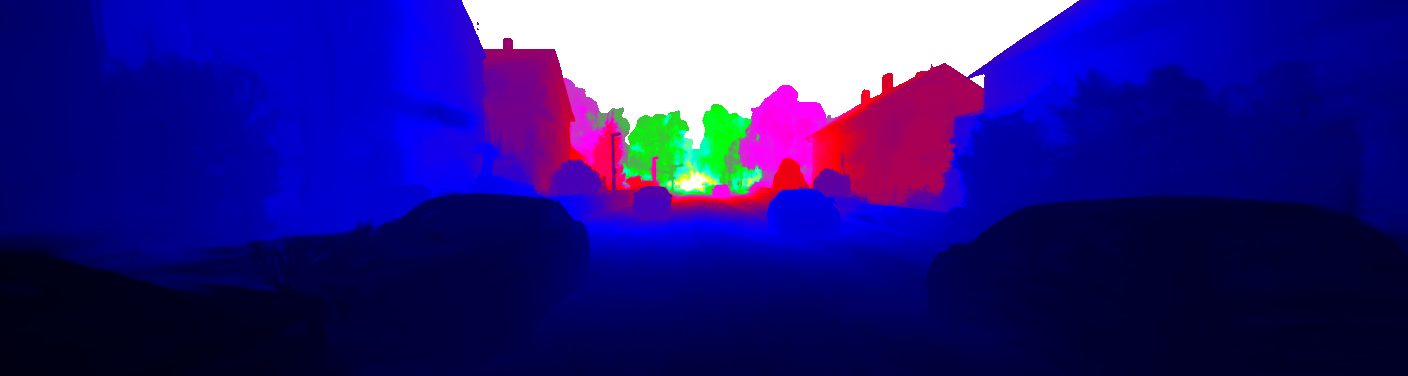}  \\

     \includegraphics[width=\mywidth]{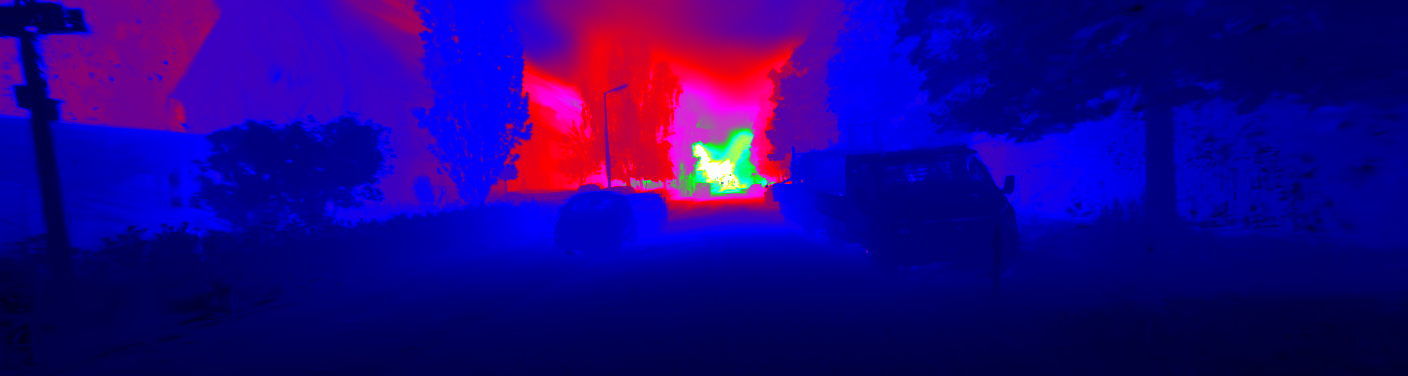}  & \includegraphics[width=\mywidth]{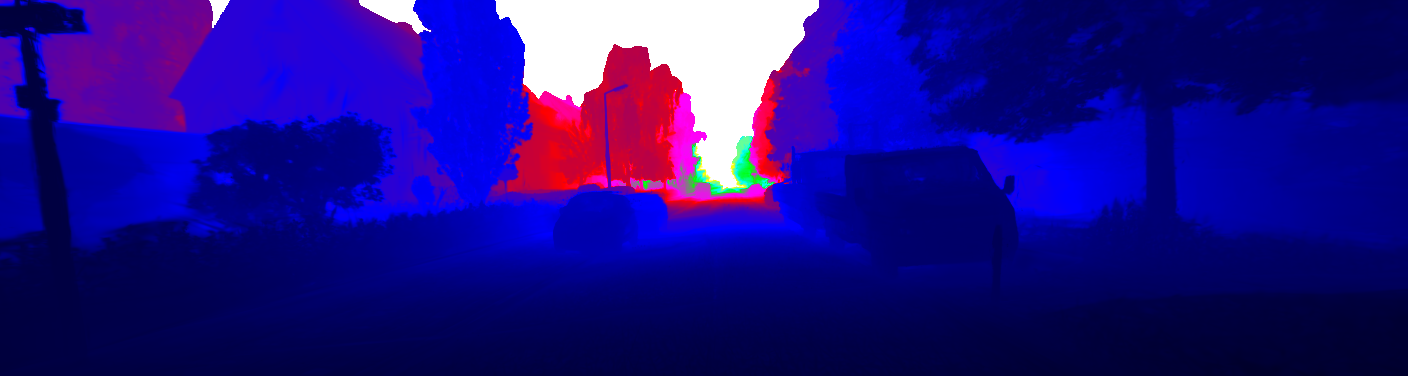}  \\

     \includegraphics[width=\mywidth]{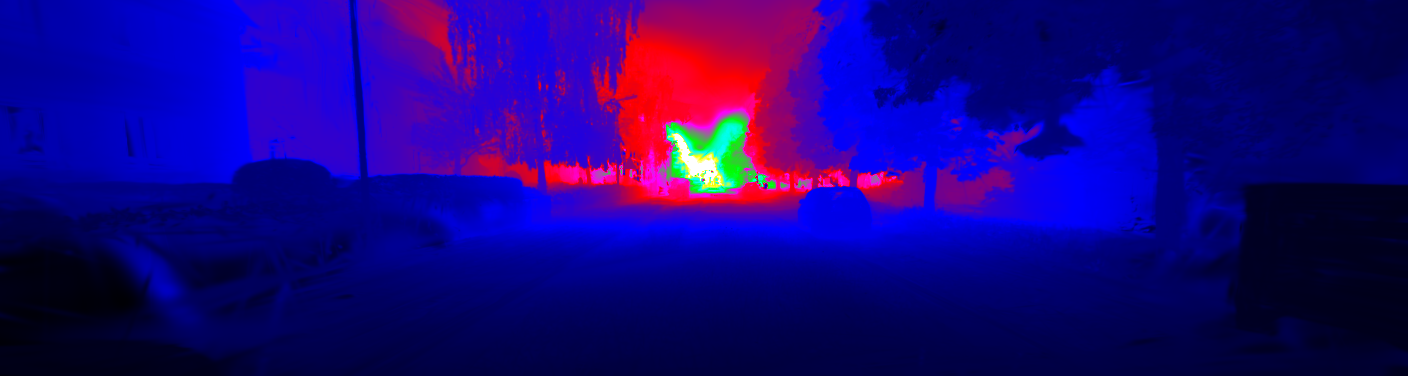}  & \includegraphics[width=\mywidth]{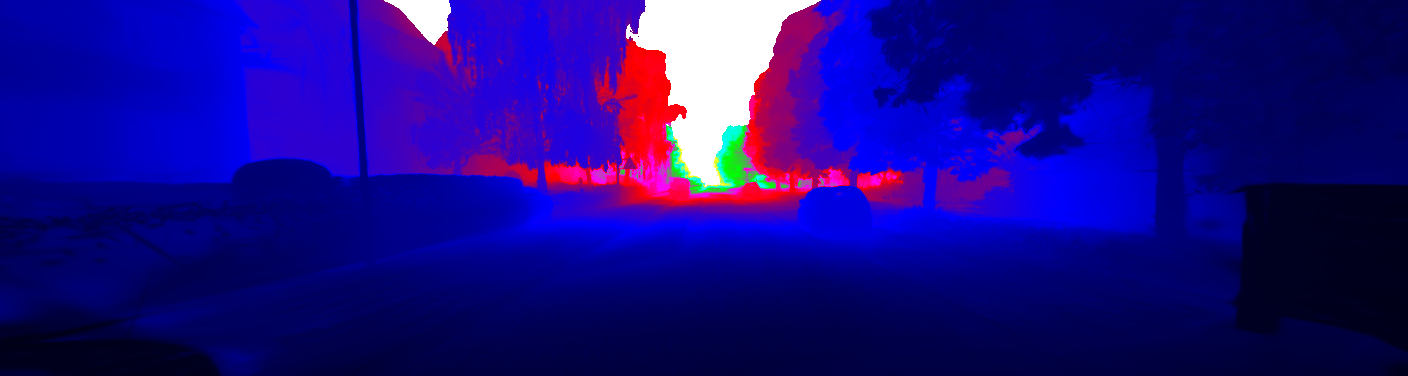}  \\

     \includegraphics[width=\mywidth]{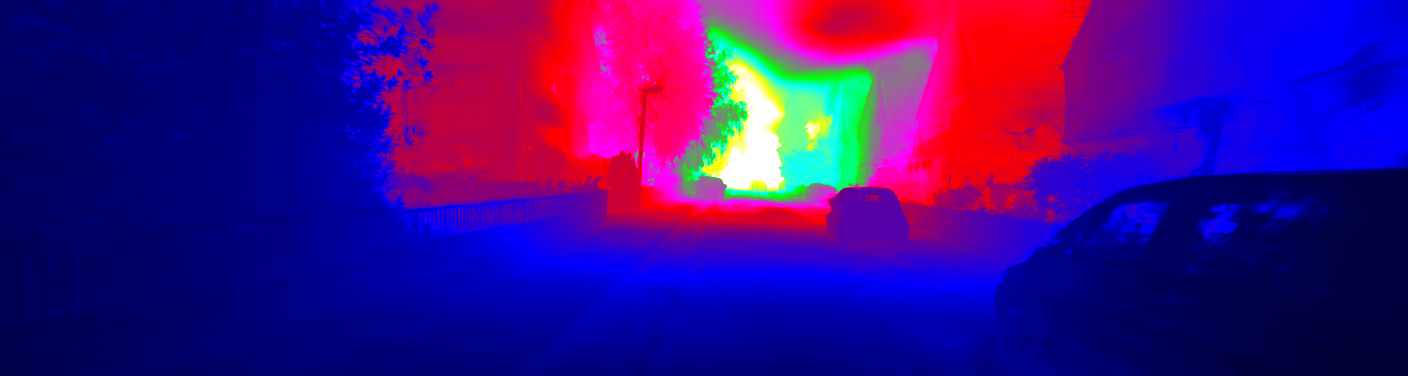}  & \includegraphics[width=\mywidth]{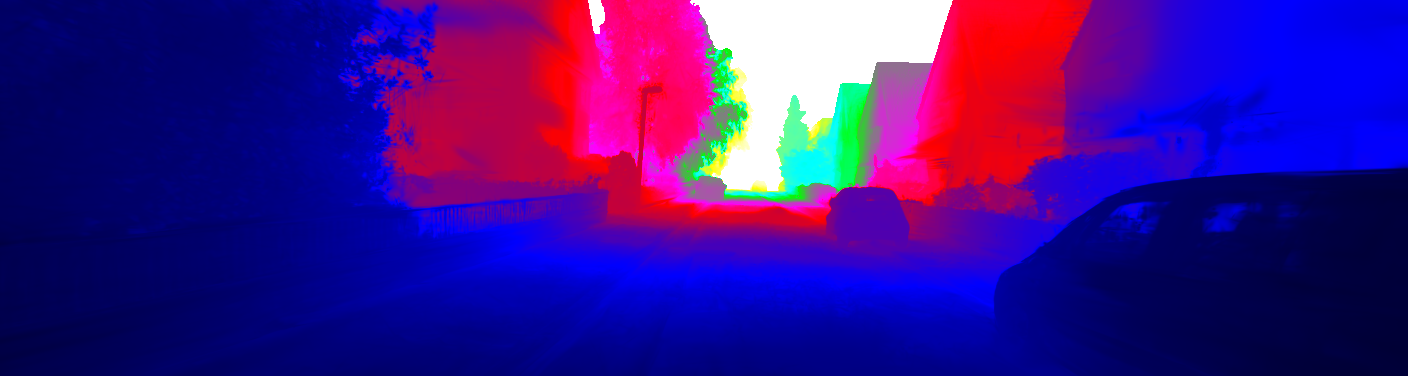}  \\

     \includegraphics[width=\mywidth]{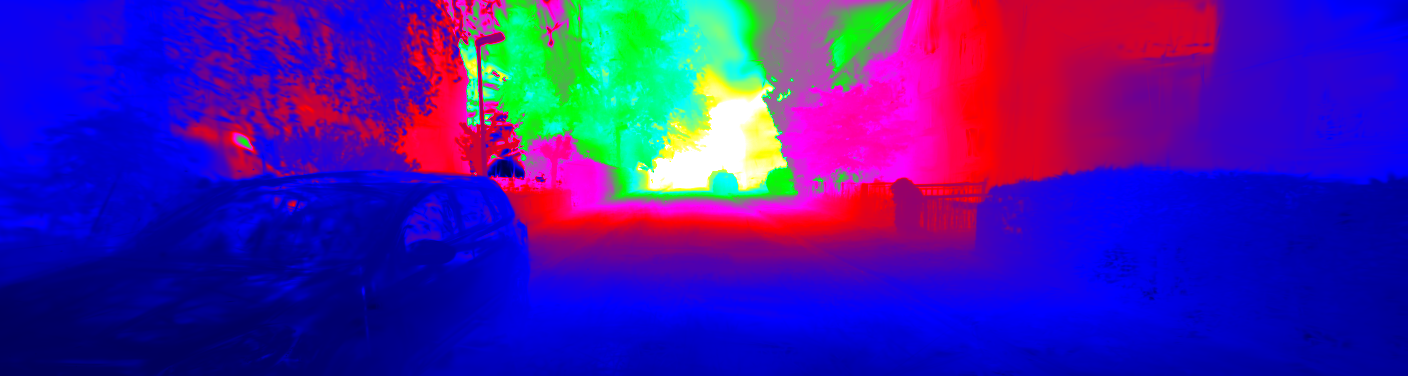}  & \includegraphics[width=\mywidth]{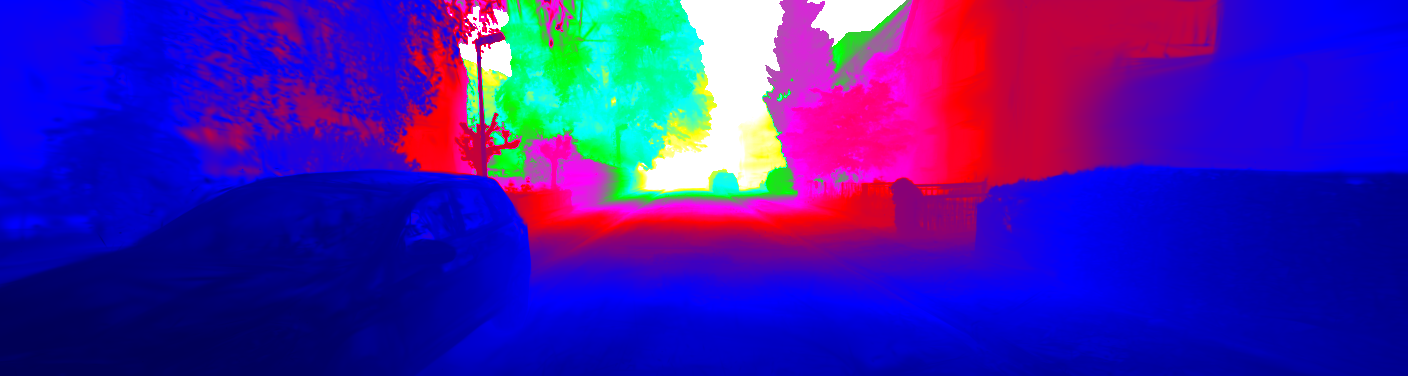}  \\

    \rotatebox{0}{w/o $\cL_\bS$} & \rotatebox{0}{w/ $\cL_\bS$}
     \end{tabular}
     \caption{\textbf{Qualitative Comparison} on depth. In the presence of the semantic loss $\cL_\bS$, We set the sky region's depth infinite based on its semantic label.}
\label{fig:depth_comp}
\vspace{-0.2cm}
\end{figure}

%% file: top.bbl
\begin{thebibliography}{10}
\providecommand{\url}[1]{#1}
\csname url@samestyle\endcsname
\providecommand{\newblock}{\relax}
\providecommand{\bibinfo}[2]{#2}
\providecommand{\BIBentrySTDinterwordspacing}{\spaceskip=0pt\relax}
\providecommand{\BIBentryALTinterwordstretchfactor}{4}
\providecommand{\BIBentryALTinterwordspacing}{\spaceskip=\fontdimen2\font plus
\BIBentryALTinterwordstretchfactor\fontdimen3\font minus \fontdimen4\font\relax}
\providecommand{\BIBforeignlanguage}[2]{{%
\expandafter\ifx\csname l@#1\endcsname\relax
\typeout{** WARNING: IEEEtranS.bst: No hyphenation pattern has been}%
\typeout{** loaded for the language `#1'. Using the pattern for}%
\typeout{** the default language instead.}%
\else
\language=\csname l@#1\endcsname
\fi
#2}}
\providecommand{\BIBdecl}{\relax}
\BIBdecl

\bibitem{agarwal2011building}
S.~Agarwal, Y.~Furukawa, N.~Snavely, I.~Simon, B.~Curless, S.~M. Seitz, and R.~Szeliski, ``Building rome in a day,'' \emph{Communications of the ACM}, vol.~54, no.~10, pp. 105--112, 2011.

\bibitem{barron2021mip}
J.~T. Barron, B.~Mildenhall, M.~Tancik, P.~Hedman, R.~Martin-Brualla, and P.~P. Srinivasan, ``Mip-nerf: A multiscale representation for anti-aliasing neural radiance fields,'' in \emph{Proc. of the IEEE International Conf. on Computer Vision (ICCV)}, 2021, pp. 5855--5864.

\bibitem{binkowski2018demystifying}
M.~Bi{\'n}kowski, D.~J. Sutherland, M.~Arbel, and A.~Gretton, ``Demystifying mmd gans,'' \emph{arXiv preprint arXiv:1801.01401}, 2018.

\bibitem{borse2021inverseform}
S.~Borse, Y.~Wang, Y.~Zhang, and F.~Porikli, ``Inverseform: A loss function for structured boundary-aware segmentation,'' in \emph{Proc. IEEE Conf. on Computer Vision and Pattern Recognition (CVPR)}, 2021, pp. 5901--5911.

\bibitem{bradski2000opencv}
G.~Bradski, ``The opencv library,'' \emph{Dr. Dobb’s Journal of Software Tools}, 2000.

\bibitem{brockman2016openai}
G.~Brockman, ``Openai gym,'' \emph{arXiv preprint arXiv:1606.01540}, 2016.

\bibitem{cabon2020virtual}
Y.~Cabon, N.~Murray, and M.~Humenberger, ``Virtual kitti 2,'' \emph{arXiv preprint arXiv:2001.10773}, 2020.

\bibitem{caesar2020nuscenes}
H.~Caesar, V.~Bankiti, A.~H. Lang, S.~Vora, V.~E. Liong, Q.~Xu, A.~Krishnan, Y.~Pan, G.~Baldan, and O.~Beijbom, ``nuscenes: A multimodal dataset for autonomous driving,'' in \emph{Proc. IEEE Conf. on Computer Vision and Pattern Recognition (CVPR)}, 2020, pp. 11\,621--11\,631.

\bibitem{caesar2021nuplan}
H.~Caesar, J.~Kabzan, K.~S. Tan, W.~K. Fong, E.~Wolff, A.~Lang, L.~Fletcher, O.~Beijbom, and S.~Omari, ``nuplan: A closed-loop ml-based planning benchmark for autonomous vehicles,'' \emph{arXiv preprint arXiv:2106.11810}, 2021.

\bibitem{chen2020category}
X.~Chen, Z.~Dong, J.~Song, A.~Geiger, and O.~Hilliges, ``Category level object pose estimation via neural analysis-by-synthesis,'' in \emph{Proc. of the European Conf. on Computer Vision (ECCV)}.\hskip 1em plus 0.5em minus 0.4em\relax Springer, 2020, pp. 139--156.

\bibitem{chen2024mvsplat}
Y.~Chen, H.~Xu, C.~Zheng, B.~Zhuang, M.~Pollefeys, A.~Geiger, T.-J. Cham, and J.~Cai, ``Mvsplat: Efficient 3d gaussian splatting from sparse multi-view images,'' \emph{arXiv preprint arXiv:2403.14627}, 2024.

\bibitem{chen2023periodic}
Y.~Chen, C.~Gu, J.~Jiang, X.~Zhu, and L.~Zhang, ``Periodic vibration gaussian: Dynamic urban scene reconstruction and real-time rendering,'' \emph{arXiv:2311.18561}, 2023.

\bibitem{chen2024omnire}
Z.~Chen, J.~Yang, J.~Huang, R.~de~Lutio, J.~M. Esturo, B.~Ivanovic, O.~Litany, Z.~Gojcic, S.~Fidler, M.~Pavone \emph{et~al.}, ``Omnire: Omni urban scene reconstruction,'' \emph{arXiv preprint arXiv:2408.16760}, 2024.

\bibitem{Chitta2023PAMI}
K.~Chitta, A.~Prakash, B.~Jaeger, Z.~Yu, K.~Renz, and A.~Geiger, ``Transfuser: Imitation with transformer-based sensor fusion for autonomous driving,'' \emph{IEEE Trans. on Pattern Analysis and Machine Intelligence (PAMI)}, 2023.

\bibitem{codevilla2018end}
F.~Codevilla, M.~M{\"u}ller, A.~L{\'o}pez, V.~Koltun, and A.~Dosovitskiy, ``End-to-end driving via conditional imitation learning,'' in \emph{Proc. IEEE International Conf. on Robotics and Automation (ICRA)}.\hskip 1em plus 0.5em minus 0.4em\relax IEEE, 2018, pp. 4693--4700.

\bibitem{codevilla2019exploring}
F.~Codevilla, E.~Santana, A.~M. L{\'o}pez, and A.~Gaidon, ``Exploring the limitations of behavior cloning for autonomous driving,'' in \emph{Proc. of the IEEE International Conf. on Computer Vision (ICCV)}, 2019, pp. 9329--9338.

\bibitem{dauner2023parting}
D.~Dauner, M.~Hallgarten, A.~Geiger, and K.~Chitta, ``Parting with misconceptions about learning-based vehicle motion planning,'' in \emph{Proc. Conf. on Robot Learning (CoRL)}.\hskip 1em plus 0.5em minus 0.4em\relax PMLR, 2023, pp. 1268--1281.

\bibitem{dauner2024navsim}
D.~Dauner, M.~Hallgarten, T.~Li, X.~Weng, Z.~Huang, Z.~Yang, H.~Li, I.~Gilitschenski, B.~Ivanovic, M.~Pavone \emph{et~al.}, ``Navsim: Data-driven non-reactive autonomous vehicle simulation and benchmarking,'' \emph{arXiv preprint arXiv:2406.15349}, 2024.

\bibitem{dosovitskiy2017carla}
A.~Dosovitskiy, G.~Ros, F.~Codevilla, A.~Lopez, and V.~Koltun, ``Carla: An open urban driving simulator,'' in \emph{Proc. Conf. on Robot Learning (CoRL)}.\hskip 1em plus 0.5em minus 0.4em\relax PMLR, 2017, pp. 1--16.

\bibitem{du20243drealcar}
X.~Du, H.~Sun, S.~Wang, Z.~Wu, H.~Sheng, J.~Ying, M.~Lu, T.~Zhu, K.~Zhan, and X.~Yu, ``3drealcar: An in-the-wild rgb-d car dataset with 360-degree views,'' \emph{arXiv preprint arXiv:2406.04875}, 2024.

\bibitem{feng2024rogs}
Z.~Feng, W.~Wu, and H.~Wang, ``Rogs: Large scale road surface reconstruction based on 2d gaussian splatting,'' \emph{arXiv preprint arXiv:2405.14342}, 2024.

\bibitem{fischer2024dynamic}
T.~Fischer, J.~Kulhanek, S.~R. Bul{\`o}, L.~Porzi, M.~Pollefeys, and P.~Kontschieder, ``Dynamic 3d gaussian fields for urban areas,'' \emph{arXiv preprint arXiv:2406.03175}, 2024.

\bibitem{fu2022panoptic}
X.~Fu, S.~Zhang, T.~Chen, Y.~Lu, L.~Zhu, X.~Zhou, A.~Geiger, and Y.~Liao, ``Panoptic nerf: 3d-to-2d label transfer for panoptic urban scene segmentation,'' in \emph{Proc. of the International Conf. on 3D Vision (3DV)}.\hskip 1em plus 0.5em minus 0.4em\relax IEEE, 2022, pp. 1--11.

\bibitem{gallup2010piecewise}
D.~Gallup, J.-M. Frahm, and M.~Pollefeys, ``Piecewise planar and non-planar stereo for urban scene reconstruction,'' in \emph{Proc. IEEE Conf. on Computer Vision and Pattern Recognition (CVPR)}.\hskip 1em plus 0.5em minus 0.4em\relax IEEE, 2010, pp. 1418--1425.

\bibitem{gao2024vista}
S.~Gao, J.~Yang, L.~Chen, K.~Chitta, Y.~Qiu, A.~Geiger, J.~Zhang, and H.~Li, ``Vista: A generalizable driving world model with high fidelity and versatile controllability,'' \emph{arXiv preprint arXiv:2405.17398}, 2024.

\bibitem{geiger2012we}
A.~Geiger, P.~Lenz, and R.~Urtasun, ``Are we ready for autonomous driving? the kitti vision benchmark suite,'' in \emph{Proc. IEEE Conf. on Computer Vision and Pattern Recognition (CVPR)}.\hskip 1em plus 0.5em minus 0.4em\relax IEEE, 2012, pp. 3354--3361.

\bibitem{gulino2024waymax}
C.~Gulino, J.~Fu, W.~Luo \emph{et~al.}, ``Waymax: An accelerated, data-driven simulator for large-scale autonomous driving research,'' in \emph{Advances in Neural Information Processing Systems (NIPS)}, 2023.

\bibitem{guo2023streetsurf}
J.~Guo, N.~Deng, X.~Li, Y.~Bai, B.~Shi, C.~Wang, C.~Ding, D.~Wang, and Y.~Li, ``Streetsurf: Extending multi-view implicit surface reconstruction to street views,'' \emph{arXiv preprint arXiv:2306.04988}, 2023.

\bibitem{han2024ggs}
H.~Han, K.~Zhou, X.~Long, Y.~Wang, and C.~Xiao, ``Ggs: Generalizable gaussian splatting for lane switching in autonomous driving,'' \emph{arXiv preprint arXiv:2409.02382}, 2024.

\bibitem{Hanselmann2022ECCV}
N.~Hanselmann, K.~Renz, K.~Chitta, A.~Bhattacharyya, and A.~Geiger, ``King: Generating safety-critical driving scenarios for robust imitation via kinematics gradients,'' in \emph{Proc. of the European Conf. on Computer Vision (ECCV)}, 2022.

\bibitem{hu2023gaia}
A.~Hu, L.~Russell, H.~Yeo, Z.~Murez, G.~Fedoseev, A.~Kendall, J.~Shotton, and G.~Corrado, ``Gaia-1: A generative world model for autonomous driving,'' \emph{arXiv preprint arXiv:2309.17080}, 2023.

\bibitem{hu2022monocular}
H.-N. Hu, Y.-H. Yang, T.~Fischer, T.~Darrell, F.~Yu, and M.~Sun, ``Monocular quasi-dense 3d object tracking,'' \emph{IEEE Trans. on Pattern Analysis and Machine Intelligence (PAMI)}, vol.~45, no.~2, pp. 1992--2008, 2022.

\bibitem{hu2023planning}
Y.~Hu, J.~Yang, L.~Chen, K.~Li, C.~Sima, X.~Zhu, S.~Chai, S.~Du, T.~Lin, W.~Wang \emph{et~al.}, ``Planning-oriented autonomous driving,'' in \emph{Proc. IEEE Conf. on Computer Vision and Pattern Recognition (CVPR)}, 2023, pp. 17\,853--17\,862.

\bibitem{huang20242d}
B.~Huang, Z.~Yu, A.~Chen, A.~Geiger, and S.~Gao, ``2d gaussian splatting for geometrically accurate radiance fields,'' in \emph{ACM Trans. on Graphics}, 2024, pp. 1--11.

\bibitem{jiang2023vad}
B.~Jiang, S.~Chen, Q.~Xu, B.~Liao, J.~Chen, H.~Zhou, Q.~Zhang, W.~Liu, C.~Huang, and X.~Wang, ``Vad: Vectorized scene representation for efficient autonomous driving,'' in \emph{Proc. of the IEEE International Conf. on Computer Vision (ICCV)}, 2023, pp. 8340--8350.

\bibitem{karnchanachari2024towards}
N.~Karnchanachari, D.~Geromichalos, K.~S. Tan, N.~Li, C.~Eriksen, S.~Yaghoubi, N.~Mehdipour, G.~Bernasconi, W.~K. Fong, Y.~Guo \emph{et~al.}, ``Towards learning-based planning: The nuplan benchmark for real-world autonomous driving,'' \emph{arXiv preprint arXiv:2403.04133}, 2024.

\bibitem{kerbl20233d}
B.~Kerbl, G.~Kopanas, T.~Leimk{\"u}hler, and G.~Drettakis, ``3d gaussian splatting for real-time radiance field rendering.'' \emph{ACM Trans. on Graphics}, vol.~42, no.~4, pp. 139--1, 2023.

\bibitem{khan2024autosplat}
M.~Khan, H.~Fazlali, D.~Sharma, T.~Cao, D.~Bai, Y.~Ren, and B.~Liu, ``Autosplat: Constrained gaussian splatting for autonomous driving scene reconstruction,'' \emph{arXiv preprint arXiv:2407.02598}, 2024.

\bibitem{kundu2022panoptic}
A.~Kundu, K.~Genova, X.~Yin, A.~Fathi, C.~Pantofaru, L.~J. Guibas, A.~Tagliasacchi, F.~Dellaert, and T.~Funkhouser, ``Panoptic neural fields: A semantic object-aware neural scene representation,'' in \emph{Proc. IEEE Conf. on Computer Vision and Pattern Recognition (CVPR)}, 2022, pp. 12\,871--12\,881.

\bibitem{lafarge2012hybrid}
F.~Lafarge, R.~Keriven, M.~Br{\'e}dif, and H.-H. Vu, ``A hybrid multiview stereo algorithm for modeling urban scenes,'' \emph{IEEE Trans. on Pattern Analysis and Machine Intelligence (PAMI)}, vol.~35, no.~1, pp. 5--17, 2012.

\bibitem{li2024ggrt}
H.~Li, Y.~Gao, D.~Zhang, C.~Wu, Y.~Dai, C.~Zhao, H.~Feng, E.~Ding, J.~Wang, and J.~Han, ``Ggrt: Towards generalizable 3d gaussians without pose priors in real-time,'' \emph{arXiv preprint arXiv:2403.10147}, 2024.

\bibitem{li2022metadrive}
Q.~Li, Z.~Peng, L.~Feng, Q.~Zhang, Z.~Xue, and B.~Zhou, ``Metadrive: Composing diverse driving scenarios for generalizable reinforcement learning,'' \emph{IEEE Trans. on Pattern Analysis and Machine Intelligence (PAMI)}, vol.~45, no.~3, pp. 3461--3475, 2022.

\bibitem{liang2024gs}
Z.~Liang, Q.~Zhang, Y.~Feng, Y.~Shan, and K.~Jia, ``Gs-ir: 3d gaussian splatting for inverse rendering,'' in \emph{Proc. IEEE Conf. on Computer Vision and Pattern Recognition (CVPR)}, 2024, pp. 21\,644--21\,653.

\bibitem{liao2022kitti}
Y.~Liao, J.~Xie, and A.~Geiger, ``Kitti-360: A novel dataset and benchmarks for urban scene understanding in 2d and 3d,'' \emph{IEEE Trans. on Pattern Analysis and Machine Intelligence (PAMI)}, vol.~45, no.~3, pp. 3292--3310, 2022.

\bibitem{ljungbergh2024neuroncap}
W.~Ljungbergh, A.~Tonderski, J.~Johnander, H.~Caesar, K.~{\AA}str{\"o}m, M.~Felsberg, and C.~Petersson, ``Neuroncap: Photorealistic closed-loop safety testing for autonomous driving,'' \emph{arXiv preprint arXiv:2404.07762}, 2024.

\bibitem{lu2023urban}
F.~Lu, Y.~Xu, G.~Chen, H.~Li, K.-Y. Lin, and C.~Jiang, ``Urban radiance field representation with deformable neural mesh primitives,'' in \emph{Proc. of the IEEE International Conf. on Computer Vision (ICCV)}, 2023, pp. 465--476.

\bibitem{martin2021nerf}
R.~Martin-Brualla, N.~Radwan, M.~S. Sajjadi, J.~T. Barron, A.~Dosovitskiy, and D.~Duckworth, ``Nerf in the wild: Neural radiance fields for unconstrained photo collections,'' in \emph{Proc. IEEE Conf. on Computer Vision and Pattern Recognition (CVPR)}, 2021, pp. 7210--7219.

\bibitem{mei2007single}
C.~Mei and P.~Rives, ``Single view point omnidirectional camera calibration from planar grids,'' in \emph{Proc. IEEE International Conf. on Robotics and Automation (ICRA)}.\hskip 1em plus 0.5em minus 0.4em\relax IEEE, 2007, pp. 3945--3950.

\bibitem{miao2024efficient}
S.~Miao, J.~Huang, D.~Bai, W.~Qiu, B.~Liu, A.~Geiger, and Y.~Liao, ``Efficient depth-guided urban view synthesis,'' \emph{arXiv preprint arXiv:2407.12395}, 2024.

\bibitem{moenne20243d}
N.~Moenne-Loccoz, A.~Mirzaei, O.~Perel, R.~de~Lutio, J.~M. Esturo, G.~State, S.~Fidler, N.~Sharp, and Z.~Gojcic, ``3d gaussian ray tracing: Fast tracing of particle scenes,'' \emph{arXiv preprint arXiv:2407.07090}, 2024.

\bibitem{3dgrt2024}
------, ``3d gaussian ray tracing: Fast tracing of particle scenes,'' \emph{ACM Trans. on Graphics}, 2024.

\bibitem{ost2021neural}
J.~Ost, F.~Mannan, N.~Thuerey, J.~Knodt, and F.~Heide, ``Neural scene graphs for dynamic scenes,'' in \emph{Proc. IEEE Conf. on Computer Vision and Pattern Recognition (CVPR)}, 2021, pp. 2856--2865.

\bibitem{rematas2022urban}
K.~Rematas, A.~Liu, P.~P. Srinivasan, J.~T. Barron, A.~Tagliasacchi, T.~Funkhouser, and V.~Ferrari, ``Urban radiance fields,'' in \emph{Proc. IEEE Conf. on Computer Vision and Pattern Recognition (CVPR)}, 2022, pp. 12\,932--12\,942.

\bibitem{samak2021control}
C.~V. Samak, T.~V. Samak, and S.~Kandhasamy, ``Control strategies for autonomous vehicles,'' in \emph{Autonomous driving and advanced driver-assistance systems (ADAS)}.\hskip 1em plus 0.5em minus 0.4em\relax CRC Press, 2021, pp. 37--86.

\bibitem{schonberger2016structure}
J.~L. Schonberger and J.-M. Frahm, ``Structure-from-motion revisited,'' in \emph{Proc. IEEE Conf. on Computer Vision and Pattern Recognition (CVPR)}, 2016, pp. 4104--4113.

\bibitem{shi2024dhgs}
X.~Shi, L.~Chen, P.~Wei, X.~Wu, T.~Jiang, Y.~Luo, and L.~Xie, ``Dhgs: Decoupled hybrid gaussian splatting for driving scene,'' \emph{arXiv preprint arXiv:2407.16600}, 2024.

\bibitem{shi2023gir}
Y.~Shi, Y.~Wu, C.~Wu, X.~Liu, C.~Zhao, H.~Feng, J.~Liu, L.~Zhang, J.~Zhang, B.~Zhou \emph{et~al.}, ``Gir: 3d gaussian inverse rendering for relightable scene factorization,'' \emph{arXiv preprint arXiv:2312.05133}, 2023.

\bibitem{Sun2020CVPR}
P.~Sun, H.~Kretzschmar, X.~Dotiwalla \emph{et~al.}, ``Scalability in perception for autonomous driving: Waymo open dataset,'' in \emph{Proc. IEEE Conf. on Computer Vision and Pattern Recognition (CVPR)}, June 2020.

\bibitem{tancik2022block}
M.~Tancik, V.~Casser, X.~Yan, S.~Pradhan, B.~Mildenhall, P.~P. Srinivasan, J.~T. Barron, and H.~Kretzschmar, ``Block-nerf: Scalable large scene neural view synthesis,'' in \emph{Proc. IEEE Conf. on Computer Vision and Pattern Recognition (CVPR)}, 2022, pp. 8248--8258.

\bibitem{tancik2023nerfstudio}
M.~Tancik, E.~Weber, E.~Ng, R.~Li, B.~Yi, T.~Wang, A.~Kristoffersen, J.~Austin, K.~Salahi, A.~Ahuja \emph{et~al.}, ``Nerfstudio: A modular framework for neural radiance field development,'' in \emph{ACM Trans. on Graphics}, 2023, pp. 1--12.

\bibitem{tonderski2024neurad}
A.~Tonderski, C.~Lindstr{\"o}m, G.~Hess, W.~Ljungbergh, L.~Svensson, and C.~Petersson, ``Neurad: Neural rendering for autonomous driving,'' in \emph{Proc. IEEE Conf. on Computer Vision and Pattern Recognition (CVPR)}, 2024, pp. 14\,895--14\,904.

\bibitem{treiber2000congested}
M.~Treiber, A.~Hennecke, and D.~Helbing, ``Congested traffic states in empirical observations and microscopic simulations,'' \emph{Physical review E}, vol.~62, no.~2, p. 1805, 2000.

\bibitem{turki2023suds}
H.~Turki, J.~Y. Zhang, F.~Ferroni, and D.~Ramanan, ``Suds: Scalable urban dynamic scenes,'' in \emph{Proc. IEEE Conf. on Computer Vision and Pattern Recognition (CVPR)}, 2023, pp. 12\,375--12\,385.

\bibitem{wang2024freevs}
Q.~Wang, L.~Fan, Y.~Wang, Y.~Chen, and Z.~Zhang, ``Freevs: Generative view synthesis on free driving trajectory,'' \emph{arXiv preprint arXiv:2410.18079}, 2024.

\bibitem{wang2023drivedreamer}
X.~Wang, Z.~Zhu, G.~Huang, X.~Chen, J.~Zhu, and J.~Lu, ``Drivedreamer: Towards real-world-driven world models for autonomous driving,'' \emph{arXiv preprint arXiv:2309.09777}, 2023.

\bibitem{wen2023limsim}
L.~Wenl, D.~Fu, S.~Mao, P.~Cai, M.~Dou, Y.~Li, and Y.~Qiao, ``Limsim: A long-term interactive multi-scenario traffic simulator,'' in \emph{Proc. IEEE Conf. on Intelligent Transportation Systems (ITSC)}.\hskip 1em plus 0.5em minus 0.4em\relax IEEE, 2023, pp. 1255--1262.

\bibitem{wimbauer2023behind}
F.~Wimbauer, N.~Yang, C.~Rupprecht, and D.~Cremers, ``Behind the scenes: Density fields for single view reconstruction,'' in \emph{Proc. IEEE Conf. on Computer Vision and Pattern Recognition (CVPR)}, 2023, pp. 9076--9086.

\bibitem{wu2023mars}
Z.~Wu, T.~Liu, L.~Luo, Z.~Zhong, J.~Chen, H.~Xiao, C.~Hou, H.~Lou, Y.~Chen, R.~Yang \emph{et~al.}, ``Mars: An instance-aware, modular and realistic simulator for autonomous driving,'' in \emph{International Conference on Artificial Intelligence (CICAI)}.\hskip 1em plus 0.5em minus 0.4em\relax Springer, 2023, pp. 3--15.

\bibitem{xiao2021pandaset}
P.~Xiao, Z.~Shao, S.~Hao, Z.~Zhang, X.~Chai, J.~Jiao, Z.~Li, J.~Wu, K.~Sun, K.~Jiang \emph{et~al.}, ``Pandaset: Advanced sensor suite dataset for autonomous driving,'' in \emph{Proc. IEEE Conf. on Intelligent Transportation Systems (ITSC)}.\hskip 1em plus 0.5em minus 0.4em\relax IEEE, 2021, pp. 3095--3101.

\bibitem{yan2024street}
Y.~Yan, H.~Lin, C.~Zhou, W.~Wang, H.~Sun, K.~Zhan, X.~Lang, X.~Zhou, and S.~Peng, ``Street gaussians: Modeling dynamic urban scenes with gaussian splatting,'' in \emph{Proc. of the European Conf. on Computer Vision (ECCV)}, 2024.

\bibitem{yang2023emernerf}
J.~Yang, B.~Ivanovic, O.~Litany, X.~Weng, S.~W. Kim, B.~Li, T.~Che, D.~Xu, S.~Fidler, M.~Pavone \emph{et~al.}, ``Emernerf: Emergent spatial-temporal scene decomposition via self-supervision,'' \emph{arXiv preprint arXiv:2311.02077}, 2023.

\bibitem{yang2024generalized}
J.~Yang, S.~Gao, Y.~Qiu, L.~Chen, T.~Li, B.~Dai, K.~Chitta, P.~Wu, J.~Zeng, P.~Luo \emph{et~al.}, ``Generalized predictive model for autonomous driving,'' in \emph{Proc. IEEE Conf. on Computer Vision and Pattern Recognition (CVPR)}, 2024, pp. 14\,662--14\,672.

\bibitem{yang2024drivearena}
X.~Yang, L.~Wen, Y.~Ma, J.~Mei, X.~Li, T.~Wei, W.~Lei, D.~Fu, P.~Cai, M.~Dou, B.~Shi, L.~He, Y.~Liu, and Y.~Qiao, ``Drivearena: A closed-loop generative simulation platform for autonomous driving,'' \emph{arXiv preprint arXiv:2408.00415}, 2024.

\bibitem{yang2023unisim}
Z.~Yang, Y.~Chen, J.~Wang, S.~Manivasagam, W.-C. Ma, A.~J. Yang, and R.~Urtasun, ``Unisim: A neural closed-loop sensor simulator,'' in \emph{Proc. IEEE Conf. on Computer Vision and Pattern Recognition (CVPR)}, 2023, pp. 1389--1399.

\bibitem{yang2024deformable}
Z.~Yang, X.~Gao, W.~Zhou, S.~Jiao, Y.~Zhang, and X.~Jin, ``Deformable 3d gaussians for high-fidelity monocular dynamic scene reconstruction,'' in \emph{Proc. IEEE Conf. on Computer Vision and Pattern Recognition (CVPR)}, 2024, pp. 20\,331--20\,341.

\bibitem{yu2024viewcrafter}
W.~Yu, J.~Xing, L.~Yuan, W.~Hu, X.~Li, Z.~Huang, X.~Gao, T.-T. Wong, Y.~Shan, and Y.~Tian, ``Viewcrafter: Taming video diffusion models for high-fidelity novel view synthesis,'' \emph{arXiv preprint arXiv:2409.02048}, 2024.

\bibitem{yu2024sgd}
Z.~Yu, H.~Wang, J.~Yang, H.~Wang, Z.~Xie, Y.~Cai, J.~Cao, Z.~Ji, and M.~Sun, ``Sgd: Street view synthesis with gaussian splatting and diffusion prior,'' \emph{arXiv preprint arXiv:2403.20079}, 2024.

\bibitem{zhang2023cat}
L.~Zhang, Z.~Peng, Q.~Li, and B.~Zhou, ``Cat: Closed-loop adversarial training for safe end-to-end driving,'' in \emph{Proc. Conf. on Robot Learning (CoRL)}, 2023.

\bibitem{zhang2018unreasonable}
R.~Zhang, P.~Isola, A.~A. Efros, E.~Shechtman, and O.~Wang, ``The unreasonable effectiveness of deep features as a perceptual metric,'' in \emph{Proc. IEEE Conf. on Computer Vision and Pattern Recognition (CVPR)}, 2018, pp. 586--595.

\bibitem{zhang2023nerflets}
X.~Zhang, A.~Kundu, T.~Funkhouser, L.~Guibas, H.~Su, and K.~Genova, ``Nerflets: Local radiance fields for efficient structure-aware 3d scene representation from 2d supervision,'' in \emph{Proc. IEEE Conf. on Computer Vision and Pattern Recognition (CVPR)}, 2023, pp. 8274--8284.

\bibitem{zhi2021place}
S.~Zhi, T.~Laidlow, S.~Leutenegger, and A.~J. Davison, ``In-place scene labelling and understanding with implicit scene representation,'' in \emph{Proc. of the IEEE International Conf. on Computer Vision (ICCV)}, 2021, pp. 15\,838--15\,847.

\bibitem{zhou2019does}
B.~Zhou, P.~Kr{\"a}henb{\"u}hl, and V.~Koltun, ``Does computer vision matter for action?'' \emph{Science Robotics}, vol.~4, no.~30, p. eaaw6661, 2019.

\bibitem{Zhou2024CVPR}
H.~Zhou, J.~Shao, L.~Xu, D.~Bai, W.~Qiu, B.~Liu, Y.~Wang, A.~Geiger, and Y.~Liao, ``Hugs: Holistic urban 3d scene understanding via gaussian splatting,'' in \emph{Proc. IEEE Conf. on Computer Vision and Pattern Recognition (CVPR)}, June 2024, pp. 21\,336--21\,345.

\bibitem{zhou2024drivinggaussian}
X.~Zhou, Z.~Lin, X.~Shan, Y.~Wang, D.~Sun, and M.-H. Yang, ``Drivinggaussian: Composite gaussian splatting for surrounding dynamic autonomous driving scenes,'' in \emph{Proc. IEEE Conf. on Computer Vision and Pattern Recognition (CVPR)}, 2024, pp. 21\,634--21\,643.

\bibitem{zwicker2002ewa}
M.~Zwicker, H.~Pfister, J.~Van~Baar, and M.~Gross, ``Ewa splatting,'' \emph{IEEE Transactions on Visualization and Computer Graphic (TVCG)}, vol.~8, no.~3, pp. 223--238, 2002.

\end{thebibliography}
